\documentclass[10pt,journal,compsoc]{ieee/IEEEtran}
\ifCLASSOPTIONcompsoc
  \usepackage[nocompress]{cite}
\else
  \usepackage{cite}
\fi
\ifCLASSINFOpdf
\else
\fi
\usepackage[hidelinks]{hyperref}
\usepackage{multirow}
\usepackage{makecell}
\usepackage{booktabs,array}
\usepackage{balance}
\usepackage{wrapfig}

\usepackage{qtree}
\usepackage{forest}

\usepackage{pgfplots}
\pgfplotsset{compat=1.17}
\usepackage{balance}
\usepackage{tikz}
\usepackage{tikz-3dplot} %
\usetikzlibrary{calc,patterns,arrows.meta,positioning,shapes,backgrounds,ducks}
\usepackage{wrapfig}

\usepackage{enumitem}
\usepackage{adjustbox}
\usepackage{colortbl}

\usepackage{placeins}

\usepackage{ragged2e}
\usepackage{pdfpages}
\usepackage{lipsum} 
\usepackage{listings} 
\usepackage{float} 
\usepackage{amssymb}
\usepackage{amsmath,amsthm,dsfont}
\usepackage{acronym}
\usepackage{subcaption} 
\usepackage{caption}

\usepackage[english]{babel}

\renewcommand*{\eqref}[1]{(\hyperref[#1]{\ref*{#1}})}
\newcommand*{\equaref}[1]{Eq.~\hyperref[#1]{\ref*{#1}}}
\newcommand*{\chref}[1]{Chapter~\hyperref[#1]{\ref*{#1}}}
\newcommand*{\secref}[1]{Sec.~\hyperref[#1]{\ref*{#1}}}
\newcommand*{\figref}[1]{Fig.~\hyperref[#1]{\ref*{#1}}}
\newcommand*{\figsref}[1]{Figs.~\hyperref[#1]{\ref*{#1}}}
\newcommand*{\tabref}[1]{Tab.~\hyperref[#1]{\ref*{#1}}}
\newcommand*{\Subref}[1]{\hyperref[#1]{(\subref*{#1})}}

\newcommand{\bR}{\mathbb{R}}
\newcommand{\bRt}{\mathbb{R}^3}

\newcommand{\cP}{\mathcal{P}}
\newcommand{\cp}{p}
\newcommand{\cPr}{\mathcal{P}^r}
\newcommand{\cPg}{\mathcal{P}^g}

\newcommand{\cE}{\mathcal{E}}
\newcommand{\cM}{\mathcal{M}}

\newcommand{\cV}{\mathcal{V}}
\newcommand{\cF}{\mathcal{F}}
\newcommand{\cS}{\mathcal{S}^g}
\newcommand{\cSr}{{\mathcal{S}^r}}
\newcommand{\cSg}{{\mathcal{S}^g}}
\newcommand{\cVr}{{\mathcal{V}^r}}
\newcommand{\cVg}{{\mathcal{V}^g}}

\usepackage{xspace}
\newcommand{\etal}{\textit{et~al.}\@\xspace}
\newcommand{\ie}{\textit{i.e.,}\@\xspace}
\newcommand{\eg}{\textit{e.g.,}\@\xspace}
\DeclareOldFontCommand{\bf}{\normalfont\bfseries}{\mathbf}

\DeclareMathOperator*{\argmin}{argmin}

\makeatletter
\newcommand*{\glsplainhyperlink}[2]{%
  \colorlet{currenttext}{.}
  \colorlet{currentlink}{\@linkcolor}
  \hypersetup{linkcolor=currenttext}
  \hyperlink{#1}{#2}%
  \hypersetup{linkcolor=currentlink}
}
\let\@glslink\glsplainhyperlink
\makeatother

\lccode`2=`2
\definecolor{tempblue}{rgb}{0.20,0.56,1}
\definecolor{tempgreen}{rgb}{0.42,0.95,0.9}
\definecolor{tempred}{rgb}{1,0.6,.95}
\definecolor{tempyellow}{rgb}{0.85, 0.85, 0.15}
\definecolor{tempdarkblue}{rgb}{0.0, 0.0, 0.30}
\definecolor{MyLightGray}{rgb}{0.90,0.90,0.90}

\DeclareOldFontCommand{\bf}{\normalfont\bfseries}{\mathbf}

\newcommand{\LOIC}[1]{\textcolor{black}{#1}}
\newcommand{\RAPH}[1]{\textcolor{black}{#1}}

\definecolor{uge-blue}{RGB}{40,54,117}

\makeatletter
\newcommand{\definetrim}[2]{%
  \define@key{Gin}{#1}[]{\setkeys{Gin}{trim=#2,clip}}%
}
\makeatother

\renewcommand{\bR}{\mathbb{R}}
\renewcommand{\bRt}{\mathbb{R}^3}

\renewcommand{\cP}{\mathcal{P}}
\renewcommand{\cPr}{\mathcal{P}^r}
\renewcommand{\cPg}{\mathcal{P}^g}

\renewcommand{\cE}{\mathcal{E}}
\renewcommand{\cM}{\mathcal{M}}

\renewcommand{\cV}{\mathcal{V}}
\renewcommand{\cF}{\mathcal{F}}
\renewcommand{\cS}{\mathcal{S}^g}
\renewcommand{\cSr}{{\mathcal{S}^r}}
\renewcommand{\cSg}{{\mathcal{S}^g}}

\usepackage{xspace}
\renewcommand{\etal}{\textit{et~al.}\@\xspace}
\renewcommand{\ie}{\textit{i.e.,}\@\xspace}
\renewcommand{\eg}{\textit{e.g.,}\@\xspace}
\newcommand{\cf}{\textit{cf.}\@\xspace}
\DeclareOldFontCommand{\bf}{\normalfont\bfseries}{\mathbf}

\newcommand{\optim}{$^*$}

\makeatletter
\renewcommand*{\glsplainhyperlink}[2]{%
  \colorlet{currenttext}{.}
  \colorlet{currentlink}{\@linkcolor}
  \hypersetup{linkcolor=currenttext}
  \hyperlink{#1}{#2}%
  \hypersetup{linkcolor=currentlink}
}
\let\@glslink\glsplainhyperlink
\makeatother

\lccode`2=`2
\definecolor{tempblue}{rgb}{0.20,0.56,1}
\definecolor{tempgreen}{rgb}{0.42,0.95,0.9}
\definecolor{tempred}{rgb}{1,0.6,.95}
\definecolor{tempyellow}{rgb}{0.85, 0.85, 0.15}
\definecolor{tempdarkblue}{rgb}{0.0, 0.0, 0.30}
\definecolor{MyLightGray}{rgb}{0.90,0.90,0.90}

\DeclareOldFontCommand{\bf}{\normalfont\bfseries}{\mathbf}

\definecolor{uge-blue}{RGB}{40,54,117}

\makeatletter
\renewcommand{\definetrim}[2]{%
  \define@key{Gin}{#1}[]{\setkeys{Gin}{trim=#2,clip}}%
}
\makeatother

\definecolor{SURFCOLOR}{RGB}{137, 186, 217}

\usepackage{alphalph}
\begin{document}

\newacro{2d}[$2$D]{two-dimensional}
\newacro{3d}[$3$D]{three-dimensional}
\newacro{2dt}[$2$DT]{2D Delaunay triangulation}
\newacro{3dt}[$3$DT]{3D Delaunay tetrahedralisation}
\newacro{dt}[DT]{Delaunay triangulation}

\newacro{psr}[PSR]{Poisson Surface Reconstruction}
\newacro{spsr}[SPSR]{Screened Poisson Surface Reconstruction}
\newacro{resr}[RESR]{Robust and Efficient Surface Reconstruction}

\newacro{ier}[IER]{intrinsic-extrinsic ratio}
\newacro{igr}[IGR]{Implicit Geometric Regularisation}
\newacro{lig}[LIG]{Local Implicit Grid}
\newacro{p2m}[P2M]{Point2Mesh}
\newacro{sap}[SAP]{Shape As Points}
\newacro{sapopt}[SAP\optim]{optimisation-based Shape As Points}
\newacro{p2s}[P2S]{Points2Surf}
\newacro{onet}[ONet]{Occupancy Networks}
\newacro{conet}[ConvONet]{Convolutional Occupancy Networks}
\newacro{dgnn}[DGNN]{Delaunay-Graph Neural Network}
\newacro{poco}[POCO]{Point Convolution for Surface Reconstruction}
\newacro{nksr}[NKSR]{Neural Kernel Surface Reconstruction}
\newacro{ngs}[NGS]{Neural Galerkin Solver}
\newacro{nkf}[NKF]{Neural Kernel Fields}

\newacro{iou}[IoU]{intersection over union}
\newacro{cd}[CD]{Chamfer distance}
\newacro{nc}[NC]{normal consistency}
\newacro{noc}[NoC]{number of components}

\newacro{pca}[PCA]{principal component analysis}
\newacro{mlp}[MLP]{multilayer perceptron}

\newacro{bce}[BCE]{binary cross entropy}
\newacro{mse}[MSE]{mean square error}

\newacro{cnn}[CNN]{convolutional neural network}
\newacroplural{cnn}[CNNs]{convolutional neural networks}

\newacro{gnn}[GNN]{graph neural network}
\newacroplural{gnn}[GNNs]{graph neural networks}

\newacro{lod}[LOD]{level of detail}
\newacroplural{lod}[LODs]{levels of detail}
\newacro{mvs}[MVS]{multi-view stereo}
\newacro{sfm}[SfM]{structure from motion}
\newacro{lidar}[LiDAR]{Light Detection and Ranging}
\newacro{als}[ALS]{airborne laser scanning}

\newacro{nerf}[NeRF]{neural radiance field}

\newacro{sdf}[SDF]{signed distance function}
\newacro{of}[OF]{occupancy function}

\newacro{fcn}[FCN]{fully-connected network}
\newacro{dsr}[NSR]{neural surface reconstruction}

\newacro{mise}[MISE]{multi-resolution isosurface extraction}

\newacro{tft}[TFT]{triangle-from-tetrahedra}

\newacro{nif}[NIF]{neural implicit function}


%
\title{A Survey and Benchmark of Automatic\\Surface Reconstruction from Point Clouds}

\author{Raphael~Sulzer,
        Renaud~Marlet,
        Bruno~Vallet~and~Loic~Landrieu
\IEEEcompsocitemizethanks{%
\IEEEcompsocthanksitem 
R.~Sulzer is affiliated with Centre INRIA d'Universit\'e C\^ote d'Azur, Sophia Antipolis, France. E-mail: \href{mailto:raphaelsulzer@gmx.de}{raphaelsulzer@gmx.de}.
\IEEEcompsocthanksitem
R.~Marlet is affiliated with LIGM, Ecole des Ponts, Univ Gustave Eiffel, CNRS, Marne-la-Vall\'ee, France and Valeo.ai, Paris, France.
\IEEEcompsocthanksitem B.~Vallet is affiliated with LASTIG, Univ Gustave Eiffel, IGN-ENSG, F-94160 Saint-Mandé, France.
\IEEEcompsocthanksitem L.~Landrieu is affiliated with LIGM and LASTIG.}
\thanks{Corresponding author: R.~Sulzer
}}

\IEEEtitleabstractindextext{%
\begin{abstract}
We present a comprehensive survey and benchmark of both traditional and learning-based methods for surface reconstruction from point clouds. This task is particularly challenging for real-world acquisitions due to factors such as noise, outliers, non-uniform sampling, and missing data. Traditional approaches often simplify the problem by imposing handcrafted priors on either the input point clouds or the resulting surface, a process that can require tedious hyperparameter tuning. In contrast, deep learning models have the capability to directly learn the properties of input point clouds and desired surfaces from data. 
We study the influence of handcrafted and learned priors on the precision and robustness of surface reconstruction techniques. We evaluate various time-tested and contemporary methods in a standardized manner. When both trained and evaluated on point clouds with identical characteristics, the learning-based models consistently produce higher-quality surfaces compared to their traditional counterparts---even in scenarios involving novel shape categories. However, traditional methods demonstrate greater resilience to the diverse anomalies commonly found in real-world 3D acquisitions. For the benefit of the research community, we make our code and datasets available, inviting further enhancements to learning-based surface reconstruction. This can be accessed at \url{https://github.com/raphaelsulzer/dsr-benchmark}.
\end{abstract}

}

\maketitle

\IEEEdisplaynontitleabstractindextext

%
\IEEEpeerreviewmaketitle


\IEEEraisesectionheading{\section{Introduction}\label{sec:introduction}}

\IEEEPARstart{M}{odern} 
\ac{3d} acquisition technology, such as range scanning or \ac{mvs} allows us to capture detailed geometric information of real scenes in the form of point clouds.
However, sets of \ac{3d} points are usually not sufficient for modeling complex physical processes, such as fluid dynamics or environmental phenomena. A variety of applications in science and engineering require a continuous surface representation of objects or scenes.
As a result, transforming point clouds into surface models---a process known as surface reconstruction---is a central problem in digital geometry processing.
In this paper, we offer a comprehensive survey and benchmark of traditional and learning-based methods designed to address this problem.

Without prior information about the sought surface, the task of surface reconstruction from point clouds becomes ill-posed: there are infinitely many surfaces that can potentially fit the given points. 
As illustrated in \figref{fig:ch2:3_problems}, acquisition defects such as non-uniform sampling, noise, outliers, or missing data further complicate the reconstruction of a geometrically and topologically accurate surface \cite{Berger_survey}.

Traditional surface reconstruction techniques often simplify this problem with handcrafted priors on the input cloud: point density, noise and outlier levels, or the output surface: smoothness, topological properties, or semantics. However, adapting priors to different acquisitions often requires tedious hyperparameter tuning. 
In contrast, contemporary deep learning methods learn point cloud defects and shape properties directly from training data. In consequence, they can reconstruct accurate surfaces without manual parameter tuning. However, \ac{dsr} methods have mainly been tested on artificial data sets with a limited number of object categories. Such datasets may not be representative of the endless variety of shapes in real-world scenarios.
Moreover, \ac{dsr} methods are typically applied on uniformly sampled point clouds. These do not reflect the irregularities of real-world acquisitions such as non-uniformity, missing data from occlusions, or transparent regions. 
The ability of \ac{dsr} models to reconstruct shapes in unseen shape categories and from point clouds with unseen defects has rarely been rigorously and systematically evaluated.

In this paper, we propose a series of experiments designed to benchmark surface reconstruction algorithms from point clouds. We use a diverse range of publicly available shape datasets with objects of varying complexities and for which we know the true surface. 
To emulate real-world scenarios, we synthetically scan the objects to generate point clouds with realistic characteristics. 
Working with objects with known surfaces allows us to measure the geometric and topological quality of the reconstructed surfaces. We also consider point clouds from real acquisitions for which the true surface is typically unknown. 
By comparing novel learning-based algorithms to traditional test-of-time methods, we specifically study the impact of learned \textit{v.s.} handcrafted priors in the reconstruction process.
Throughout our study, the generalizability of these methods remains a focal point.
Our main contributions are as follows:
\begin{itemize}
    \item We provide a comprehensive literature review of surface reconstruction methods from point clouds, tracing developments over three decades up to the latest learning-based approaches. We contrast popular test-of-time traditional models with novel \ac{dsr} models. \RAPH{Additionally, we discuss their relation to the new emerging field of rendering-based surface reconstruction.}
    \item We benchmark both traditional and learning-based methods under consistent conditions. We perform several experiments to assess their geometric precision, topological consistency, and robustness using openly available shape datasets and point clouds from both synthetic and real scans.
\end{itemize}

\begin{figure*}
    \centering
    \resizebox{0.9\textwidth}{!}{%

     \newcommand\myscale{0.4}
\newlength\mylinewidth
\setlength\mylinewidth{0.4mm}

\captionsetup[sub]{labelfont=scriptsize,textfont=scriptsize,justification=centering,skip=5pt}
    \centering

\begin{tabular}{@{}c@{}c@{}c@{}}

\subfloat[Unknown Topology]{
\begin{tikzpicture}

\node[circle,scale = \myscale] at (0,-0.5) (e1) {};
\node[circle,scale = \myscale] at (4.3,1.4) (e2) {};
\node[circle,scale = \myscale] at (4.2,1.1) (g) {};

\node[circle,scale = \myscale] at (2.7,1) (i1) {};

\node[circle,fill=black,scale = \myscale] at (0.5,0) (a) {};
\node[circle,fill=black,scale = \myscale] at (1.3,1) (b) {};
\node[circle,fill=black,scale = \myscale] at (2,0) (c) {};
\node[circle,fill=black,scale = \myscale] at (2.4,0.5) (c1) {};
\node[circle,fill=black,scale = \myscale] at (3.1,0.5) (d) {};
\node[circle,fill=black,scale = \myscale] at (3.8,2) (f) {};

\begin{pgfonlayer}{background}
\draw [SURFCOLOR, line width = \mylinewidth] plot [smooth, tension=0.5] coordinates {(0,0) (a) (b) (c) (c1) (i1) (d) (f) (g)};
\draw [red, line width = \mylinewidth] plot [smooth, tension=0.5] coordinates {(0,2) (b) (d) (f) (4.3,2.5)};
\draw [red, line width = \mylinewidth] plot [smooth, tension=0.5] coordinates {(0,-0.5) (a) (c) (4.3,-0.5)};
\end{pgfonlayer}
\label{fig:ch1:topology}
\end{tikzpicture}
}
&
\subfloat[Unknown Geometry]{
\begin{tikzpicture}

\node[circle,scale = \myscale] at (0,-0.5) (e1) {};
\node[circle,scale = \myscale] at (4.3,1.4) (e2) {};

\node[circle,scale = \myscale] at (2.7,1) (i1) {};

\node[circle,fill=black,scale = \myscale] at (0.5,0) (a) {};
\node[circle,fill=black,scale = \myscale] at (1.3,1) (b) {};
\node[circle,fill=black,scale = \myscale] at (2,0) (c) {};
\node[circle,fill=black,scale = \myscale] at (2.4,0.5) (c1) {};
\node[circle,fill=black,scale = \myscale] at (3.1,0.5) (d) {};
\node[circle,fill=black,scale = \myscale] at (3.8,2) (f) {};
\node[circle,scale = \myscale] at (4.2,1.1) (g) {};

\begin{pgfonlayer}{background}
\draw [SURFCOLOR, line width = \mylinewidth] plot [smooth, tension=0.5] coordinates {(0,0) (a) (b) (c) (c1) (i1) (d) (f) (g)};
\draw [red, line width = \mylinewidth] plot [smooth, tension=0.5] coordinates {(c1) (d)};
\draw [red, line width = \mylinewidth] plot [smooth, tension=0.5] coordinates {(c1) (2.7,0) (d)};
\end{pgfonlayer}
\label{fig:ch1:geometry}
\end{tikzpicture}
}
&
\subfloat[Acquisition Defects]{
\begin{tikzpicture}

\node[circle,scale = \myscale] at (0,-0.5) (e1) {};
\node[circle,scale = \myscale] at (4.3,1.4) (e2) {};

\node[circle,scale = \myscale] at (2.7,1) (i1) {};

\node[circle,fill=black,scale = \myscale] at (0.5,0) (a) {};
\node[circle,fill=black,scale = \myscale] at (1,1) (b) {};
\node[circle,fill=black,scale = \myscale] at (2,0) (c) {};
\node[circle,fill=black,scale = \myscale] at (2.4,0.5) (c1) {};
\node[circle,fill=black,scale = \myscale] at (3.1,0.5) (d) {};
\node[circle,fill=black,scale = \myscale] at (3.8,2) (f) {};
\node[circle,scale = \myscale] at (4.2,1.1) (g) {};

\begin{pgfonlayer}{background}
\draw [SURFCOLOR, line width = \mylinewidth] plot [smooth, tension=0.5] coordinates {(0,0) (a) (1.3,1) (c) (c1) (i1) (d) (f) (g)};
\draw [red, line width = \mylinewidth] plot [smooth, tension=0.5] coordinates {(0,0) (a) (b) (c)};
\end{pgfonlayer}
\label{fig:ch1:defects}
\end{tikzpicture}
}
\end{tabular}
}

        \caption[Difficulties in surface reconstruction from point clouds]{\textbf{Difficulties in surface reconstruction from point clouds:} In each plot, we show the real surface
        \protect\tikz[baseline=-.25em] \protect\draw[-, thick, draw = SURFCOLOR, line width = 0.3mm] (0,0) -- (0.5,0);,
        point samples
        \protect\tikz[baseline=-.25em] \protect\node[circle, thick, draw = none, fill = black, scale = 0.5] {};,
        and possible reconstructions 
        \protect\tikz[baseline=-.25em] \protect\draw[-, thick, draw = red, line width = 0.3mm] (0,0) -- (0.5,0);. 
        The correct topology and geometry of the real surface are not known from the point samples (\subref{fig:ch1:topology},\subref{fig:ch1:geometry}). The point samples may also include acquisition defects such as noise (\subref{fig:ch1:defects}). The goal of any surface reconstruction algorithm is finding a good approximation
        of the real surface, in terms of its geometry and topology.
        Learning-based surface reconstruction can learn shape patterns or sampling errors such as the one exemplified here, and use the learned knowledge during reconstruction for a better approximation.}
    \label{fig:ch2:3_problems}

\end{figure*}

\section{Related work}
In this section, we provide an overview of the existing surveys and benchmark studies of \ac{dsr} methods.

\subsection{Surveys}

Although several works survey the broad field of surface reconstruction from point clouds \cite{cazals2006delaunay,Berger_survey,you2020survey,bolle1991survey, huang2022survey,cheng2008survey}, the majority predate the rise of learning-based reconstruction. 
Surface reconstruction methods are often grouped as interpolation-based \cite{cazals2006delaunay} or approximation-based techniques~\cite{Berger_survey}. Interpolation methods ``connect'' points of the input point cloud, usually by linear interpolation between pairs of points. Approximation methods use smooth functions to represent the point cloud globally or locally \cite{botsch2010polygon}. 
In our review, we consider both interpolating and approximating methods and focus on novel ideas in learning-based surface reconstruction.
While many reconstruction methods can be distinguished by their surface priors \cite{Berger_survey, huang2022survey}, we argue that several successful approaches combine different priors, complicating classification solely based on priors. We thus organize the methods into two groups: surface-based and volume-based approaches. This breakdown mirrors the primary mathematical surface representations: parametric \cite{farin2014curves} and implicit \cite{gomes2009implicit}.
To the best of our knowledge, only two surveys include recent learning-based methods \cite{huang2022survey, farshian2023deepsurvey}. The work of You~\etal \cite{you2020survey} predates important developments for learning-based methods, such as the incorporation of local information \cite{Peng2020,deepLS,lig,points2surf,dgnn,rakotosaona2021dse,boulch2022poco}. 




\subsection{Benchmarks}

Historically, very few benchmarks for surface reconstruction from point clouds have been proposed. Numerous methods use bespoke datasets to evaluate their approach, typically by uniformly sampling point clouds from the ground truth surface from existing shape collections \cite{Park2019,Peng2020,deepLS,lig,rakotosaona2021dse,boulch2022poco}. However, the characteristics of these sampled point clouds often differ between studies, complicating the comparison of their results. Moreover, these point clouds typically lack common defects seen in real acquisitions, such as missing data from occlusion.

Two notable exceptions are the benchmarks of Berger \etal \cite{Berger_benchmark} and Huang \etal \cite{huang2022survey}. Both teams develop synthetic range scanning procedures to produce point clouds with realistic artifacts such as noise, non-uniformity, and misaligned scans. They also use shapes with non-trivial topology and details of various sizes. While providing interesting results, Berger \etal's benchmark predates learning-based surface reconstruction and only considers traditional approximation techniques.
Huang \etal's benchmark considers learning-based methods, but does not specifically study their generalization capabilities. 

\begin{figure*}
\centering
    \resizebox{0.9\textwidth}{!}{%
\captionsetup[sub]{labelfont=scriptsize,textfont=scriptsize,justification=centering}
    \centering
    \newcommand{\mywidth}{0.29\textwidth}
    \begin{tabular}{ccc}
         \subfloat[High Quality Mesh]{    \includegraphics[width=\mywidth]{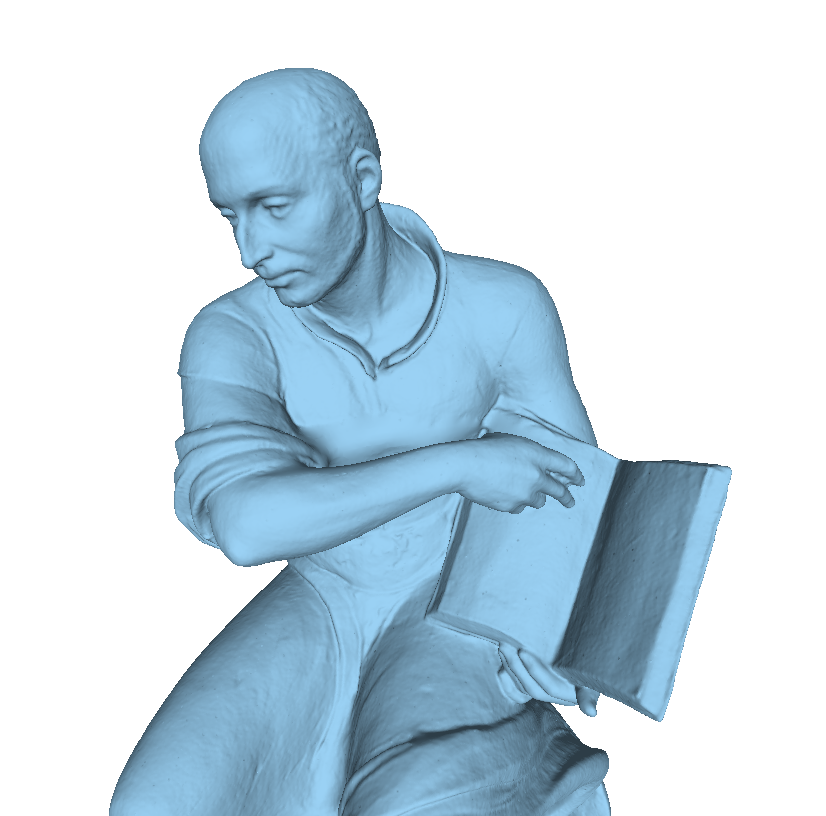}\label{fig:ign:mesh}}
         &  \subfloat[Real MVS]{    \includegraphics[width=\mywidth]{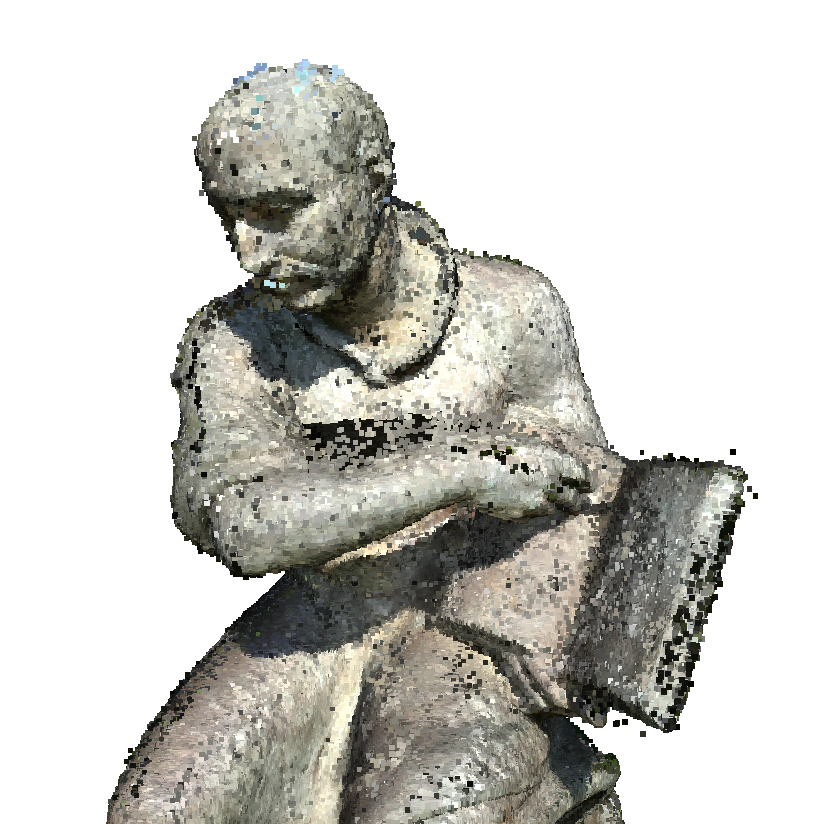}\label{fig:ign:mvs}}
         &  \subfloat[Real range scan]{    \includegraphics[width=\mywidth]{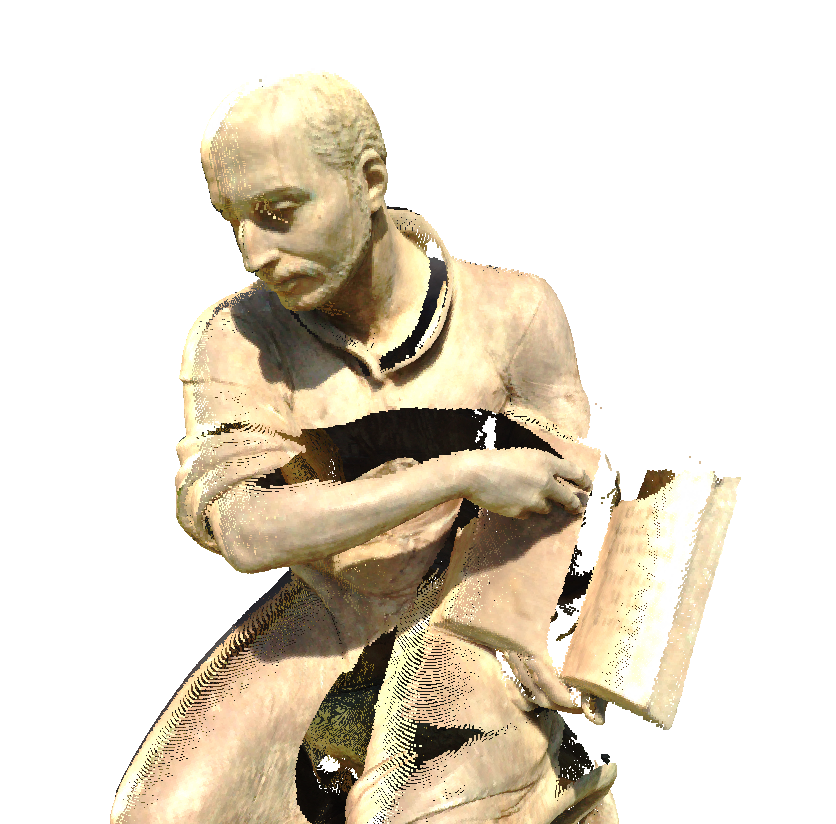}\label{fig:ign:lidar}}\\
         \subfloat[Uniform sampling]{    \includegraphics[width=\mywidth]{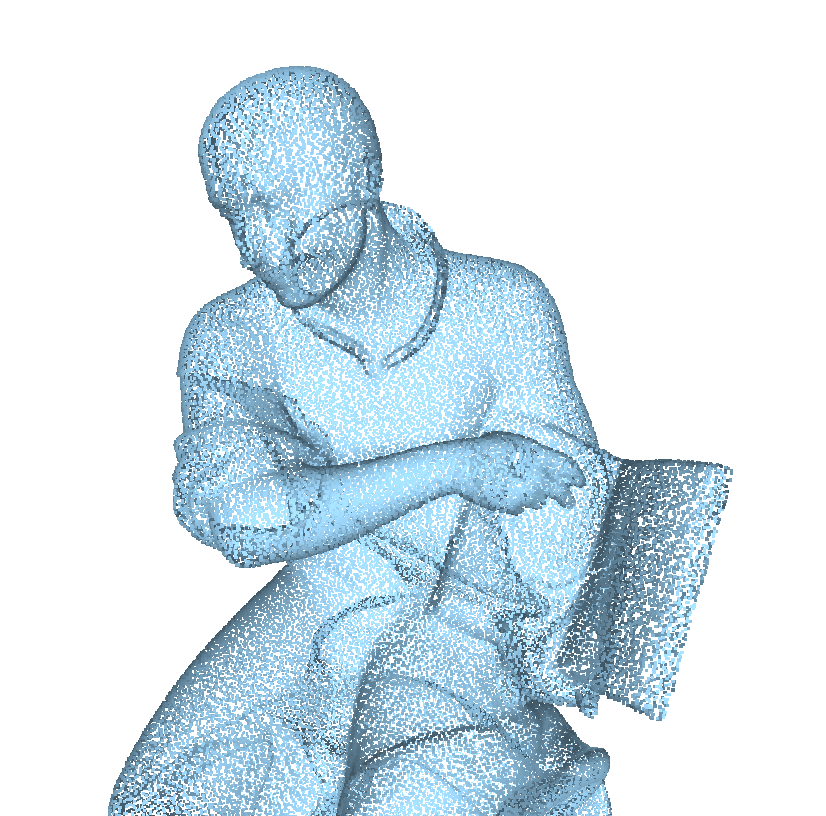}\label{fig:ign:uniform}}      
         & \subfloat[Synthetic MVS]{    \includegraphics[width=\mywidth]{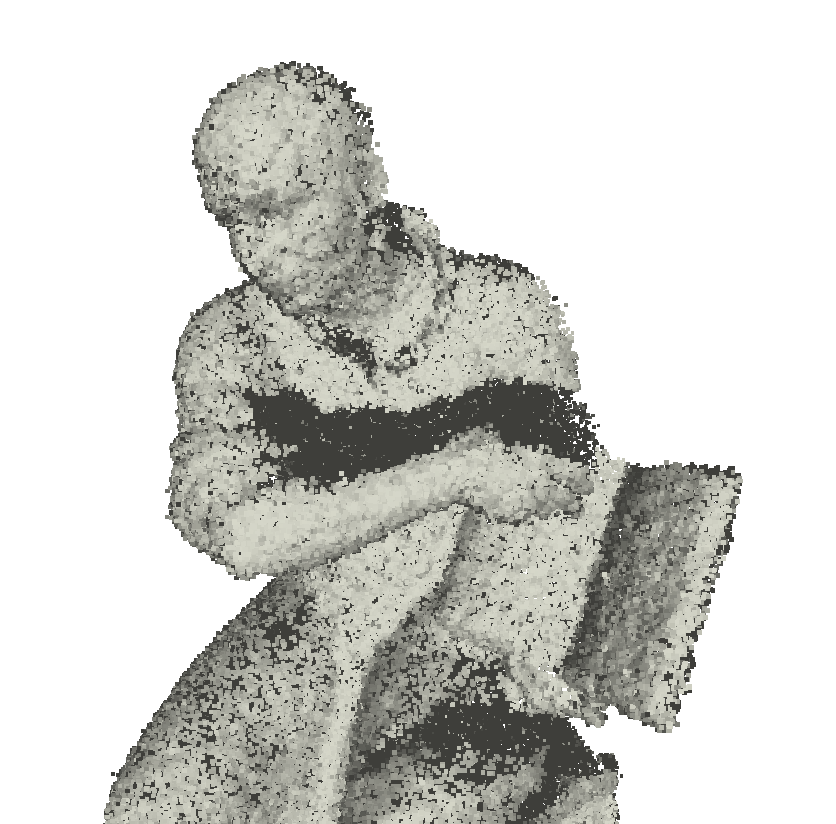}\label{fig:ign:synmvs}}         
         & \subfloat[Synthetic range scan]{    \includegraphics[width=\mywidth]{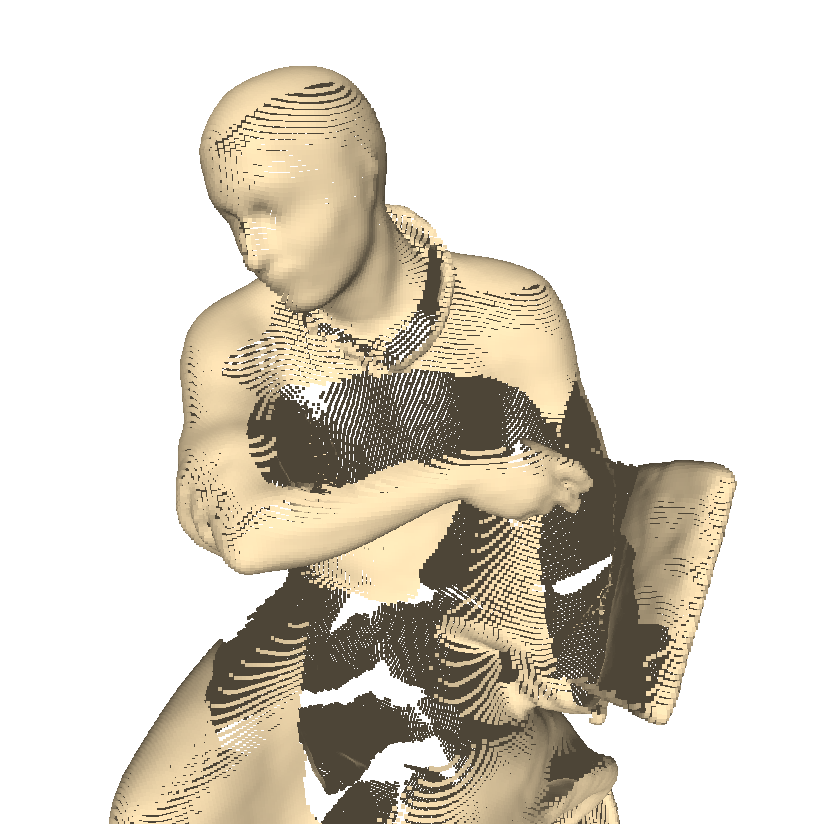}\label{fig:ign:synlidar}}         
    \end{tabular}}
\caption[Synthetic and real point clouds]{\textbf{Synthetic and real point clouds:} 
Real world point cloud acquisitions (\subref{fig:ign:mvs},\subref{fig:ign:lidar}) have defects such as missing data from occlusion.
However, surface reconstruction methods are often tested on point clouds that are produced by directly and uniformly sampling a ground truth surface (\subref{fig:ign:uniform}). While this sampling strategy allows to add artificial noise, it cannot realistically model missing data from occlusion.
Instead, we test methods on synthetic MVS (\subref{fig:ign:synmvs}) and synthetic range scans (\subref{fig:ign:synlidar}) which allows us to reproduce realistic sampling defects.}
    \label{fig:ch2:ignatius}
\end{figure*}

In our benchmark, we use the synthetic range scanning procedure of Berger~\etal and their test shapes as they provide a realistic challenge for both learning-based and traditional algorithms. We also implement our own synthetic scanning procedure to produce \ac{mvs}-like point clouds. We use synthetic scanning on existing large shape datasets to create training datasets with true surfaces and point clouds with realistic characteristics (see \figref{fig:ch2:ignatius}). 
Furthermore, we specifically study the generalization capabilities of methods by using train and test sets with different characteristics in terms of shape categories and acquisition defects.

\vspace{0.1cm}\noindent\textbf{MVS benchmarks.} The generation of point clouds from 2D information such as overlapping images is an integral problem of surface reconstruction. There exists a variety of benchmarks using data captured in a laboratory environment \cite{middlebury,dtu} or in the wild \cite{Strecha08,Schops2017,tanksandtemples}. These benchmarks often use low-quality image acquisitions as input, and use a higher-quality acquisition, \eg from LiDAR scans, to produce precise and dense point clouds that serve as reference.
{However, even with advanced sensors, producing a complete and precise point cloud remains difficult. As a result, the evaluation of the reconstructed surfaces can be flawed. A common workaround is to restrict the evaluation to specific ``trustworthy'' areas \cite{dtu, Schops2017}. 
However, in contrast to true and complete surfaces, this approach prevents the computation of topological metrics such as component count, or differential metrics such as surface normals. Furthermore, many learning-based methods require a large dataset of closed reference surfaces and not just a small set of precise point clouds.}

\section{Representations, properties, and reconstruction of surfaces}

\label{ch2:definitionAndProperties}

In this section, we first define the notion of surface and its mathematical and digital representations. 
We then discuss the most critical properties of reconstructed surfaces.
Finally, we propose a grouping of surface reconstruction algorithms featured in our survey based on the mathematical representations of surfaces: parametric or implicit.

Throughout this article, we denote by $\cP$ the input point cloud, $\cSr$ the reconstructed surface, and $\cS$ the real surface from which $\cP$ is sampled.



\subsection{Representations}

A surface is defined as an orientable, continuous 2-manifold within \( \bR^3 \), which may or may not have boundaries \cite{botsch2010polygon,hoppe1992surface,o2006elementary}. These properties hold significance for surface visualization and processing, which we will elaborate on in subsequent sections. From a mathematical perspective, surface representations fall primarily into two categories: \emph{parametric} and \emph{implicit}.


\vspace{0.1cm}\noindent\textbf{Parametric surfaces.} These surfaces are represented by a continuous and bijective function $\textbf{f}: \Omega \mapsto \cS$ that maps a parameter domain $\Omega \in \bR^2$ (\eg $[-1,1]^2$) to a 3D surface $\cS = \textbf{f}(\Omega) \in \bRt$. 
However, it may be impractical to devise a single function to parameterize complex surfaces.
Therefore, the parameter domain $\Omega$ is usually divided into smaller sub-regions, each equipped with its own function \cite{botsch2010polygon}.
Typically, $\Omega$ is partitioned into triangles.
A set of triangles approximating $\cS$ can be efficiently stored and processed as a triangle surface mesh $\cM = (\cV,\cE,\cF)$, where $\cV$ denotes the vertices of the triangles, $\cE$ their edges and $\cF$ their facets.

\vspace{0.1cm}\noindent\textbf{Implicit surfaces.} These surfaces are defined by the level-set $c$ of a scalar-valued function $F: \bRt \mapsto \bR$:
\begin{align}
    \cS_c = \{ \textbf{x} \in \bRt \mid F(\textbf{x})=c\}~.
\label{eq:implicit}
\end{align}
The most common choice for $F$ are \ac{sdf} or \ac{of}.
An \ac{sdf} gives the distance from a \ac{3d} point $\textbf{x}$ to the closest part of the surface. Points within the interior of the surface are assigned negative values, while those outside are given positive values. An \ac{of} typically has a value of $1$ inside the surface and $0$ outside. The $c$-level-set of $F$ represent the surface \( \cS \) with $c=0$ for \ac{sdf}s and $c=0.5$ for \ac{of}s. 
Similar to the parametric case, the domains of implicit functions are often split into sub-regions, such as voxels~\cite{Peng2020}, OcTree-nodes~\cite{screened_poisson}, or tetrahedra~\cite{Vu2012}. 
{In practice, once computed, 
implicit surfaces are often converted into a parametric mesh, which is easier to handle for downstream applications.}


\begin{figure*}[ht!]
	\definetrim{mytrim}{55 90 70 280}
	\definetrim{mytrim2}{190 80 70 250}

    \centering
    \begin{tabular}{c@{}c@{}c@{}c}
     \subfloat[Non-watertight]{  
        \includegraphics[width=.24\linewidth,mytrim]{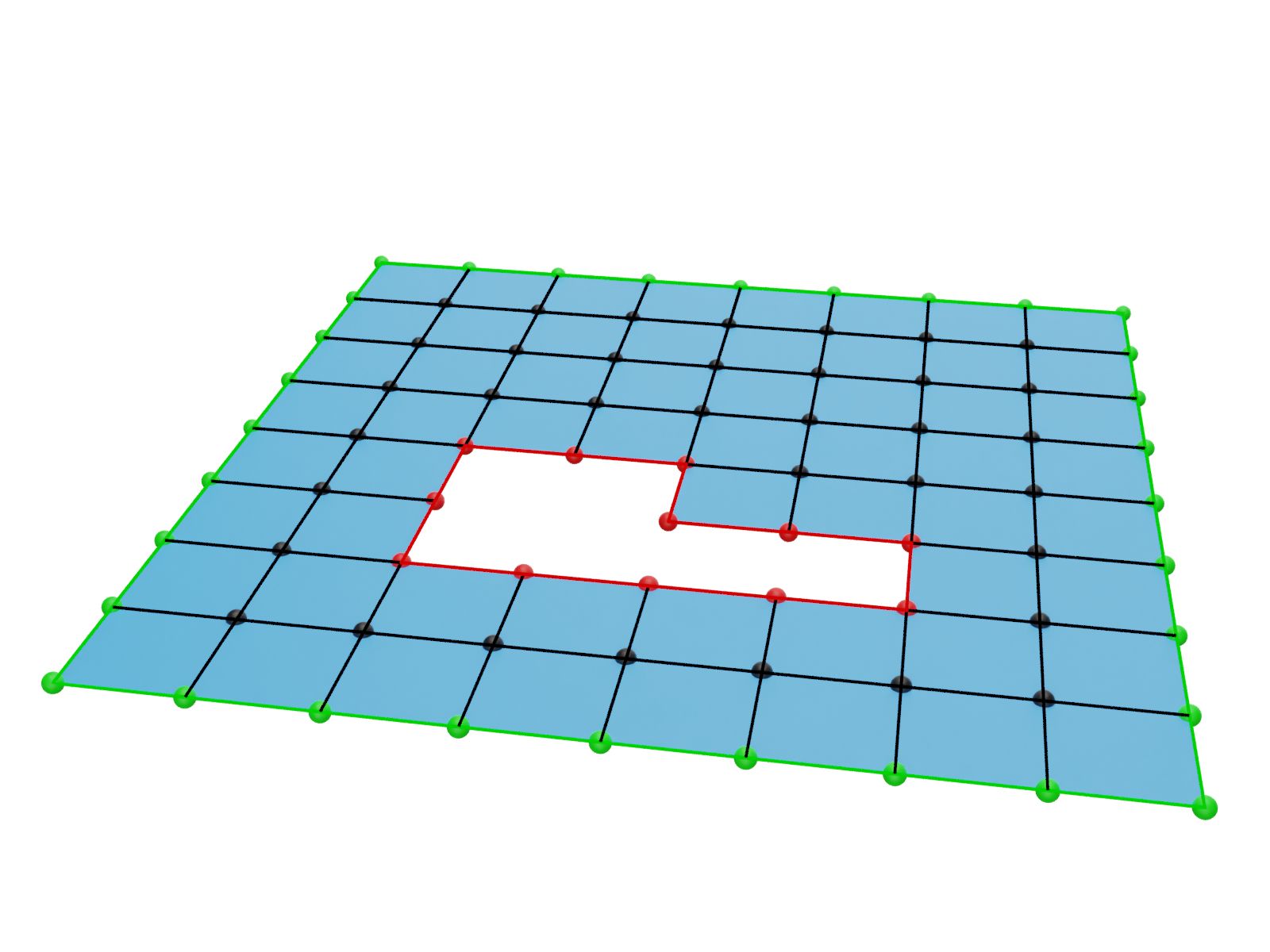}\label{properties:a}
    }
    &
     \subfloat[Non-manifold]{  
        \includegraphics[width=.24\linewidth,mytrim]{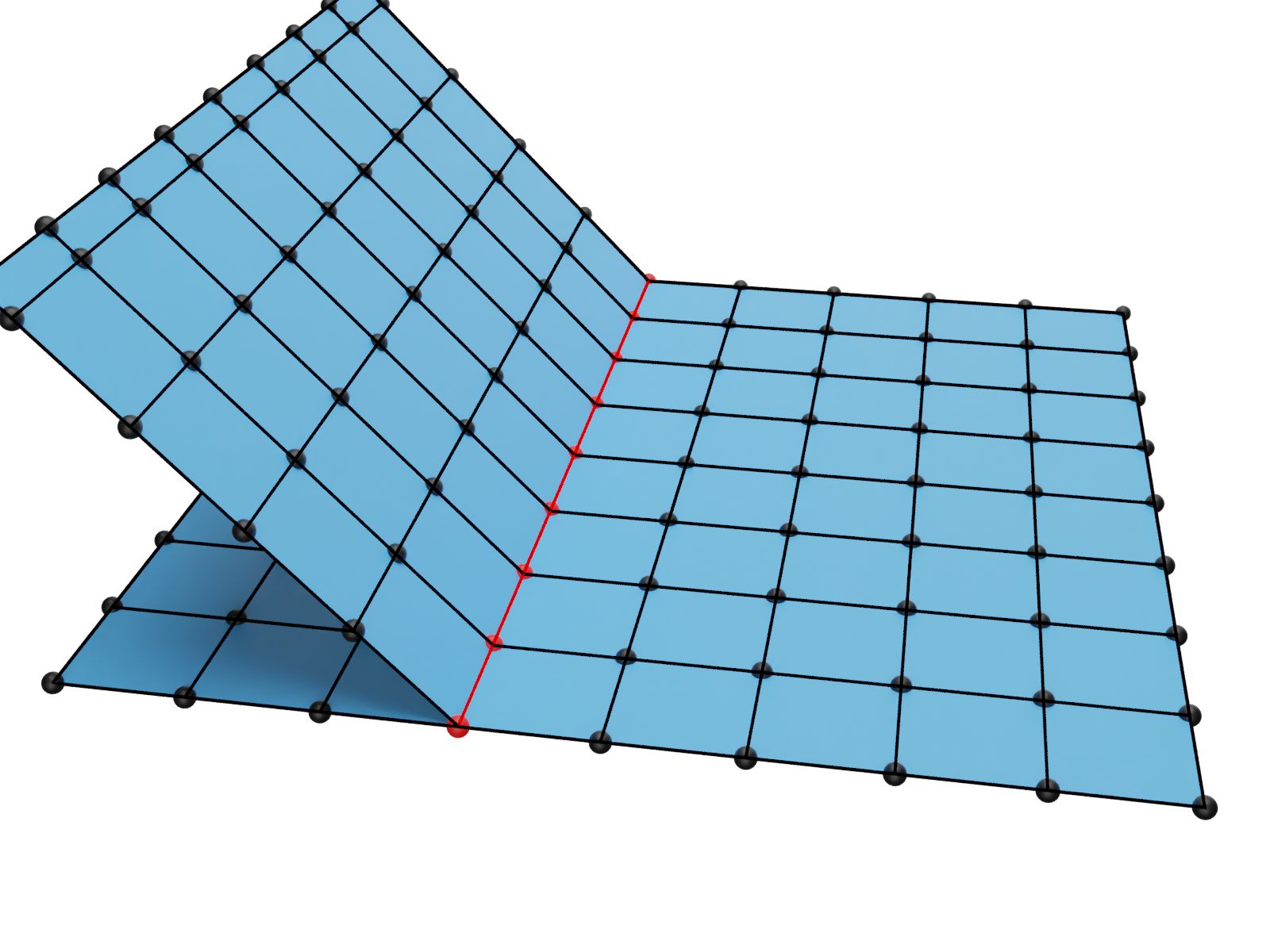}\label{properties:b}
        }
    &
     \subfloat[Intersecting]{  
        \includegraphics[width=.24\linewidth,mytrim]{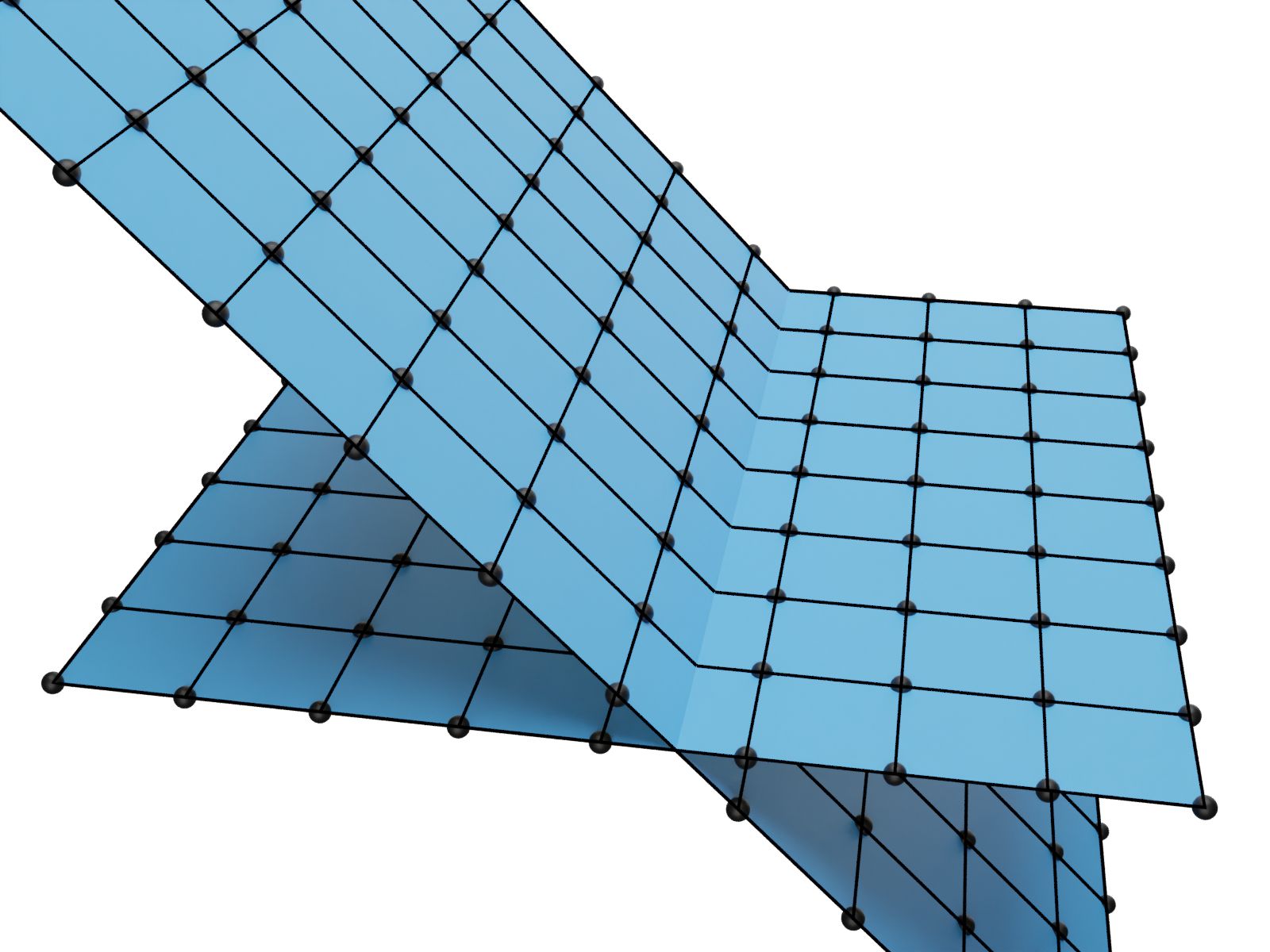}\label{properties:c}
    }
    &
     \subfloat[Non-orientable]{  
        \includegraphics[width=.24\linewidth,mytrim2]{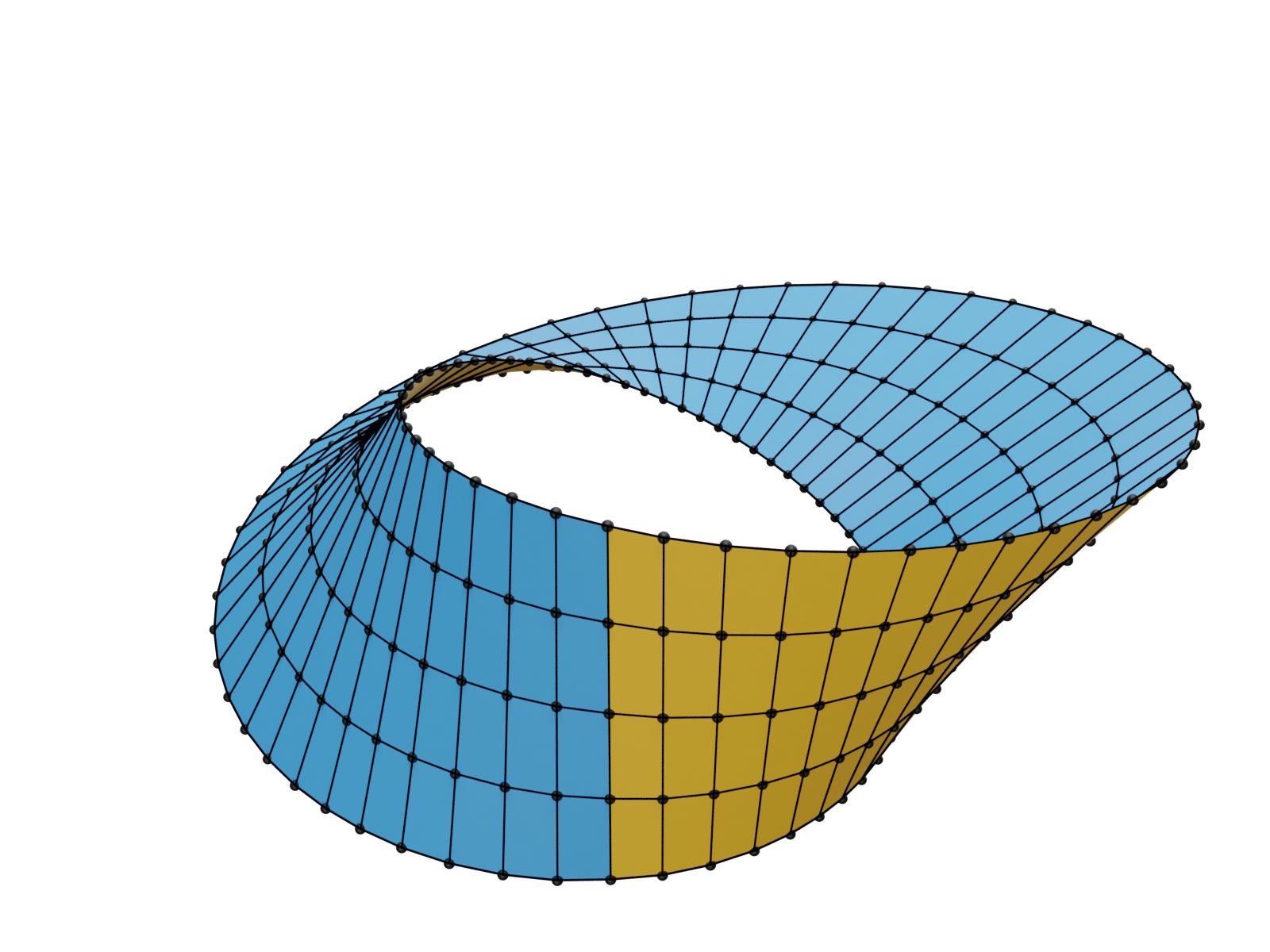}\label{properties:d}
    }
    \end{tabular}
    \caption{{\bf Properties of surface meshes:} In (\subref{properties:a}), we show a surface mesh that is non-watertight due to the hole in the center marked with red vertices and edges. The green vertices and edges mark the intersection with the domain boundary. 
        In (\subref{properties:b}), we show a surface mesh with non-manifold edges and non-manifold vertices marked in red. 
        In (\subref{properties:c}), we show a non-manifold and intersecting surface mesh. 
        Here, the non-manifoldness is harder to detect, because it does not happen along edges of the mesh.
        In (\subref{properties:d}), we show a non-orientable surface.
    }
    \label{fig:prop}
\end{figure*}

\subsection{Properties}
\label{properties}

We want the reconstructed surface $\cSr$ to be close in terms of geometry and topology to the real surface $\cS$ from which the point cloud $\cP$ is sampled.
Real surfaces exhibit certain intrinsic properties, such as watertightness, manifoldness, and orientability, which are crucial for a variety of downstream applications. Therefore, it is essential that the reconstructed surface $\cSr$ also possess these properties. In this section, we define these properties in the context of mesh representations. For a visual interpretation of these concepts, the reader is directed to \figref{fig:prop}.
\begin{itemize}
    \item \textbf{Watertightness:}
    A geometric surface is considered watertight, or closed, when it is boundary-free. This implies that it forms a continuous and uninterrupted surface that encloses a finite volume without gaps or openings. For example, a sphere is a closed surface as one can move anywhere along its surface without ever encountering any boundaries. 
     A mesh $\cM$ is boundary-free (or closed) if no edge is incident to only one facet: every edge should be shared between two facets to avoid gaps. 
      {In practice, surfaces reconstructed from real-world scenes will necessarily have boundaries due to the finite nature of the scanning coverage. 
      Thus, we define a surface as watertight if it is boundary-free, except for any intersections with its domain boundary. \figref{properties:a} illustrates a non-watertight surface.}
    
    \item \textbf{Manifoldness:} 
    {Real and geometric surfaces are 2-manifold if and only if any of its point has a local neighborhood that is homeomorphic to an open subset of the Euclidean plane.
    In other words, every small enough region of the surface, irrespective of its curvature, can be continuously deformed into a 2D plane. A mesh $\cM$ is manifold if it has the following attributes:
    \begin{itemize}
    \item \textit{Edge-manifold:} For each edge in $\cE$, the set of facets that share this edge form a topological half-disk. 
    This means that no edge can be incident to more than two facets. See \figref{properties:b} for an illustration of a mesh violating the edge-manifoldness property.
    \item \textit{Vertex-manifold:} For each vertex in $\cV$, the set of facets that share this vertex forms a topological half-disk. This implies that facets that share a common vertex are organized in an open or closed fan.
    In other words, each vertex is part of at least two edges and each edge is part of two facets.
    \item \textit{Intersection-free:} 
    All pairs of facets not sharing an edge or vertex do not intersect. In \figref{properties:c}, we represent a mesh that self-intersect outside of an edge.
    
    \end{itemize}}
  \item \textbf{Orientability:}
    {A mesh $\cM$ is orientable if and only if one can consistently assign a direction (or ``side'') to its facets such that adjacent facets always share the same orientation. This allows us to distinguish between ``inside'' and ``outside'' of the mesh.
    If the order of the vertices dictates the direction of the outward-facing normal with the right-hand rule, this has implications on the vertex ordering as well. Specifically, when two facets share a common edge, the ordering of the shared vertices should be reversed for the two facets. See \figref{properties:d} for an illustration of a non-orientable surface.}
    
\end{itemize}

{The watertightness property is crucial for physical simulations such as fluid dynamics.
Both manifoldness and orientability are prerequisites for certain mesh data structures, such as the widely used half-edge structure \cite{kettner1999halfedge,mantyla1987halfedge}.
These structures ensure consistent and predictable behavior during geometric transformations, topological modifications, and various mesh optimization and query operations, and many mesh processing algorithms require a manifold mesh.
Lastly, intersection-free and orientable surfaces lead to a well-defined notion of inside and outside, which is important for mesh visualization and various geometric operations and metrics.
}




\subsection{Reconstruction}

{Surface reconstruction from point clouds consists in creating a continuous surface approximating the original surface from which the 3D points were sampled.
In our survey, we group the methods for surface reconstruction from point clouds into two groups: surface-based and volume-based.}

\vspace{0.1cm}\noindent{\textbf{Surface-based methods.} These approaches 
focus on finding one or several parametric surfaces $\cSr$ approximating the input point cloud $\cP$, either in the form of triangles or \ac{2d} patches, or by deforming parameterized enclosing envelops such as meshed spheres. 
The main limitation of surface-based methods with a single function $\textbf{f}$ is that the topology of the parameter domain $\Omega$ must be equivalent to the topology of the true surface $S$, which is usually unknown. 
Conversely, the main challenge for surface-based methods that define $\cSr$ with multiple functions for different regions is to guarantee consistent transitions between subsurfaces. In particular, it is difficult to keep the final surface intersection-free, manifold, and watertight.}

\vspace{0.1cm}\noindent{\textbf{Volume-based methods.} These methods segment a subset of $\bRt$ into interior (inside) and exterior (outside) subspaces. The surface is implicitly defined as the interface between the two subspaces.
Most, but not all algorithms in this class formulate the problem as finding an implicit function indicating where a point in space is full (inside) or empty (outside).
Surfaces reconstructed with volume-based methods are guaranteed to be watertight and intersection-free, but not necessarily manifold \cite{cazals2006delaunay}.} 

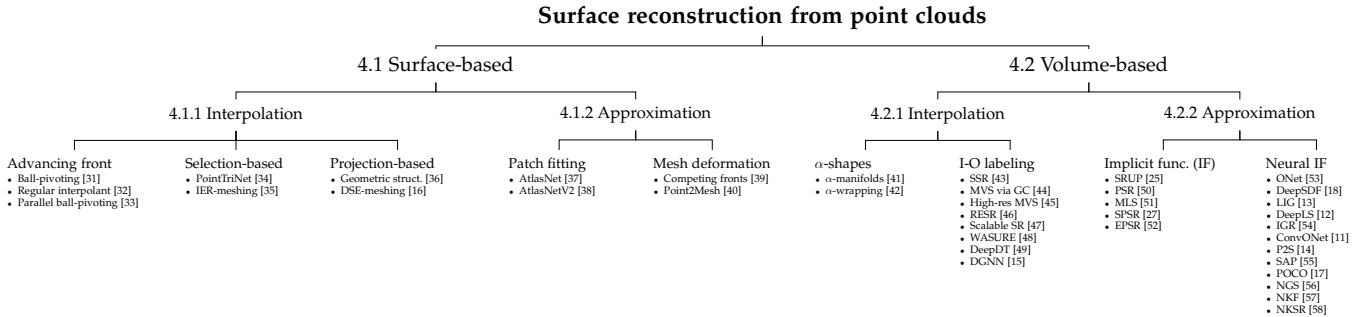
\begin{figure*}
\newcommand{\fsvolsurf}{\LARGE}
\newcommand{\fsapproxinter}{\Large}
\newcommand{\fsgroups}{\large}
\newcommand{\fsmethods}{\small}
\forestset{
  qtree/.style={
    for tree={
      parent anchor=south,
      anchor=north,
      child anchor=north,
      align=center,
      inner sep=4pt,
      l=1.5cm,
      s sep=1cm, 
      level distance=200pt,
      edge={-, >=latex, shorten >=2pt, shorten <=2pt},
      edge path={
        \noexpand\path[\forestoption{edge}]
        (!u.south) -- +(0,-10pt) -| (.child anchor)
        \forestoption{edge label};
      },
      rounded corners,
    },
    for siblings={
      align=north,
      baseline=(current bounding box.north),
    },
  },
}
          
\begin{adjustbox}{max width=\textwidth}
        \begin{forest}, baseline, qtree
[\huge{\textbf{Surface reconstruction from point clouds}}
    [\fsvolsurf{\ref{survey:surface-based} Surface-based}
        [\fsapproxinter{\ref{survey:surface-based-interpolation} Interpolation}
            [\parbox{4cm}{
            \fsgroups{Advancing front}     
                \begin{itemize}[leftmargin=*,before=\fsmethods]
                \item Ball-pivoting~\cite{bernardini1999ball}
                \item Regular interpolant~\cite{petitjean2001regular}
                \item Parallel ball-pivoting~\cite{digne2014bpa_parallel}
                \end{itemize}}
              ]
            [\parbox{3cm}{
            \fsgroups{Selection-based}    
                \begin{itemize}[leftmargin=*,before=\fsmethods]
                \item PointTriNet~\cite{PointTriNet}
                \item IER-meshing~\cite{liu2020ier}
               \end{itemize}}
              ]
            [\parbox{4cm}{
            \fsgroups{Projection-based}     
                \begin{itemize}[leftmargin=*,before=\fsmethods]
                \item Geometric struct.~\cite{boissonat1984geometric}
                \item DSE-meshing~\cite{rakotosaona2021dse}
                \end{itemize}}
            ]
        ]
        [\fsapproxinter{\ref{survey:surface-based-approximation} Approximation}
                [\parbox{3cm}{
                \fsgroups{Patch fitting}
                \begin{itemize}[leftmargin=*,before=\fsmethods]
                \item AtlasNet~\cite{Groueix2018}
                \item AtlasNetV2~\cite{deprelle2019atlasnetv2}
                \end{itemize}}
            ]
                [\parbox{3.5cm}{
                \fsgroups{Mesh deformation}
                \begin{itemize}[leftmargin=*,before=\fsmethods]
                \item Competing fronts~\cite{sharf2006competing}
                \item \Acl{p2m}~\cite{point2mesh}
                \end{itemize}}
            ]
        ]
    ]
    [\fsvolsurf{\ref{survey:volume-based} Volume-based},    
        [\fsapproxinter{\ref{survey:volume-based-interpolation} Interpolation}
                                    [\parbox{3cm}{
            \fsgroups{$\alpha$-shapes}     
                \begin{itemize}[leftmargin=*,before=\fsmethods]
                \item $\alpha$-manifolds~\cite{bernardini1997alphashapes}
                \item $\alpha$-wrapping~\cite{portaneri2022alpha}
                \end{itemize}}
            ]
                                                [\parbox{3cm}{
            \fsgroups{I-O labeling}     
                \begin{itemize}[leftmargin=*,before=\fsmethods]
                \item SSR~\cite{kolluri2004ssr}
                \item MVS via GC~\cite{sinha2007multi}
                \item High-res MVS~\cite{hiep2009towards}
                \item RESR~\cite{Labatut2009a}
                \item Scalable SR~\cite{mostegel2017scalable}
                \item WASURE~\cite{wasure}
                \item DeepDT~\cite{luo2021deepdt}
                \item DGNN~\cite{dgnn}
                \end{itemize}}
            ]
        ]
        [\fsapproxinter{\ref{survey:volume-based-approximation} Approximation}
                                                [\parbox{3.5cm}{
            \fsgroups{Implicit func. (IF)}     
                \begin{itemize}[leftmargin=*,before=\fsmethods]
                \item SRUP~\cite{hoppe1992surface}
                \item PSR~\cite{kazhdan_poisson_2006}
                \item MLS~\cite{kolluri2008provably}
                \item SPSR~\cite{screened_poisson}
                \item EPSR~\cite{Kazhdan2020envelope}
                \end{itemize}}
            ]
                                                [\parbox{3cm}{
            \fsgroups{Neural IF}     
                \begin{itemize}[leftmargin=*,before=\fsmethods]
                \item \acs{onet}~\cite{Mescheder2019}
                \item DeepSDF~\cite{Park2019}
                \item \acs{lig}~\cite{lig}
                \item DeepLS~\cite{deepLS}
                \item \acs{igr}~\cite{Gropp2020}
                \item \acs{conet}~\cite{Peng2020}
                \item \acs{p2s}~\cite{points2surf}
                \item \acs{sap}~\cite{Peng2021SAP}
                \item \acs{poco}~\cite{boulch2022poco}
                \item \acs{ngs}~\cite{huang2022neuralgalerkin}
                \item \acs{nkf}~\cite{williams2022nkf}
                \item \acs{nksr}~\cite{huang2023nksr}
                \end{itemize}}
            ]
        ]
    ]
]
  \end{forest}
  \end{adjustbox}
\caption{\textbf{Survey overview:}
{Surface reconstruction algorithms can be categorized based on their primary representation approach: surface-based or volume-based. Within each category, methods are further classified as either interpolation-based or approximation-based. Subsequently, specific subgroups and various individual methods are identified under each classification, providing a comprehensive framework for understanding the landscape of surface reconstruction techniques.}}
\label{fig:survey_tree}
\end{figure*}

\vspace{0.1cm}\noindent\textbf{Mesh extraction.} Surface-based methods typically directly yield a mesh, \eg by triangulating $\Omega$. In contrast, volume-based methods usually require an additional processing step.
If the implicit field is discretized with tetrahedra using \ac{3dt}, a mesh can then be built from all triangles that are adjacent to one inside-tetrahedra and one outside-tetrahedra.
Another option is the algorithm of Boissonnat and Oudot \cite{boissonat2005provably} that iteratively samples the implicit function $F$ by drawing lines traversing the volume and building a triangle mesh from their intersection with the underlying surface~$S$.
One of the most popular methods for extracting a mesh from an implicit field is Marching Cubes \cite{marching_cubes}, which discretizes the implicit function into voxels, constructs triangles inside each voxel that have at least one inside and one outside vertex, and extracts a triangulation as the union of all triangles. 
Recently, mesh extraction has also been addressed with deep learning methods \cite{maruani2023voromesh,chen2022neuraldualcontouring,chen2021neuralmarchingcubes, neuralMeshing}. Neural meshing \cite{neuralMeshing} specifically addresses the case where an implicit function is encoded in the weights of a neural network and aims to extract meshes with fewer triangles compared to Marching Cubes. A novel interesting approach in the direction of mesh extraction is the use of Voronoi diagrams to extract concise polygon meshes from an implicit function \cite{maruani2023voromesh}.

\vspace{0.1cm}\noindent\textbf{Robustness.} When the point cloud $\cP$ is sampled densely and without noise, both surface-based and volume-based methods can provide theoretical guarantees about the topology and geometry of the reconstructed surface \cite{cazals2006delaunay}. However, in this paper, we focus on the robustness of various methods to defect-laden input point clouds from \ac{3d} scanning.

\vspace{0.1cm}\noindent\textbf{Input.} \RAPH{The characteristics of the input point cloud mainly depend on the sensor used for their acquisition.}

\RAPH{
LiDAR point clouds typically offer high precision and accuracy without a significant amount of noise or outliers due to the robustness of the active sensor to challenging conditions. However, LiDAR point clouds commonly suffer from occlusion due to the sensor's limited field of view.}

\RAPH{%
Image-based point clouds are produced using a technique called \acf{mvs}. Their quality heavily depends on the quality and quantity of input images. Pixels from overlapping images are matched which allows to compute their 3D location relative to the camera position. \ac{mvs} can produce dense point clouds, especially in areas with abundant texture and features. However, poor lighting conditions or lack of texture can result in less accurate point clouds, and variations in image scales to strongly varying point densities \cite{Schops2017}.}
\section{Survey}
\label{sec:survey}

In this section, we review important surface- and volume-based surface reconstruction methods and discuss their robustness against different types of point cloud defects. We also explore the numerous links between learning-based and traditional methods. \RAPH{Finally, we discuss recently emerging rendering-based methods for surface reconstruction.}

\subsection{Surface-based reconstruction}
\label{survey:surface-based}

Surface-based approaches directly produce a surface, typically as a mesh, by connecting some of the points of $\cP$ or deforming a reference mesh.

\subsubsection{Surface-based interpolation}
\label{survey:surface-based-interpolation}

{Most traditional surface-based approaches linearly interpolate between all or a selection of 3D points of $\cP$. A critical aspect lies in how these points are connected into triangles, leading to a variety of algorithms.}

\vspace{0.1cm}\noindent\textbf{Advancing-front.}
{Computing the \ac{3dt} of $\cP$ can be done by forming triangles for all triplets of points respecting the  \emph{empty ball property}, \ie such that no other points lie within their circumsphere, which is the smallest sphere passing through three non-co-linear points. The Ball-pivoting algorithm \cite{bernardini1999ball} employs a greedy strategy to find such triplets and consists of several steps. First, it selects a seed triplet of points with the empty ball property such that their circumsphere has a radius close to a defined value related to the density of $\cP$. This triplet defines the first triangle of the mesh, and its circumsphere forms a ball used for the rest of the algorithm.
This ball is then pivoted around each edge of the triangle until it touches a new point, forming a new triplet, and hence a new triangle. Once all possible edges have been processed, the algorithm starts over with a new seed triangle until all points of $\cP$ have been considered.}

\begin{figure*}[t]
	\definetrim{mytrim}{120 0 50 0}

\captionsetup[sub]{labelfont=scriptsize,textfont=scriptsize,justification=centering}
    \centering
    \newcommand{\mywidth}{0.25\textwidth}
    \begin{tabular}{@{}c@{}c@{}c@{}c@{}}
         \subfloat[Surface-based interpolation]{\includegraphics[width=\mywidth,mytrim]{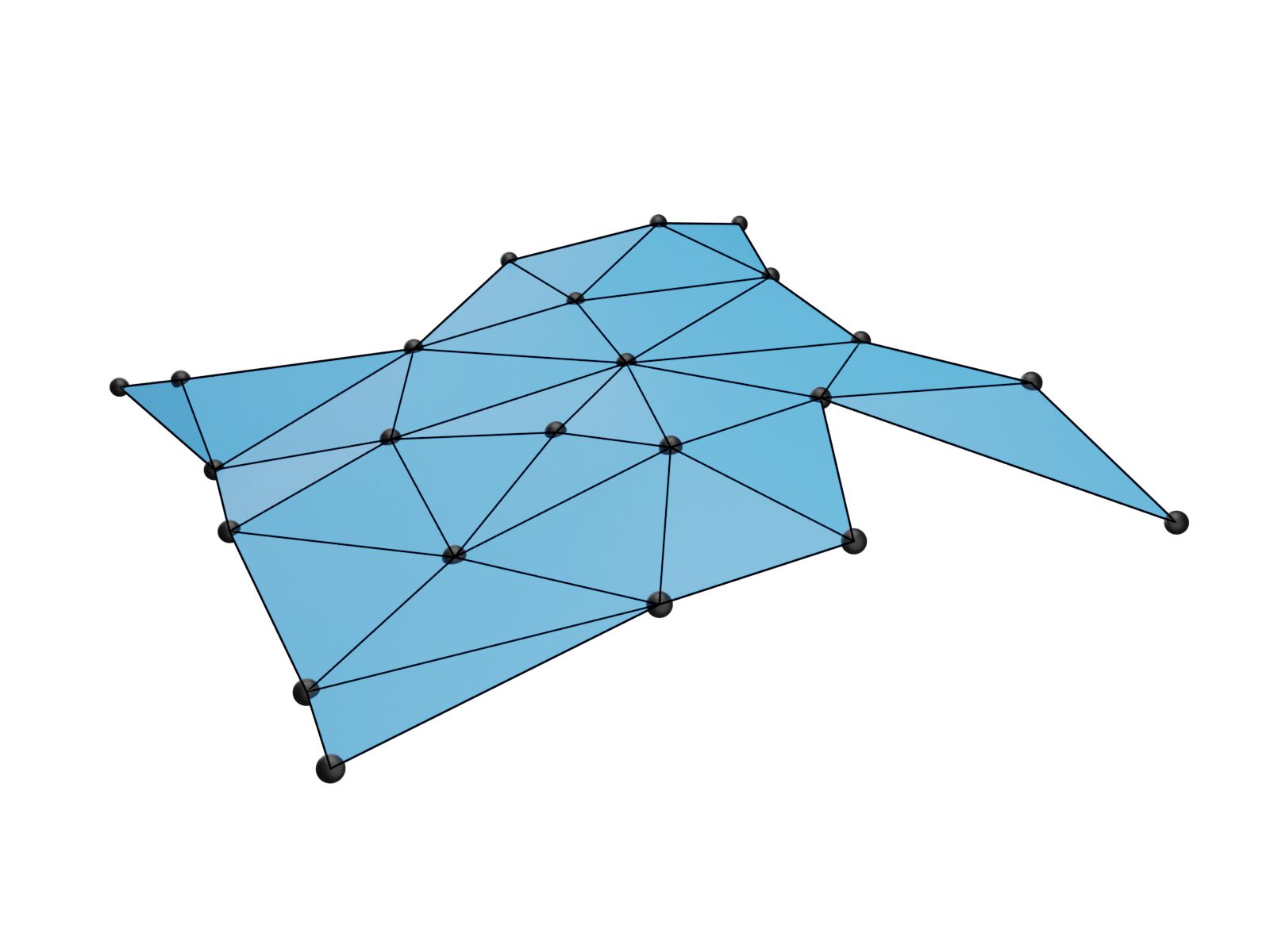}\label{fig:survey:inter_surf}}
         &
         \subfloat[Surface-based approximation]{\includegraphics[width=\mywidth,mytrim]{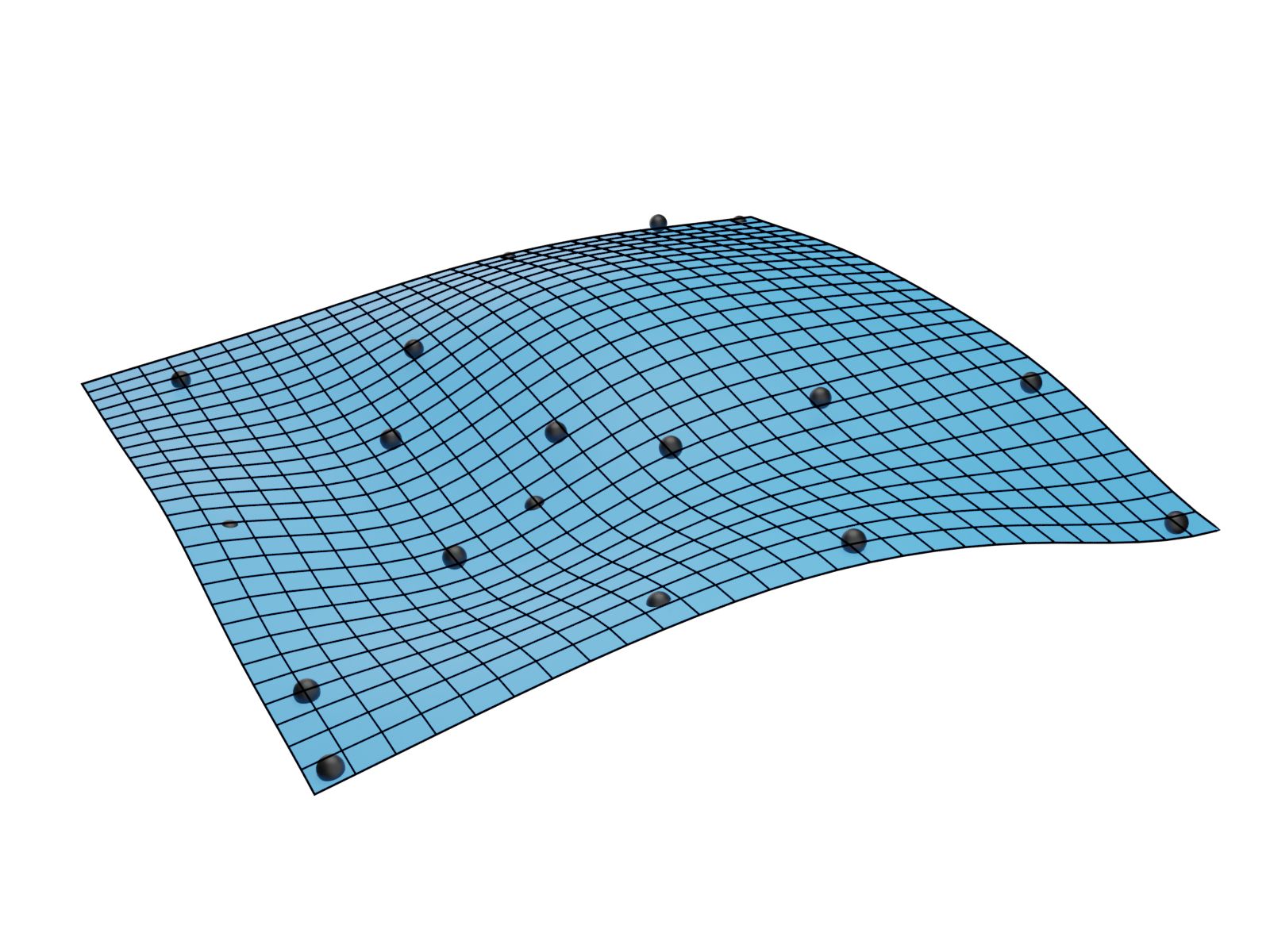}\label{fig:survey:approx_surf}}
         &  
         \subfloat[Volume-based interpolation]{\includegraphics[width=\mywidth,mytrim]{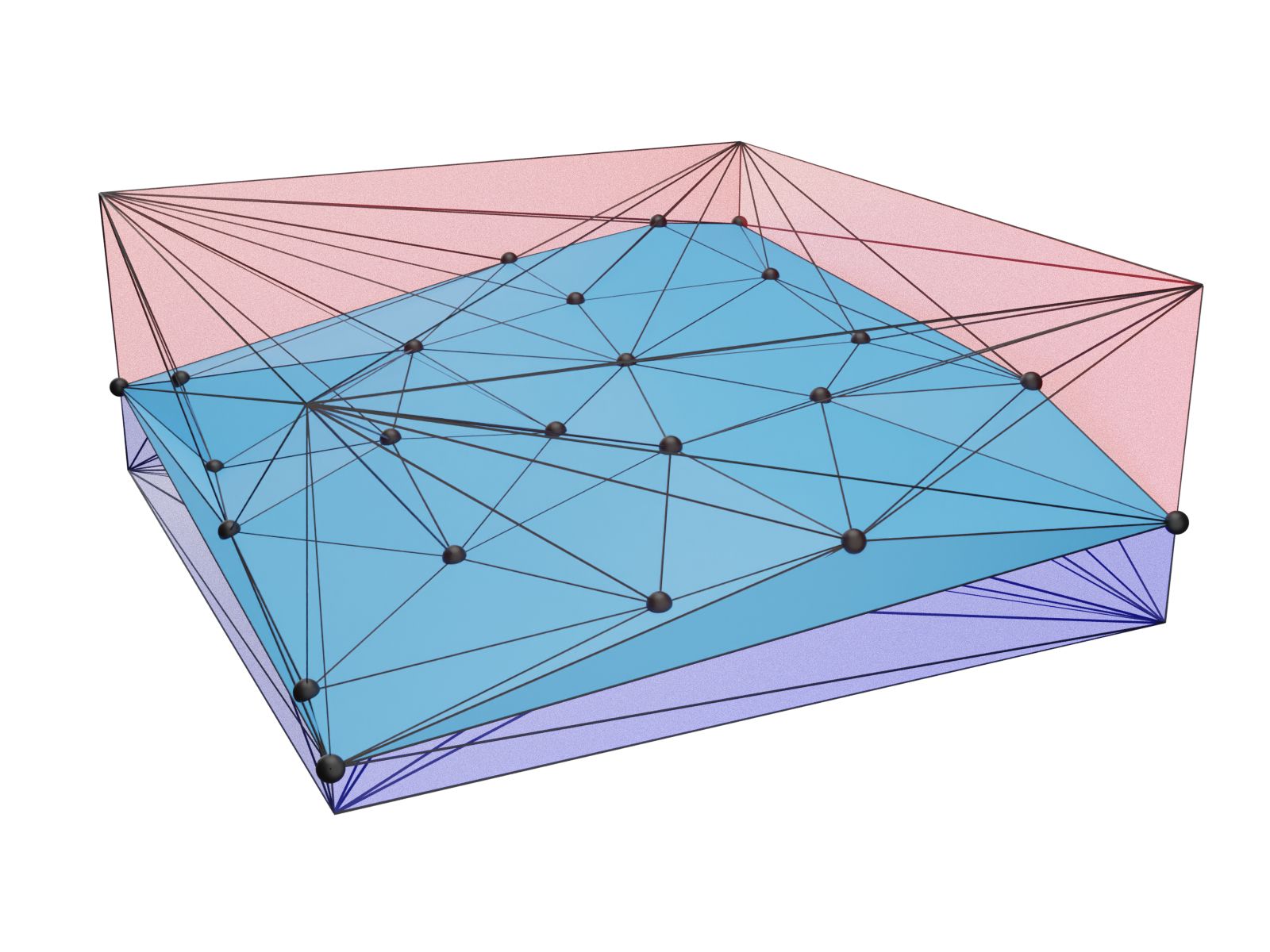}\label{fig:survey:inter_vol}}
         &
         \subfloat[Volume-based approximation]{\includegraphics[width=\mywidth,mytrim]{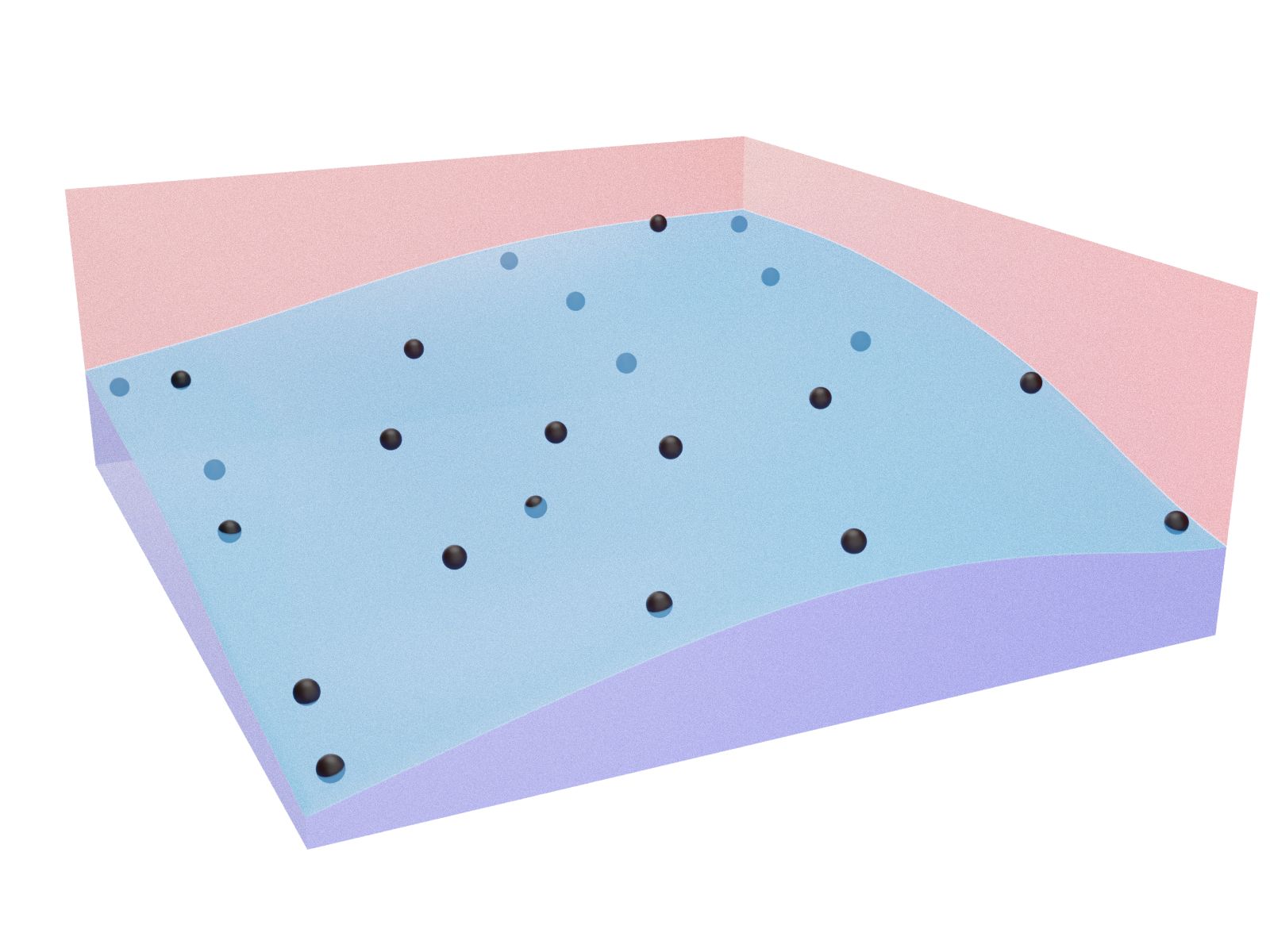}\label{fig:survey:approx_vol}}                 
    \end{tabular}
\caption{\textbf{Examples of surface reconstruction algorithms:} In (\subref{fig:survey:inter_surf}), we exemplify a reconstruction from a surface-based interpolation approach. Close by points are connected with triangles to reconstruct the surface. In (\subref{fig:survey:approx_surf}), we show a reconstruction of a surface-based approximation approach. An initial mesh, \eg a plane, is deformed to fit to the input points. In (\subref{fig:survey:inter_vol}), we show a volume-based interpolation. Input points are connected using a 3D Delaunay tetrahedralization. The resulting tetrahedra are then labelled as inside (dark blue) or outside the surface (red) and the interface triangles constitute the final mesh. In (\subref{fig:survey:approx_vol}), we show a volume-based approximation. The surface is represented as an appropriate level set of an occupancy function.}
    \label{fig:survey}
\end{figure*}

{A similar approach to the Ball-pivoting algorithm, avoiding the fixed ball size, is the regular interpolant method \cite{petitjean2001regular}. It also starts by triangulating a seed triplet of points that adheres to the empty ball property. Then, for each existing edge $\cp_i,\cp_j$ a new triangle $\cp_i,\cp_j,\cp_k$ is added to the triangulation. The point $\cp_k$ is chosen such that the new triangle has the smallest radius circumsphere for all $\cp \in \cP \setminus \{\cp_i,\cp_j\}$.
}

A refinement of the Ball-pivoting algorithm \cite{digne2014bpa_parallel} also use a ball of varying size, and additionally proposes a parallel implementation. Both follow-up works have enhanced the Ball-pivoting algorithm's robustness against nonuniform sampling. However, these methods remain sensitive to point cloud defects such as noise or large occlusions.

\vspace{0.1cm}\noindent\textbf{Selection-based.}
{Similar to advancing-front techniques, some learning-based methods \cite{liu2020ier,PointTriNet} also iteratively build the triangulation from an initial set of candidate triangles.
PointTriNet~\cite{PointTriNet} computes the $k$-nearest neighbor graph of $\cP$ and creates seed triangles.
Then, a first network selects some of these triangles based on their probability of being part of the final surface, which is obtained by considering their neighbouring points and already-formed triangles. Finally, a second network proposes new candidate triangles formed by points adjacent to selected triangles. The new candidates are processed by the first network and the algorithm continues for a user-defined number of iterations. As the loss function is the Chamfer distance between the input points and the reconstructed surface, this method can be trained without ground-truth meshes.}

{The \ac{ier} between two points is the ratio of the Euclidean distance and the geodesic distance along the true surface. An \ac{ier} value near $1$ for a pair of points suggests that the edge connecting them is likely to be part of the surface.
IER-meshing~\cite{liu2020ier} starts with a large set of seed triangles generated from the $k$-nearest neighbor graph of the point cloud. They then estimate the \ac{ier} for the vertices of these triangles using an \ac{mlp} supervised from a ground-truth mesh. Triangles whose vertices exhibit \ac{ier} values close to $1$ are then selected for the mesh, as this indicates a high probability of their edges being part of the actual surface.}
{While they are robust against minor noise to the input point clouds, the aforementioned methods typically result in reconstructed surfaces that are neither manifold nor watertight.}

\vspace{0.1cm}\noindent{\textbf{Projection-based.}
A notable example of projection-based methods is the algorithm by Boissonnat \cite{boissonat1984geometric}, which may be the earliest surface reconstruction algorithm \cite{cazals2006delaunay}. This method starts by computing the tangent plane of each point, projecting their neighboring points onto these planes, computing a 2D Delaunay triangulation of the projected 2D point clouds, and merging these local reconstructions. }

{The use of tangent planes is difficult in areas with high curvature or for reconstructing thin structures. DSE-meshing~\cite{rakotosaona2021dse} addresses these challenges by employing \emph{logarithmic maps} as an alternative. Logarithmic maps offer local surface parametrizations around a point $p$ based on the geodesics from that point, providing a more adaptable approach for complex geometries. This method uses an \ac{mlp} to obtain approximate logarithmic maps for each point. Finally, neighboring maps are aligned and triangulated.
This approach leads to fewer non-manifold edges compared to methods that process triangles independently. However, the surface may still not be watertight.
}

\subsubsection{Surface-based approximation}
\label{survey:surface-based-approximation}

\vspace{0.1cm}\noindent\textbf{Patch fitting.}
{AtlasNet~\cite{Groueix2018}, one of the pioneering learning-based methods in surface reconstruction, employs an original approach. It starts with a triangulated 2D patch and uses a neural network to adjust its vertices' positions. This network is trained by minimizing the distance between the deformed surface and the point cloud, eliminating the need for actual mesh data for training purposes. Similar to interpolating methods, AtlasNet does not inherently ensure that the reconstructed surface is both watertight and manifold.} 

\vspace{0.1cm}\noindent\textbf{Mesh deformation.}
A variation of the same idea is the mesh-fitting approach, which starts with an intial mesh based on some of the input points \cite{sharf2006competing} or a low-resolution Poisson \cite{screened_poisson} reconstruction \cite{point2mesh}, and adjusts the position of the vertices to fit the point cloud. Such methods demonstrate robustness in scenarios with missing data, but they require meticulous parameter tuning to effectively manage noise or outliers.
A notable limitation is the fixed topology of the initial mesh throughout the reconstruction process. Consequently, if the initial mesh does not accurately reflect the true topology of the target surface, this discrepancy cannot be corrected later in the process. For example, if the initial mesh is convex and the desired surface includes holes, it can not be reconstructed with consistent orientation.
A significant advantage of mesh deformation methods compared to other surface-based approaches is that they  guarantee a watertight surface.

\subsection{Volume-based reconstruction}
\label{survey:volume-based}

{Instead of directly generating a surface mesh, volume-based approaches consider discrete or continuous 3D volumes from which a mesh can be extracted. These methods offer a different perspective on the reconstruction process, initially emphasizing volume over surface geometry.}

\subsubsection{Volume-based interpolation}
\label{survey:volume-based-interpolation}

{Volume-based interpolating methods commonly start by constructing a \acs*{3dt} of the the convex hull of the point cloud $\cP$   into tetrahedra (see \figref{fig:survey:inter_vol}). This \ac{3dt} is formed such that no point in $\cP$ lies within the circumspheres of any of these tetrahedra. For point clouds that are well distributed, the \ac{3dt} can be efficiently constructed with a time complexity of $O(n\log{}n)$ \cite{attali2003delaunaycomplexity} where $n$ is the number of vertices. While the Delaunay triangulation itself does not directly yield a surface mesh, a dense-enough sampling of $\cP$ from $\cS$ ensures that a subcomplex of the \ac{3dt} will accurately approximate both the geometry and the topology of $\cS$ \cite{cazals2006delaunay}.}

\vspace{0.1cm}\noindent{\textbf{$\alpha$-shapes.} One straightforward way to extract such a subcomplex from a \ac{3dt} involves two key steps: discarding all tetrahedra whose circumspheres are larger than a predetermined radius $\alpha$, and then retaining only the boundary triangles, \ie triangles between tetrahedra that are discarded and tetrahedra that are kept, resulting in the point cloud's $\alpha$-shape \cite{bernardini1997alphashapes}. Like the Ball Pivoting algorithm, the selection of $\alpha$ depends on the density of points. For samplings that are error-free and dense, $\alpha$-shapes, along with other interpolation methods \cite{cazals2006delaunay, amenta1998crust, boissonat1984geometric}, can offer provable guarantees of topological accuracy of the reconstructed surface \cite{cazals2006delaunay}.
A recent approach constructs a \ac{3dt} that is greedily refined and carved with an empty ball of size $\alpha$ to reconstruct a watertight 2-manifold surface \cite{portaneri2022alpha}. The surface strictly encloses the input point set by a positive user-defined offset.}

 \vspace{0.1cm}\noindent{\textbf{Inside-outside labeling.} An alternative method to derive a surface from a \ac{3dt} is by labeling each tetrahedron as being inside or outside \cite{kolluri2004ssr,sinha2007multi,hiep2009towards,Zhou2019,mostegel2017scalable,Vu2012,Labatut2009a,Jancosek2011,Jancosek2014,wasure, dgnn}. In this approach, every tetrahedron in the \ac{3dt} of $\cP$ is labeled as either \emph{inside} or \emph{outside} relative to $\cSr$ (dark blue and red regions in \figref{fig:survey:inter_vol}). The surface is then defined as the interface between tetrahedra with different labels. This technique guarantees that the resulting surfaces are both intersection-free and watertight.}
 {The problem of labeling tetrahedra is typically formulated with a global energy under a form that can be efficiently minimized by a graph cut \cite{boykov2004experimental}. Unary inside/outside potentials are determined based on visibility information. Meanwhile, binary terms incorporate priors related to the smoothness or area of the reconstructed surface.  This approach is robust to various moderate acquisition defects \cite{Jancosek2011,Labatut2009a,Vu2012} and can handle large-scale scenes \cite{mostegel2017scalable, caraffa2021efficiently}. 
 {In particular, \ac{resr} \cite{Labatut2009a} has a specific treatment to prevent tiny elongated tetrahedra from being wrongly labeled.} 
  DeepDT~\cite{luo2021deepdt} uses a \ac{gnn} operating on a \ac{3dt} to predict the global energy parameters. \ac{dgnn} \cite{dgnn} improves on this idea by integrating visibility information and using a scalable graph formulation operating on small subgraphs, allowing it to scale to large point clouds.}

\subsubsection{Volume-based approximation}
\label{survey:volume-based-approximation}

\vspace{0.1cm}\noindent\textbf{Implicit functions.} Implicit functions form the cornerstone of many popular volume-based surface approximation techniques. The seminal work by Hoppe~\etal~\cite{hoppe1992surface} pioneered this approach by defining an \ac{sdf} based on the local tangent planes at each point. However, the computation of these tangent planes proved costly and susceptible to noise, as well as to variations in point density \cite{kolluri2008provably}. Over the years, implicit function methods have been refined and improved to address these challenges.

Arguably the most well-known method for surface reconstruction, Poisson surface reconstruction was introduced in 2006 by Kazhdan \etal~\cite{kazhdan_poisson_2006}. The method uses an indicator function $F$ and a level set that defines the reconstructed surface $\cSr$. The main idea of the approach is that the Laplacian of $F$ should equate to the divergence of a vector field $\vec{N}$ defined by the oriented normals of the points of $\cP$:
\begin{equation}
\label{eq:poisson}
	\Delta F = \nabla \cdot \vec{N}~.
\end{equation}
The reconstruction process starts by building an octree from $\cP$ and defines a system of hierarchical functions for each cell of the tree. The function coefficients are computed by solving a sparse linear system, making the method time and memory-efficient.

{While Poisson reconstruction yields a closed and smooth surface, it tends to erase small details and structures. \ac{spsr}~\cite{screened_poisson} addresses this problem by constraining the implicit function, and thus reconstructed surface, to pass through or near all points of $\cP$. 
Moreover, this refined variant adds Neumann boundary conditions \cite{cheng2005heritage}, which allow the reconstructed surface to intersect the boundary of the domain on which $F$ is defined. This was further refined by adding Dirichlet boundary conditions around a tight envelope enclosing the point cloud $\cP$ and leading to better reconstructions in areas of missing data \cite{Kazhdan2020envelope}.
Poisson methods generate watertight meshes and are robust against a wide array of acquisition defects of moderate amplitude. However, a notable prerequisite is the need for well-oriented normals, which can be a significant challenge in real-world data acquisition.}

\vspace{0.5cm} Rather than being characterized by its Laplacian and boundary conditions, the implicit function $F$ can be represented as a neural field with a neural network \cite{Mescheder2019,Park2019,IM-Net}. This broad category of methods is further subdivided into three distinct subgroups based on the nature of the network employed: set-based, grid-based, or neighborhood-based. Each subgroup represents a unique approach to leveraging neural networks for the representation of implicit functions.

\vspace{0.1cm}\noindent{\textbf{Set-based \acp{nif}.}
This first class of approaches consider the input point cloud as an unordered set of points and compute a global embedding for its global structure. }
\ac{onet} \cite{Mescheder2019} defines $F$ as a \ac{fcn} conditioned by the input point cloud $\cP$. This network is trained to predict whether a set of preset points lie inside or outside the surface $\cS$. To accurately train the network, true watertight surfaces are required for reference, {such as those found in some CAD-based object databases.}

{DeepSDF~\cite{Park2019} takes a different approach by conditioning $F$ on a latent shape code optimized for a given input point cloud $\cP$ during inference. This optimization is facilitated by an auto-decoder network trained on a collection of shapes. However, the optimization process requires accurate signed distance values for the input point cloud, which can be challenging for complex or ambiguously oriented shapes.}

{The \ac{lig}~\cite{lig} and DeepLS~\cite{deepLS} approaches split the input point cloud $\cP$ into overlapping subregions and process them independently with weight-sharing networks. A benefit of this approach is that similar areas belonging to different objects, such as flat surfaces on tabletops or floors, are processed uniformly, thus reducing the risk of the network overfitting to specific training scenes. 
However, as DeepSDF, these techniques also require true values of the implicit function $F$ for the shape during inference. Furthermore, the dimension of the subregions is a critical parameter that requires careful calibration to ensure successful reconstruction.}

{
\ac{igr}~\cite{Gropp2020} proposes a different approach in which a \ac{fcn} is optimized during inference for each test shape such that the signed distance's $0$-level set goes through the points of $\cP$ and its Laplacian follows the points normal. As this approach is optimized for each shape independently, it does not require true meshes for supervision, but it also cannot learn shape priors.
}

\vspace{0.1cm}\noindent{\textbf{Grid-based \acp{nif}.}
As they rely on a single {point cloud embedding} to define the implicit function $F$, set-based \ac{nif} methods tend to discard local point cloud information, which often results in oversmoothing \cite{Peng2020}. To address this issue, the 3D space can be discretized into a voxel grid {which can be used to learn local features.}. Then, \ac{2d} or \ac{3d} U-Nets \cite{ronneberger2015unet2d,cciccek2016unet3d} process the resulting maps to capture local and global shape information and, in turn, connect a \acl{fcn} predicting occupancy.  
\ac{conet}~\cite{Peng2020} exemplifies this approach and also proposes a sliding-window strategy to scale to larger scans. However, since the content of the windows must be consistent with the scale of training shapes, this approach can be computationally expensive for large scenes.
}

Using a similar encoder architecture, \ac{sap}~\cite{Peng2021SAP} merges neural implicit fields with Poisson reconstruction techniques. The process starts by densifying the input point cloud and estimating the point normals. This enhanced point cloud is then fed into a differentiable Poisson solver, which determines the occupancy for each cell of a regular grid {(\cf the indicator function $F$ of Poisson)}.
These predicted values are then compared to true occupancies with the L2 loss, leading to a fully differentiable pipeline where point encoding, densification, and normal estimation are all learned end-to-end.
{\acl{sap} actually comes in two flavors: a purely \ac{sapopt}, which does not use any training, and a learning-based setting, for which we retain the name \ac{sap}.}

{\ac{nkf}~\cite{williams2022nkf} also combine learned features with techniques of the classical Poisson surface reconstruction. Input points are encoded into a feature grid, which is then used to compute local kernel basis functions for every point.
Similar to PSR, the function coefficients are computed by solving a sparse linear system and using occupancy information from oriented normals.}
{Recently, \ac{nkf} has been updated to make the method more scalable and robust to noise, by using hierarchical features from multiple voxel grids at different resolutions and aligning the implicit function gradients with the normal direction of the input points \cite{huang2023nksr,huang2022neuralgalerkin}.}
{In contrast to the original \acl{psr}, the aforementioned methods allow to incorporate learned priors into the implicit function computation. Additionally, for \ac{sap} the input point cloud does not need to be equipped with oriented normals.}

\vspace{0.1cm}\noindent{\textbf{Neighborhood-based \acp{nif}.} The choice of the voxel size for 3D convolution-based methods is often critical: opting for small voxels increases computational and memory requirements, whereas coarse grids discard geometric details. \ac{p2s} \cite{points2surf} proposes an alternative approach that predicts signed distance functions by encoding both the local neighborhood of each point (for unsigned distance prediction) and a downsampled representation of the entire shape (for orientation prediction). Leveraging $k$-nearest neighbor sampling, this method offers better adaptability to variations in point density. However, it may result in extended processing times since the local neighborhoods of each point needs to be processed individually during the inference stage.
}
{\ac{poco}~\cite{boulch2022poco} proposes instead to use a continuous convolution backbone \cite{boulch2020convpoint} to learn point features without spatial discretization. The occupancy of each point is then determined by an attention-based scheme.
}

\subsection{Rendering-based reconstruction}
\label{survey:rendering-based}
\RAPH{%
Our paper specifically addresses surface reconstruction methods from 3D point clouds. 
However, a new emerging field is the reconstruction of surfaces directly from images using rendering techniques from computer graphics.
Originally, so-called novel view synthesis methods \cite{mildenhall2020nerf,mildenhall2021nerf} were specifically designed for generating new views of an object or scene starting from a set of given images with their accurate pose. These methods present many advantages, including the ability to generate photo-realistic content, to deal with complex lighting and transparencies \cite{IchnowskiAvigal2021DexNeRF}, and the fact that they do not rely on complex sensors. 
Recently, more and more rendering-based methods have been designed specifically for surface reconstruction \cite{UNISURF,yariv2023bakedsdf,guedon2023sugar}. 
They can be divided into the two categories, \acp{nerf} and 3D Gaussian Splatting, and are closely related to the two corresponding, more general, principles from computer graphics, namely ray tracing and rasterization.}


\subsubsection{\ac{nerf} and Gaussian Splatting}

\LOIC{\ac{nerf} encode scenes within the weights of a neural network \cite{mildenhall2020nerf} by mapping 3D locations and viewing directions with a volume density and emitted color. Although initially designed for novel view synthesis, extensions such as NeuS \cite{wang2021neus} and VolSDF \cite{yariv2021volume} reparameterize the volume density using signed distance functions, allowing explicit surface extraction from radiance fields. \ac{nerf} have also been extended from object-level to entire scenes, such as Urban Radiance Fields \cite{rematas2022urban}, which incorporate point clouds, and Neuralangelo \cite{li2023neuralangelo}, which employs instant neural graphics primitives to improve reconstruction speed.}

\LOIC{3D Gaussian Splatting \cite{kerbl20233d} provides an efficient rendering technique based on differentiable rasterization of 3D Gaussians initialized from sparse point clouds. While primarily aimed at novel view synthesis, preliminary works like SuGaR \cite{guedon2023sugar} and Gaussian Frosting \cite{guedon2024gaussian} explore efficient mesh reconstruction and real-time rendering within this framework. These methods highlight the potential of rasterization-based techniques for surface reconstruction, although they are still in the early stages.}

\section{Benchmark setup}
\label{ch2:sec:benchmark}

{In this section, we propose a series of experiments to benchmark surface reconstruction algorithms discussed in the previous section. 
We first describe our approach to generate realistic point clouds (\secref{ch2:sec:scanning}) and detail the different datasets that we use (\secref{ch2:sec:dataset}). We then define the experimental setup (\secref{ch2:sec:setup}), the methods evaluated  (\secref{ch2:sec:methods}) and finally the benchmark's metrics (\secref{ch2:sec:metrics}).}


\subsection{Synthetic scanning for point cloud generation}
\label{ch2:sec:scanning}

{In an ideal scenario, we would use real point cloud acquisitions and compare the reconstructed surfaces with the associated true surfaces. However, it is often impractical to obtain the true surfaces of real objects, as it requires high-density and error-free scans as well as substantial manual intervention. }
{Many existing \ac{mvs} benchmarks derive their point clouds from images, and the true surfaces are obtained with extensive acquisitions from multiple stationary \ac{lidar} scans \cite{Strecha08,middlebury,dtu,Schops2017,tanksandtemples}. 
However, even these surfaces may be incomplete or not watertight due to occlusions. Additionally, these datasets are typically limited in scale, featuring only a few objects or scenes. As a result, while they are valuable for evaluation purposes, they are less suitable for training models.}
{Instead, we propose to use synthetic surfaces from extensive shape collections for training, since they come with strong topological guarantees. In order to  reproduce the defects commonly encountered during real scans, such as {density variations} 
and missing data from occlusion, we implement realistic artificial scanning procedures, illustrated in \figref{fig:scan}.}

\subsubsection{Synthetic MVS}
We propose a procedure designed to replicate the typical characteristics of point clouds obtained from MVS techniques. While our approach doesn't follow an MVS pipeline, it aims to produce point clouds with similar defects as those found in MVS-generated point clouds. However, it does not model failure cases from MVS methods, such as the lack of \RAPH{points in areas with difficult lighting conditions or without texture, or strong variations in density and noise levels.} 

We randomly position virtual sensors around a sphere enclosing the object, and shoot rays towards random points on the object's circumsphere. A new 3D point is recorded at the first intersection of the ray with the object’s surface. Further intersections are ignored, leading to similar occlusion patterns as in real world acquisitions. We create two types of scans using this methodology:
(i) sparse point clouds, consisting of approximately 3,000 points per object;
(ii) dense point clouds, comprising around 10,000 points per object, with 10\% of these points being outliers.
Consistent with the approach suggested by Peng \etal \cite{Peng2020}, we apply Gaussian noise to each point position, characterized by a zero mean and a standard deviation of 0.5\% of the bounding box diagonal. Both sparse and dense scans are conducted from 10 uniformly distributed sensor positions.


\subsubsection{Synthetic range scanning}
We use a modified version of the range scanning procedure from the surface reconstruction benchmark of Berger \etal \cite{Berger_benchmark}.
We choose five distinct scanning settings, each presenting varying levels of difficulty, to scan each shape: 
\begin{itemize}
    \item \textbf{LR:} \textit{Low resolution} to  replicates point clouds obtained from long-range scanning;
     \item \textbf{HR:} \textit{High resolution} with close to no defects;
     \item \textbf{HRN:} \textit{Noisy acquisition} by adding random jitter on the point positions;
     \item \textbf{HRO:} \textit{Outliers} by adding points uniformly sampled the bounding box of the object;
     \item \textbf{HRNO:} \textit{Noise and outliers} combines the last two defects.
\end{itemize}
See the supplementary material for the exact numbers and the original benchmark paper \cite{Berger_benchmark} for further details.


\subsection{Datasets}
\label{ch2:sec:dataset}

{To train and evaluate various surface reconstruction algorithms from point clouds, we use a mix of synthetic and real shapes. See the supplementary material for visual examples.}

\vspace{0.1cm}
\noindent{\textbf{Training sets.}
To have enough shapes to train deep networks, we use the synthetic and closed shapes from ShapeNet \cite{chang2015shapenet} and ModelNet \cite{wu20153d}. Point clouds are generated using our \ac{mvs} procedure, and the true surfaces are processed with ManifoldPlus \cite{huang2020manifoldplus} to make them watertight.
We generate point clouds with 3000 and 10,000 points for ScanNet, and 3000 points for ModelNet as they are typically simpler models. Following the standard practice in this field, we use Choy \etal's \cite{choy20163dr2n2} 13-class subset of ShapeNet and the corresponding train/val/test split.
We use the training sets to parametrize non-learning-based methods, ensuring a fair comparison across different algorithms.
}

\vspace{0.1cm}
\noindent{\textbf{Evaluation.}
{We evaluate the methods} using five shapes from the benchmark of Berger \etal \cite{Berger_benchmark} with challenging features, such as intricate details and complex topology. These shapes present a higher reconstruction difficulty compared to those in ModelNet. For each shape, we generate $4$ point clouds: with 3000 and 10,000, and using synthetic \ac{mvs} and range scanning procedures.
For qualitative evaluation, we also consider real scans, including open surfaces.
We select a range scan from Tanks and Temples \cite{tanksandtemples} (Truck), and two \ac{mvs} point clouds from DTU \cite{dtu} (scan1) and from Middlebury \cite{middlebury} (Temple), each subsampled to 50,000 points.
}

\subsection{Design of the experiments}
\label{ch2:sec:setup}

\setlength{\fboxrule}{5pt}
\setlength{\fboxsep}{0pt}

\newcommand{\pic}[1]{
\begin{tikzpicture}
 \foreach \X [count=\Z]in {a,b,c,d}
 {\node[] at (0,0,\Z/3) {\frame{#1}};}
\end{tikzpicture}
}

\def \eil {figures/exp/02691156/1bea1445065705eb37abdc1aa610476c}
\def \eir {figures/exp/02691156/d18592d9615b01bbbc0909d98a1ff2b4}
\def \eiir {figures/exp/table/0008}
\def \eiiir {figures/exp/dc/dc}
\def \eiil {figures/exp/table/0470}
\def \eivr {figures/exp/dc/dc}

\begin{table*}
    \newcommand{\mywidth}{0.18\textwidth}
    \definetrim{mytrim}{0 0 0 0}
\caption{\textbf{Benchmark setup:} %
We propose several experiments to evaluate the robustness of surface reconstruction models.
In \protect\hyperlink{e1}{E1} to \protect\hyperlink{e4}{E4}, 
we train  learning and traditional methods on a set of shapes and evaluate them on a test set with varying characteristics: identical (\protect\hyperlink{e1}{E1}), with different point cloud defects (\protect\hyperlink{e2}{E2}), with simpler (\protect\hyperlink{e3}{E3}) or more complex (\protect\hyperlink{e4}{E4}) shapes. 
We evaluate on ShapeNet and ModelNet, with synthetic MVS point clouds. In \protect\hyperlink{e5}{E5}, we evaluate neural-based and traditional optimization methods that do not tune their parameters; we use the few but varied synthetic range scans of Berger \etal's dataset. In \protect\hyperlink{e6}{E6} and \protect\hyperlink{e7}{E7}, we compare all methods quantitatively and qualitatively on synthetic and real point clouds.
}
\resizebox{\textwidth}{!}{
\centering
\begin{tabular}{ccccc}
\toprule
\Large\textbf{Experiment}
& \multicolumn{2}{c}{\textbf{\Large\textbf{Training set}}}        &  \multicolumn{2}{c}{\textbf{\Large\textbf{Test set}}} \\\midrule
\multicolumn{5}{c}{\Large \raisebox{5mm}{}Evaluation of methods with dataset-driven parameterization}\\[3mm]
\makecell[c]{\LARGE E1 \vspace{8pt} \\  \large In-distribution}
& \multicolumn{1}{c}{\raisebox{-.5\height}{\pic{\includegraphics[width=\mywidth]{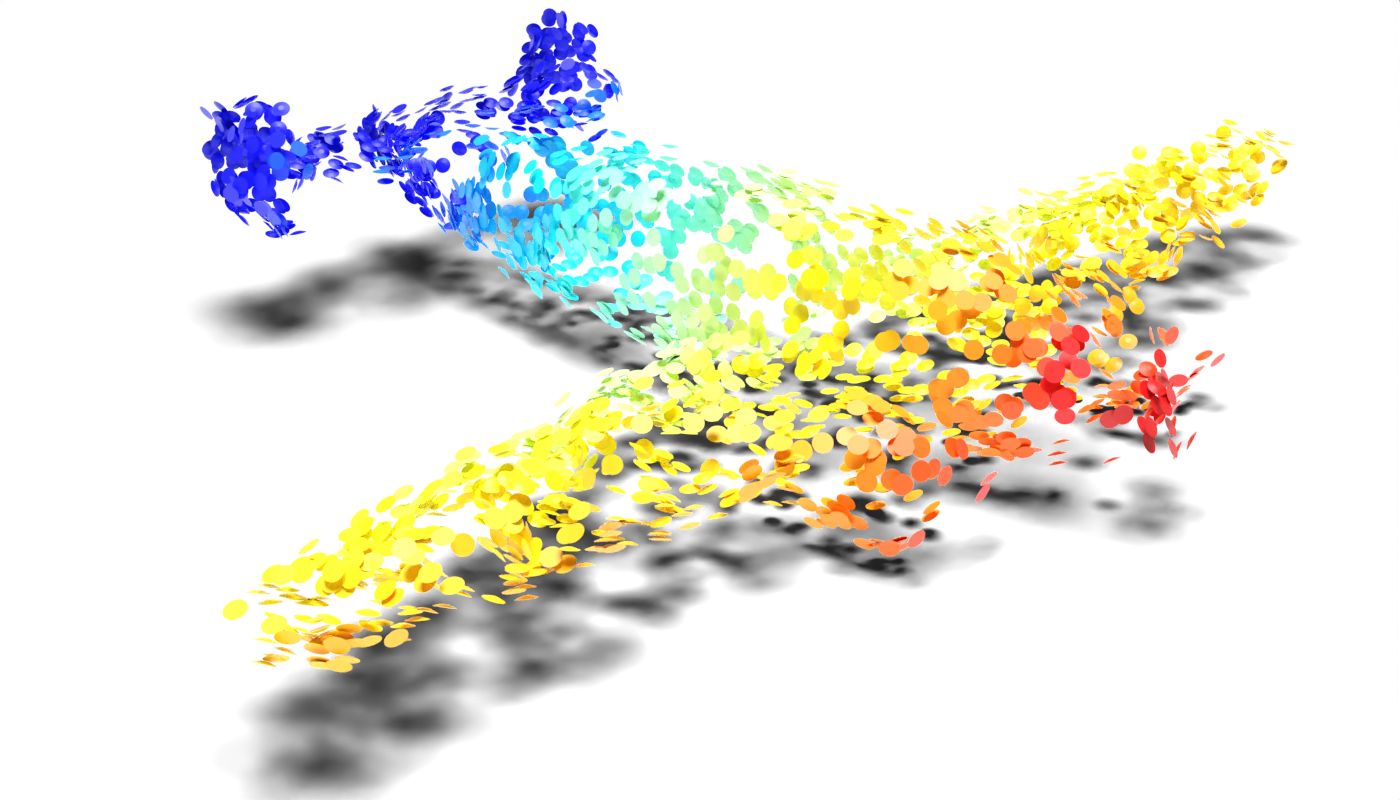}}}}
& \multicolumn{1}{c|}{\raisebox{-.5\height}{\pic{\includegraphics[width=\mywidth]{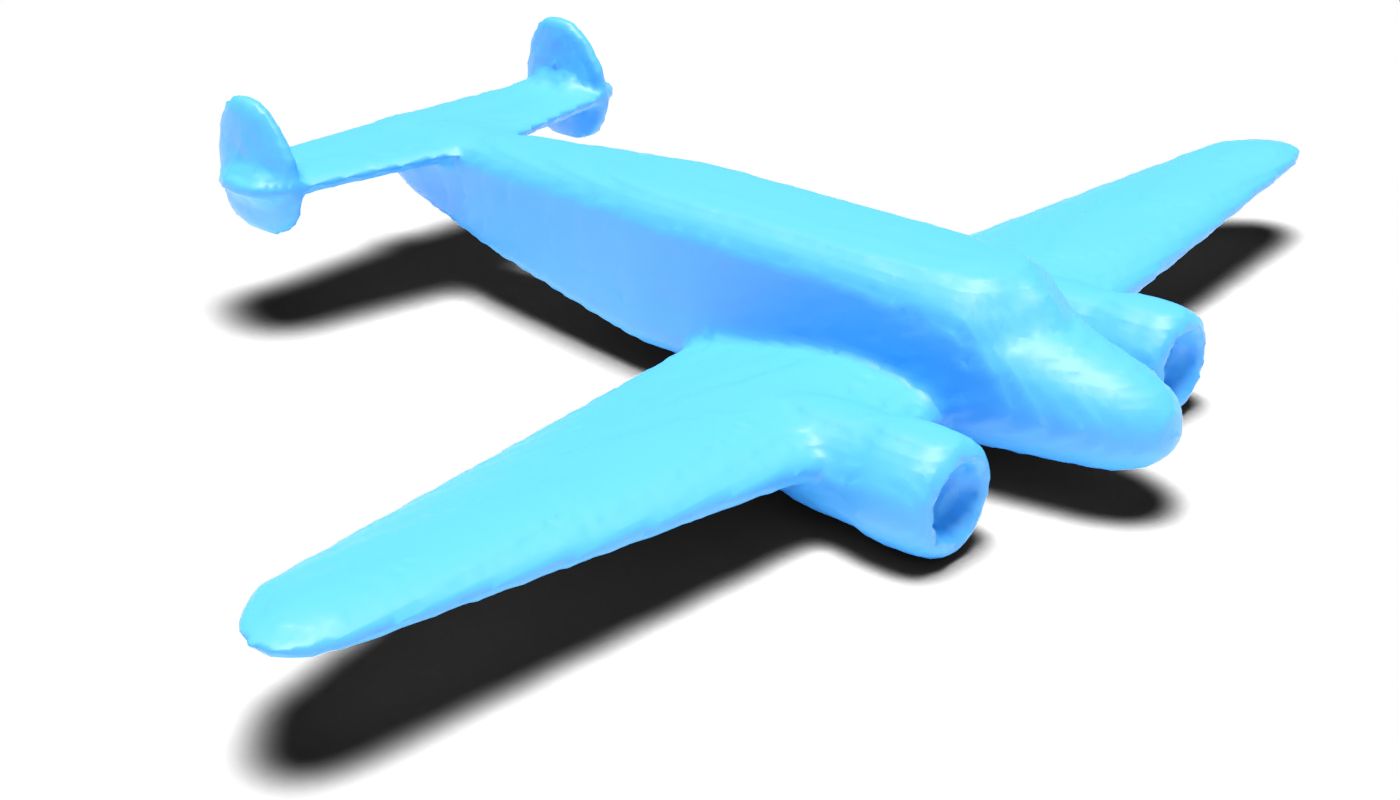}}}}
& \multicolumn{1}{c}{\raisebox{-.5\height}{\pic{\includegraphics[width=\mywidth]{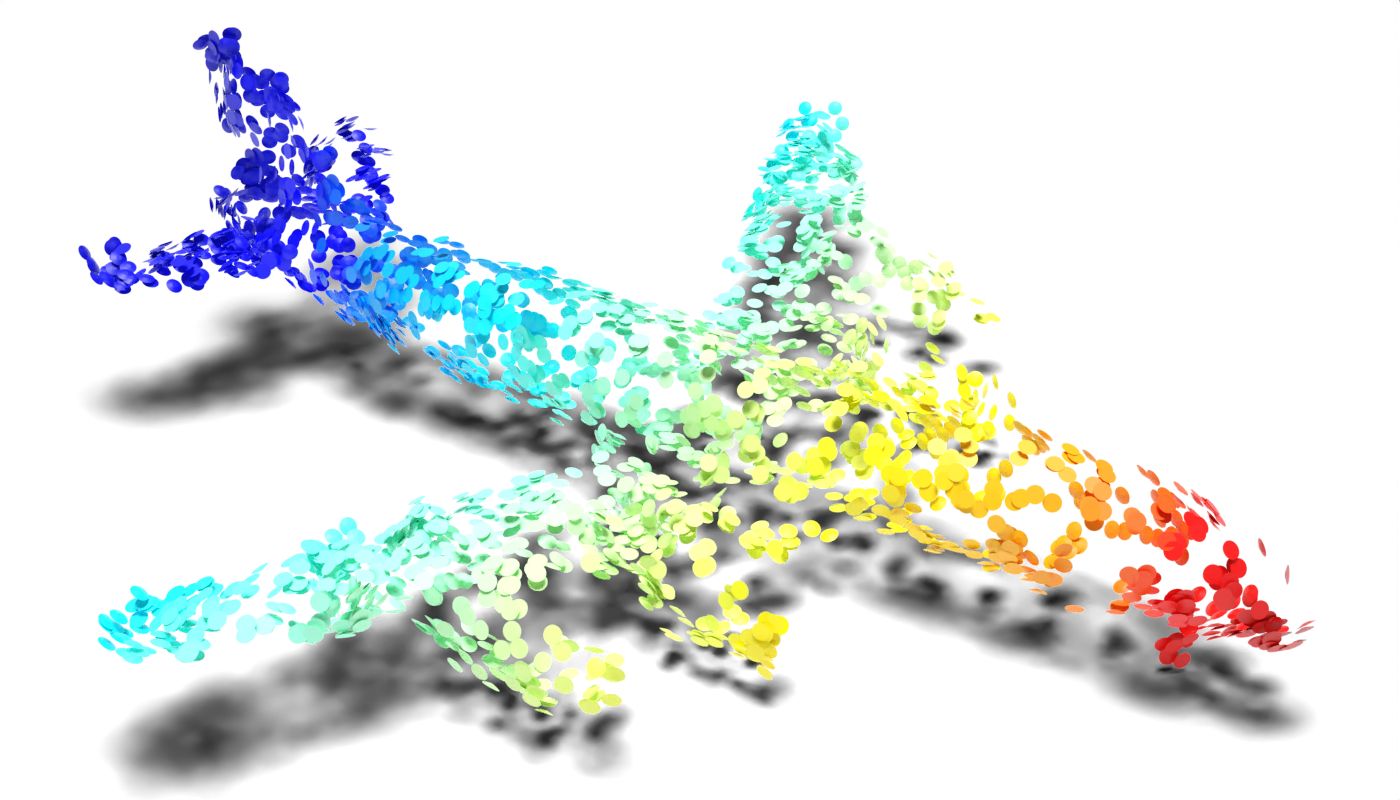}}}}
& \multicolumn{1}{c}{\raisebox{-.5\height}{\pic{\includegraphics[width=\mywidth]{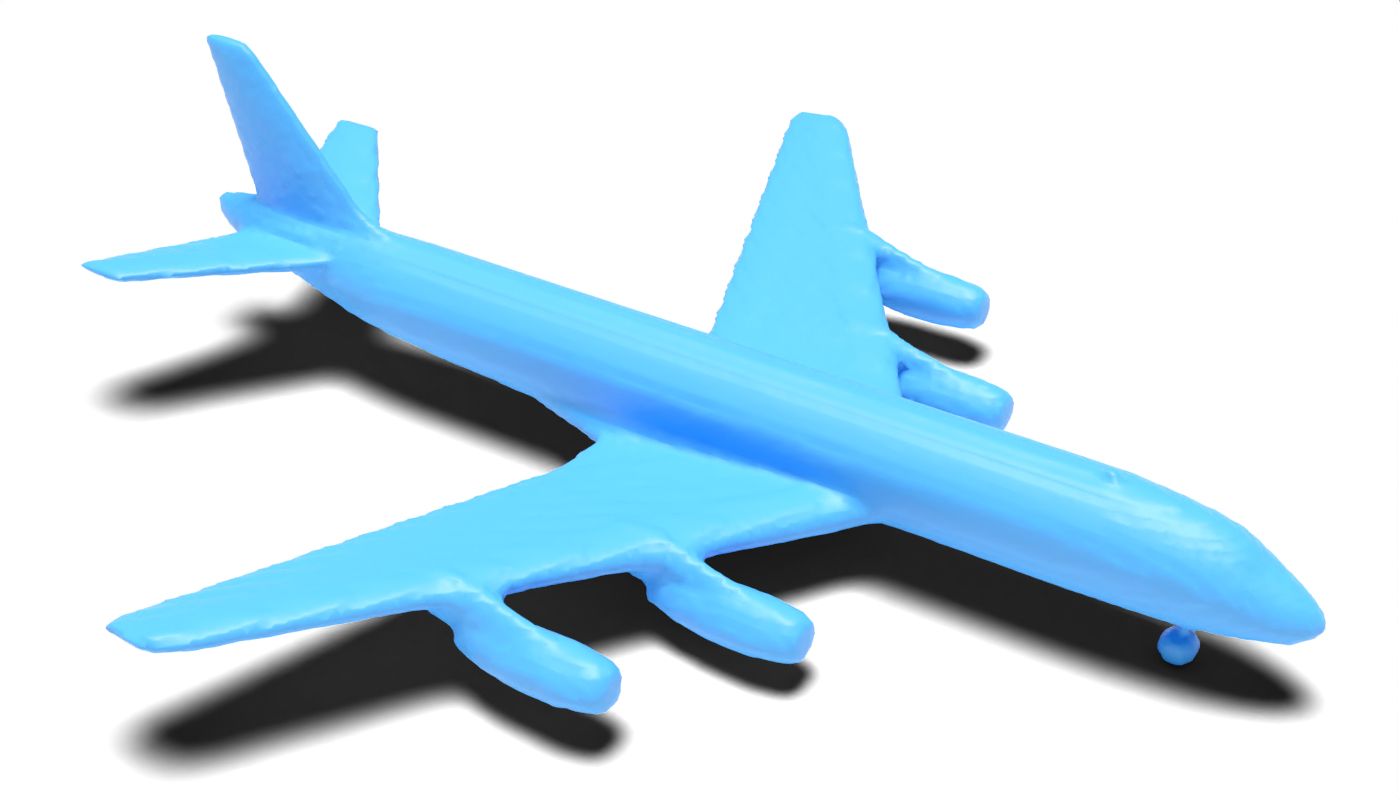}}}}
\\           
& \multicolumn{2}{c|}{\textbf{\textbf{ShapeNet (synthetic MVS)}}}        &  \multicolumn{2}{c}{\textbf{\textbf{ShapeNet (synthetic MVS)}}} 
\\
\makecell[c]{\LARGE E2 \vspace{8pt} \\  \large Out-of-distribution\\
\normalsize \textit{unseen point cloud characteristics:} \\
\normalsize \textit{density, outliers}}
& \multicolumn{1}{c}{\raisebox{-.5\height}{\pic{\includegraphics[width=\mywidth]{\eil/input3000.jpg}}}}
& \multicolumn{1}{c|}{\raisebox{-.5\height}{\pic{\includegraphics[width=\mywidth]{\eil/mesh.jpg}}}}
& \multicolumn{1}{c}{\raisebox{-.5\height}{\pic{\includegraphics[width=\mywidth]{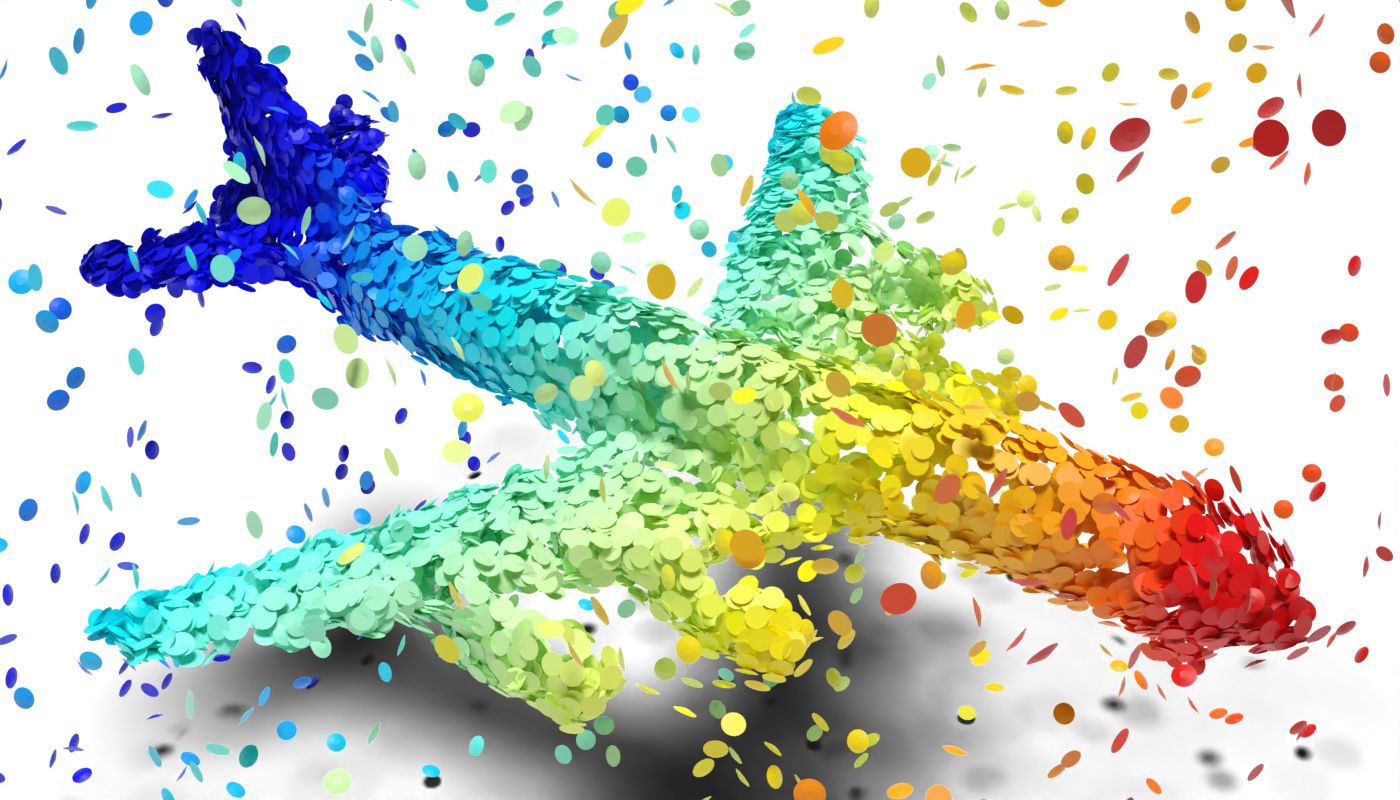}}}}
& \multicolumn{1}{c}{\raisebox{-.5\height}{\pic{\includegraphics[width=\mywidth]{\eir/mesh.jpg}}}}
\\           
& \multicolumn{2}{c|}{\textbf{\textbf{ShapeNet (synthetic MVS)}}}        &  \multicolumn{2}{c}{\textbf{\textbf{ShapeNet (synthetic MVS)}}} 
\\
\makecell[c]{\LARGE E3 \vspace{8pt} \\  \large Out-of-distribution\\
\normalsize \textit{unseen shape categories:} \\ \normalsize \textit{less complex shapes}}
& \multicolumn{1}{c}{\raisebox{-.5\height}{\pic{\includegraphics[width=\mywidth]{\eil/input3000.jpg}}}}
& \multicolumn{1}{c|}{\raisebox{-.5\height}{\pic{\includegraphics[width=\mywidth]{\eil/mesh.jpg}}}}
& \multicolumn{1}{c}{\raisebox{-.5\height}{\pic{\includegraphics[width=\mywidth]{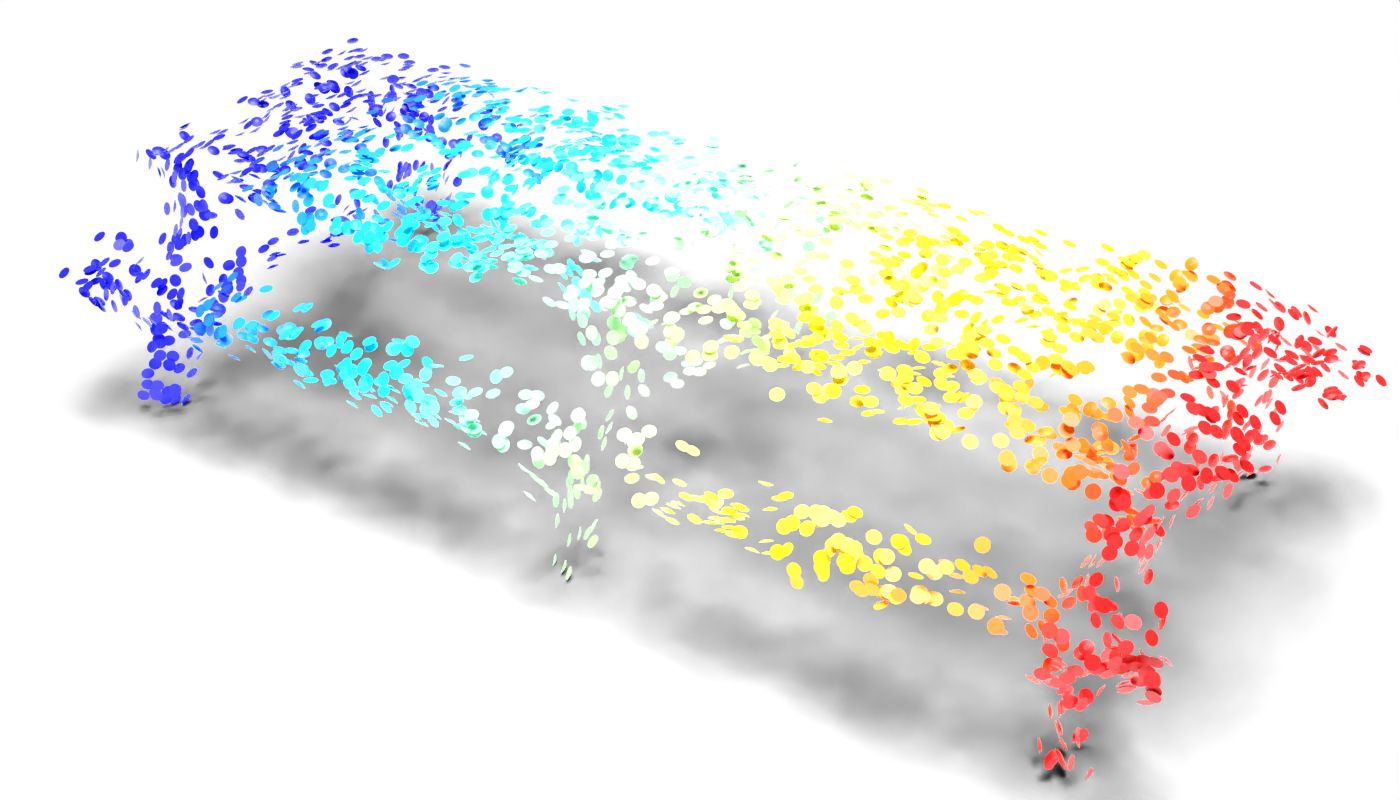}}}}
& \multicolumn{1}{c}{\raisebox{-.5\height}{\pic{\includegraphics[width=\mywidth]{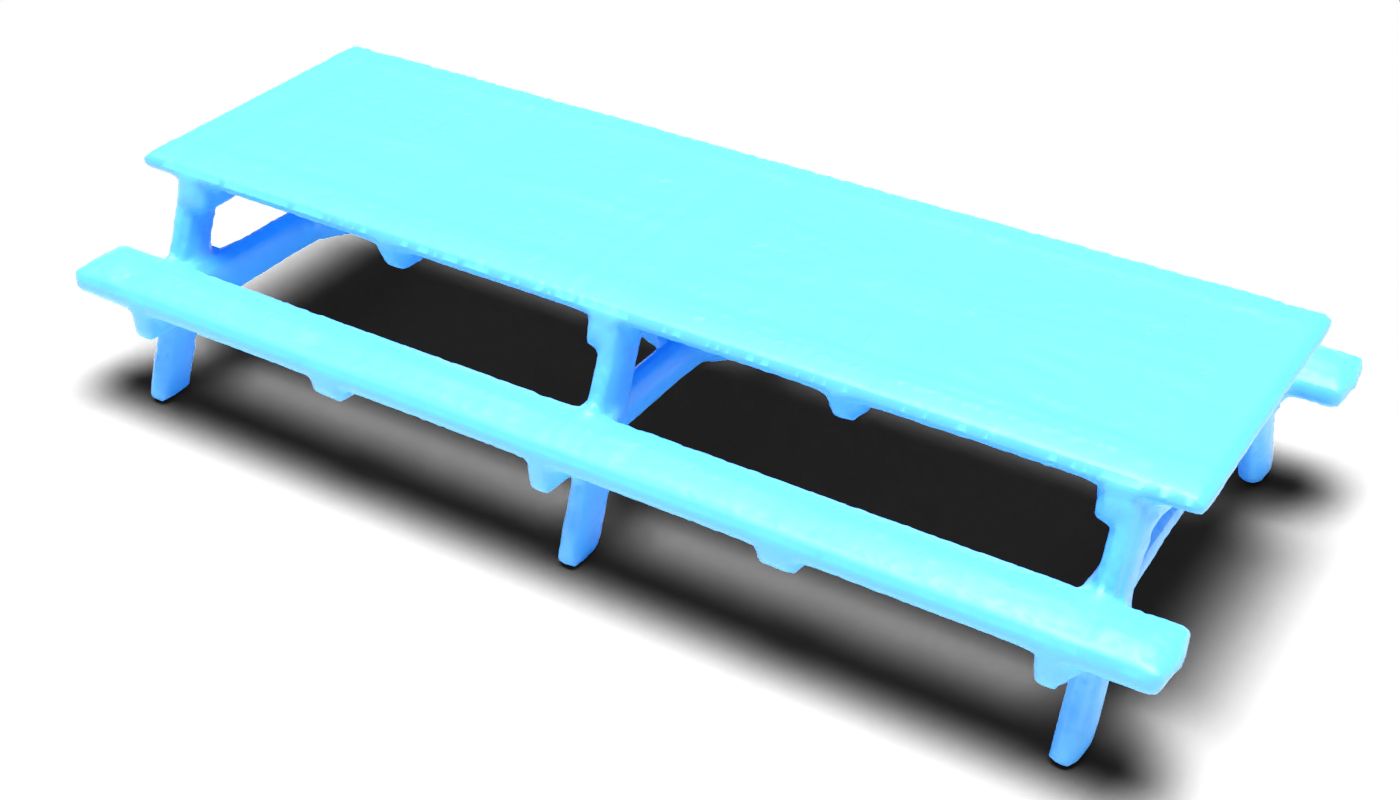}}}}
\\           
& \multicolumn{2}{c|}{\textbf{\textbf{ShapeNet (synthetic MVS)}}}        &  \multicolumn{2}{c}{\textbf{\textbf{ModelNet (synthetic MVS)}}} 
\\
\if 1 0
\makecell[c]{\LARGE E4 \vspace{8pt} \\  \large Out-of-distribution\\
\normalsize \textit{unseen shape categories:} \\ \normalsize \textit{similar complexity}}
& \multicolumn{1}{c}{\raisebox{-.5\height}{\pic{\includegraphics[width=\mywidth]{\eil/input3000.jpg}}}}
& \multicolumn{1}{c|}{\raisebox{-.5\height}{\pic{\includegraphics[width=\mywidth]{\eil/mesh.jpg}}}}
& \multicolumn{1}{c}{\raisebox{-.5\height}{\pic{\includegraphics[width=\mywidth]{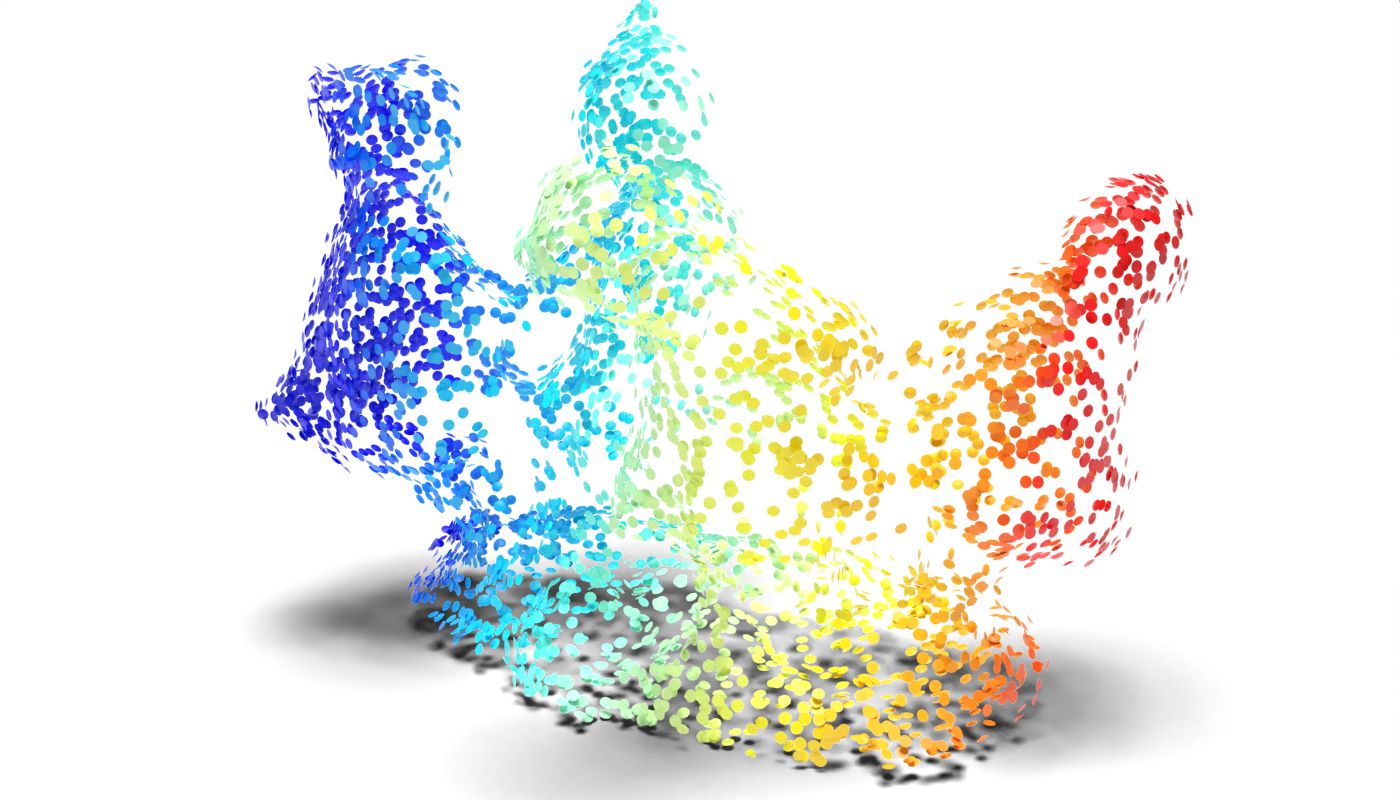}}}}
& \multicolumn{1}{c}{\raisebox{-.5\height}{\pic{\includegraphics[width=\mywidth]{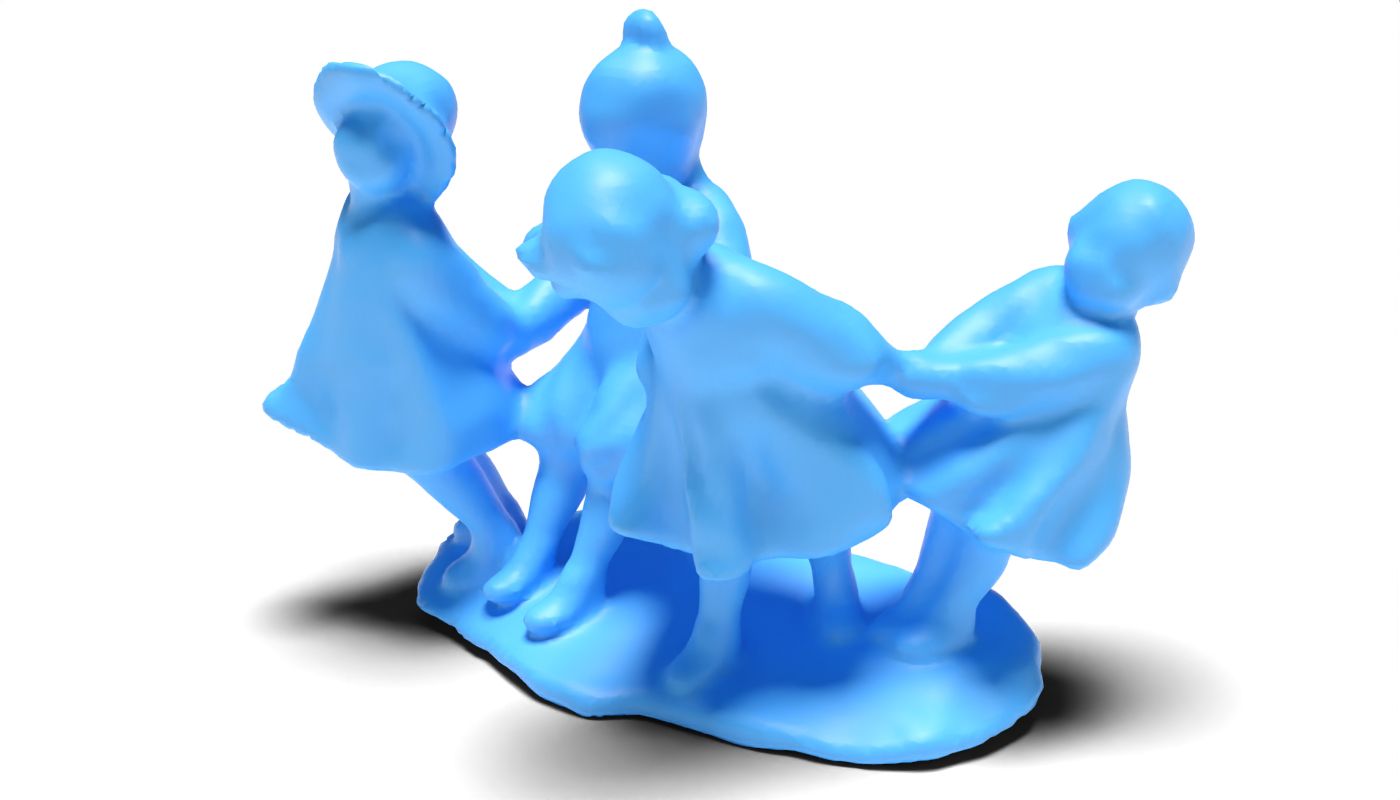}}}}
\\           
& \multicolumn{2}{c|}{\textbf{\textbf{ShapeNet (synthetic MVS)}}}        &  \multicolumn{2}{c}{\textbf{\textbf{Berger \etal (synthetic MVS)}}}
\\
\fi
\makecell[c]{\LARGE E4 \vspace{8pt} \\  \large Out-of-distribution\\
\normalsize \textit{unseen shape categories:} \\ \normalsize \textit{more complex shapes}}
& \multicolumn{1}{c}{\raisebox{-.5\height}{\pic{\includegraphics[width=\mywidth]{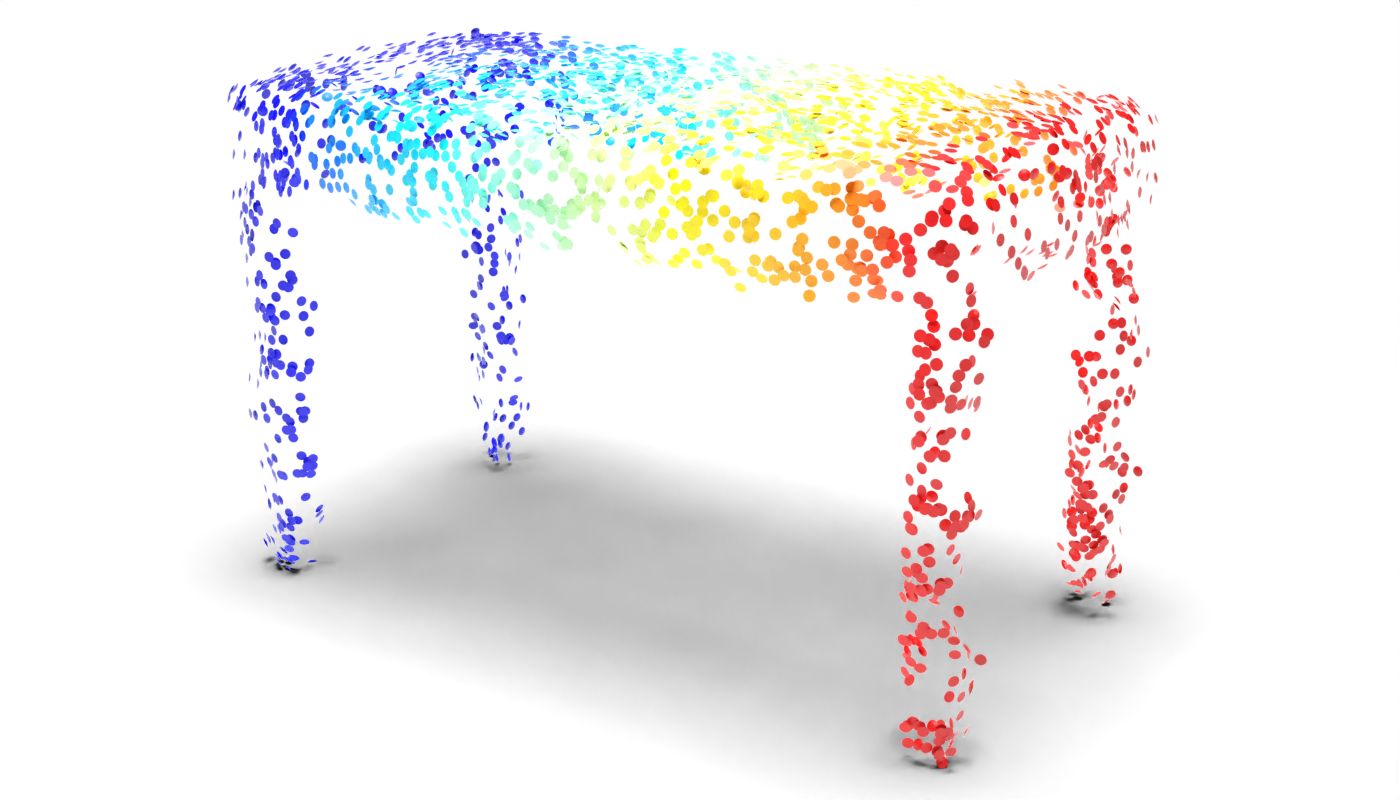}}}}
& \multicolumn{1}{c|}{\raisebox{-.5\height}{\pic{\includegraphics[width=\mywidth]{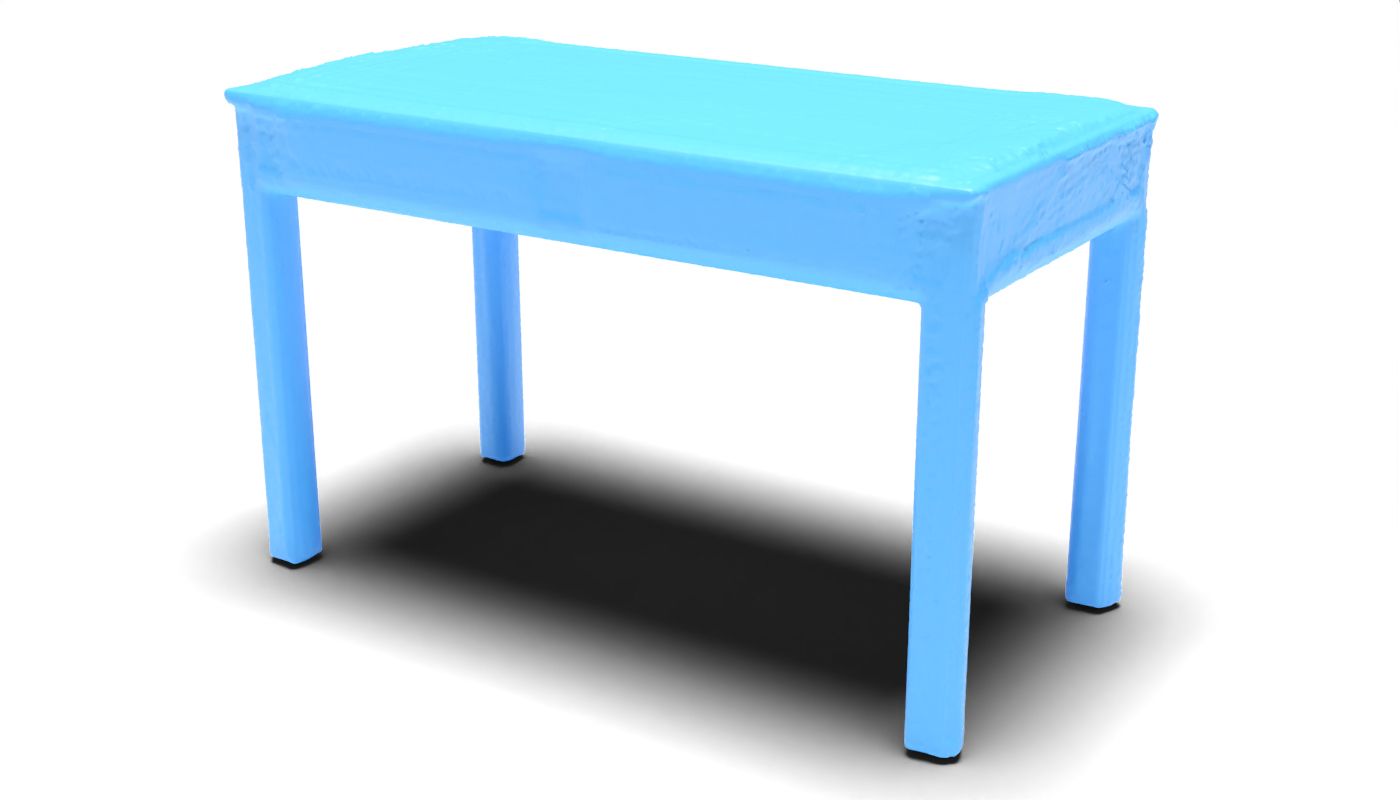}}}}
& \multicolumn{1}{c}{\raisebox{-.5\height}{\pic{\includegraphics[width=\mywidth]{\eir/input3000.jpg}}}}
& \multicolumn{1}{c}{\raisebox{-.5\height}{\pic{\includegraphics[width=\mywidth]{\eir/mesh.jpg}}}}
\\           
& \multicolumn{2}{c}{\textbf{\textbf{ModelNet (synthetic MVS)}}}        &  \multicolumn{2}{c}{\textbf{\textbf{ShapeNet (synthetic MVS)}}} 
\\
\midrule
\multicolumn{5}{c}{\Large  \raisebox{5mm}{}Evaluation of methods without dataset-driven parameterization}\\[3mm]
\makecell[c]{\LARGE E5 \vspace{8pt} \\  \large  Shape variety with defects \\ \normalsize \textit{shapes with varying complexities,} \\ \normalsize \textit{density, noise and outliers}}
   & \multicolumn{1}{c}{\raisebox{-.5\height}{\Huge --}}
& \multicolumn{1}{c|}{\raisebox{-.5\height}{\Huge --}}
& \multicolumn{1}{c}{\raisebox{-.5\height}{\pic{\includegraphics[width=\mywidth,mytrim]{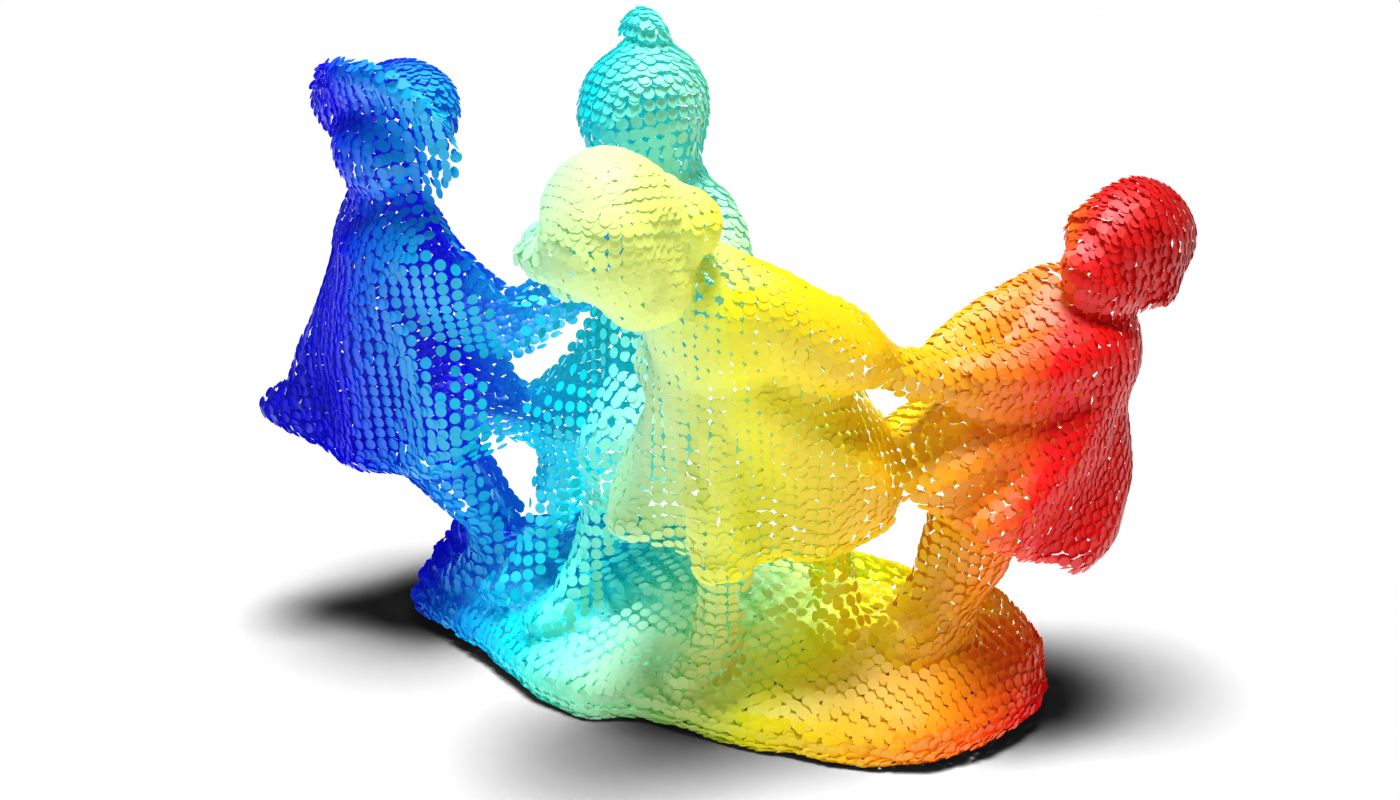}}}}
& \multicolumn{1}{c}{\raisebox{-.5\height}{\pic{\includegraphics[width=\mywidth,mytrim]{\eivr/mesh.jpg}}}}
\\           
& \multicolumn{2}{c}{}        &  \multicolumn{2}{c}{\textbf{\textbf{Berger \etal (synthetic range scan)}}} 
\\                               
\midrule
\multicolumn{5}{c}{\Large \raisebox{5mm}{}Evaluation of all methods on a common ground }\\[3mm]
\makecell[c]{\LARGE E6 \vspace{8pt} \\  \large Synthetic data \\ \normalsize \textit{with ground truth} \\ \normalsize \textit{(quantitative evaluation)}}
& \multicolumn{1}{c}{\raisebox{-.5\height}{\pic{\includegraphics[width=\mywidth]{\eil/input3000.jpg}}}}
& \multicolumn{1}{c|}{\raisebox{-.5\height}{\pic{\includegraphics[width=\mywidth]{\eil/mesh.jpg}}}}
& \multicolumn{1}{c}{\raisebox{-.5\height}{\pic{\includegraphics[width=\mywidth]{\eiiir/input3000.jpg}}}}
& \multicolumn{1}{c}{\raisebox{-.5\height}{\pic{\includegraphics[width=\mywidth]{\eiiir/mesh.jpg}}}}
\\           
& \multicolumn{2}{c|}{\textbf{\textbf{ShapeNet (synthetic MVS)}}}        &  \multicolumn{2}{c}{\textbf{\textbf{Berger \etal (synthetic MVS)}}} 
\\
\makecell[c]{\LARGE E7 \vspace{8pt} \\  \large Real data\\ \normalsize \textit{without ground truth} \\ \normalsize \textit{(qualitative evaluation)}}
& \multicolumn{1}{c}{\raisebox{-.5\height}{\pic{\includegraphics[width=\mywidth]{\eil/input3000.jpg}}}}
& \multicolumn{1}{c|}{\raisebox{-.5\height}{\pic{\includegraphics[width=\mywidth]{\eil/mesh.jpg}}}}
& \multicolumn{1}{c}{\raisebox{-.5\height}{\pic{\includegraphics[width=\mywidth]{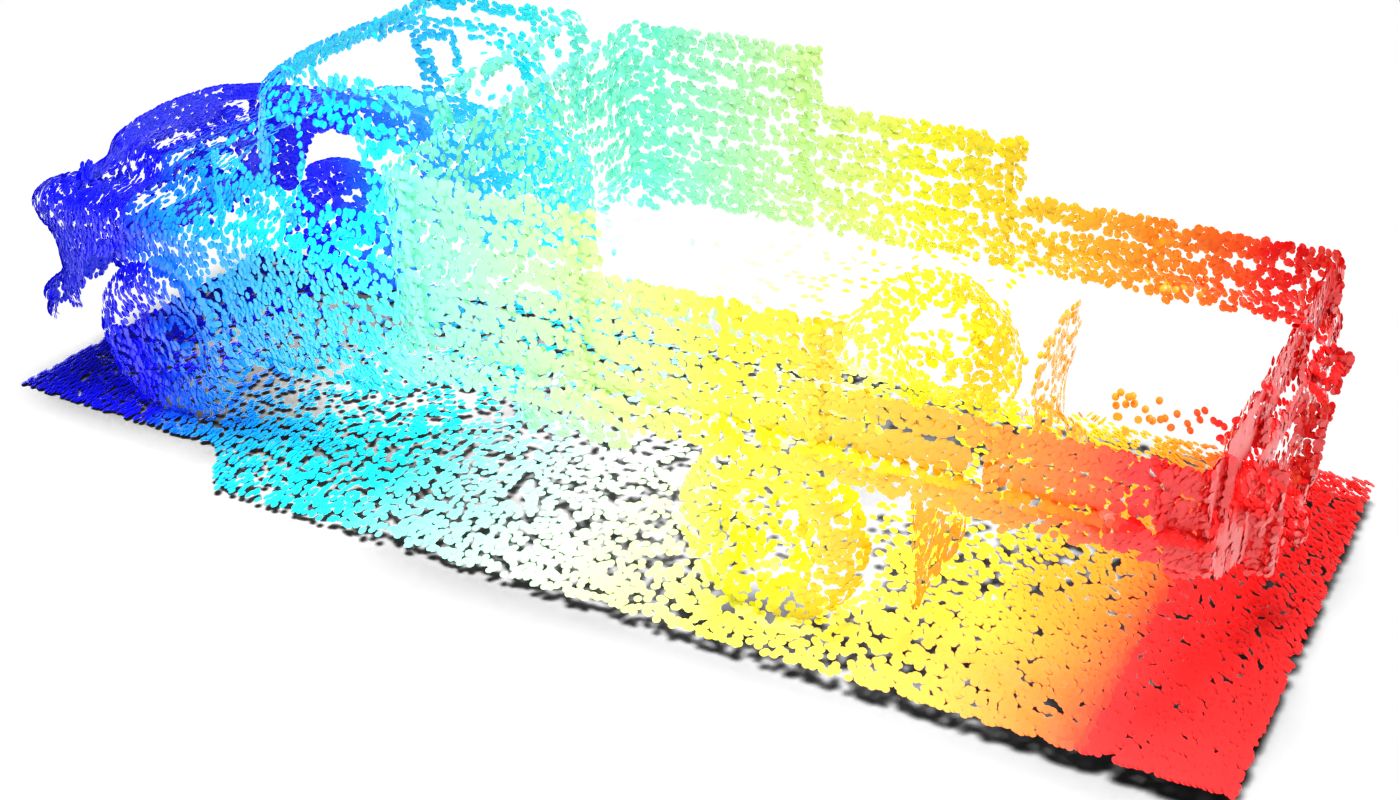}}}}
& \multicolumn{1}{c}{\raisebox{-.5\height}{\Huge --}}
\\           
& \multicolumn{2}{c}{\textbf{\textbf{ShapeNet (synthetic MVS)}}}        &  \multicolumn{2}{c}{\textbf{\textbf{Middlebury (MVS), DTU (MVS), T\&T (range scan)}}}

\end{tabular}
}
\label{ch2:tab:benchmark}
\end{table*}
{We  design a series of experiments (\tabref{ch2:tab:benchmark}) to thoroughly evaluate surface reconstruction models.}

\noindent{\textbf{Evaluation of methods with dataset-driven parameterization.}}
We benchmark both learning-based methods and traditional optimization-based methods that have parameters that can be set according to a given dataset and which can be evaluated on a complete dataset.
\begin{itemize}
    \item \textbf{\hypertarget{e1}{(E1)} In-distribution.}
    We trained models on the ShapeNet training set and test on its test set. All point clouds are generated using the same \ac{mvs} scanning procedure, featuring 3000 points with Gaussian noise (zero mean, standard deviation of 0.5\% of the bounding box diagonal). This experiment assesses the models' ability to handle sparse point clouds, to complete missing data, and to eliminate noise.
    \item \textbf{\hypertarget{e2}{(E2)} Out-of-distribution - unseen point cloud characteristics.} We test models trained in \hyperlink{e1}{E1} on denser point clouds with 10,000 points {with the same 0.5\% Gaussian noise, and with the addition of 10\% outliers}.5
    This tests the {ability of the model to generalize} 
    to different point cloud characteristics.
    \item \textbf{\hypertarget{e3}{(E3)} Out-of-distribution - unseen shape categories: less complex.} 
    We evaluate models from \hyperlink{e1}{E1} on ModelNet categories not present in the ShapeNet training set, maintaining the same point cloud characteristics. This explores generalization to new shape categories in an easy setting.
    \item \textbf{\hypertarget{e4}{(E4)} Out-of-distribution - unseen shape categories: more complex.} 
    We train the learning methods on the simpler shapes from ModelNet and evaluate them on the test set of ShapeNet. Here, we assess whether learning methods can generalize from simple shapes to more complex ones, a difficult out-of-distribution setting.
\end{itemize}

\noindent{\textbf{Evaluation of methods without dataset-driven parameterization.}}
{As the tested neural optimization-based methods are significantly slower than learning-based and traditional optimization-based methods (several minutes \textit{v.s} a few seconds for a single shape), we had to evaluate them separately, on a much smaller dataset than the ShapeNet or ModelNet test sets (which contain several thousands shapes).}
\begin{itemize}
    \item \textbf{\hypertarget{e5}{(E5)} Shape variety with defects.} We evaluate both neural and traditional optimization-based methods that do not require parameterization based on a training set on Berger \etal's synthetic range-scanning dataset.
\end{itemize}

\noindent{\textbf{Evaluation of all  methods on a common ground.} Last, we evaluate all types of methods on the same, small-enough datasets, featuring both synthetic and real point clouds.}
\begin{itemize}
    \item \textbf{\hypertarget{e6}{(E6)} Synthetic data with ground truth (quantitative evaluation).} We compare learning- and optimization-based methods for the \ac{mvs}-scanned version of five shapes from Berger \etal's dataset, which do not correspond to any of ShapeNet's categories but are comparable in complexity. We use the models trained in \hyperlink{e1}{E1}.
    \item \textbf{\hypertarget{e7}{(E7)} Real data without ground truth (qualitative evaluation).} 
    Finally, we compare the learning-based network trained in \hyperlink{e1}{E1} to optimization-based methods on real point clouds obtained by \ac{mvs} and range scanning. As these real shapes do not have a ground truth surface, we only perform qualitative evaluation.
\end{itemize}

\begin{figure}[t]

\captionsetup[sub]{labelfont=scriptsize,textfont=scriptsize,justification=centering}
    \centering
    \newcommand{\mywidth}{0.49\columnwidth}
    \begin{tabular}{@{}c@{}c@{}}
        \subfloat[Synthetic MVS]{\includegraphics[width=\mywidth]{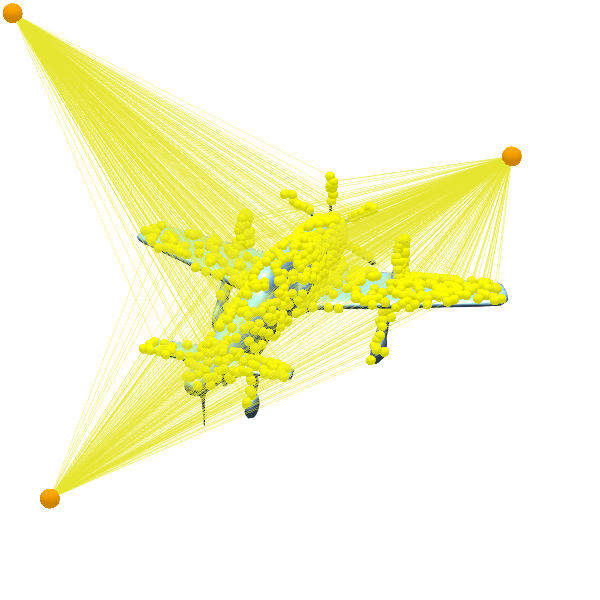}\label{fig:scanning:mvs}}   
         &  \subfloat[Synthetic range scanning]{\includegraphics[width=\mywidth]{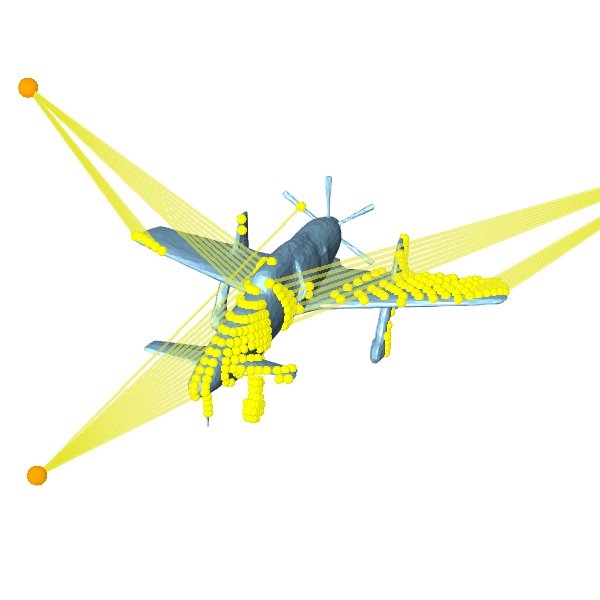}\label{fig:scanning:range}}
    \end{tabular}
    \caption{\textbf{Synthetic scanning:} 
    We randomly place sensors 
    (\protect\tikz[baseline=-.25em] \protect\node[circle, thick, draw = none, fill = orange, scale = 0.5] {};)
    on bounding spheres with multiple radii around an object. To produce MVS-like point clouds, we consider rays 
    (\protect\tikz[baseline=-.25em] \protect\draw[-, thick, draw = yellow, line width = 0.4mm] (0,0) -- (0.5,0);) from sensors aiming at uniformly sampled points on the circumsphere of the object (\subref{fig:scanning:mvs}). This produces non-uniform point clouds with self-occlusions similar to real MVS point clouds. For synthetic range scanning, we use a modified version of Berger \etal's \cite{Berger_benchmark} pipeline, which considers rays arranged on a uniform line grid aiming at the object (\subref{fig:scanning:range}). This produces point clouds with uniformity and self-occlusions similar to real range scanning point clouds.  
}
    \label{fig:scan}
\end{figure}

\subsection{Evaluated surface reconstruction methods}
\label{ch2:sec:methods}

{In this section, we offer a concise overview and some specifics regarding the implementation of the methods evaluated in our benchmark. For a more comprehensive analysis, please refer to our detailed survey presented in \secref{sec:survey}.}

{To ensure a fair evaluation of traditional methods, when running experiments with a training set (E1-E4), we perform a grid-search on the parameters and select the configuration with the best mean volumetric \acs*{iou} on the training set. In the experiments E5-E7, we use default parameters for all optimization methods.}

\vspace{0.1cm}\noindent
{\textbf{Learning-based methods}.
The parameters of these methods are learned on a training set.
\begin{itemize}
\item\textbf{ConvONet~\cite{Peng2020}}
first extracts point features and averages them on cells of three 2D grids (2D variant), or one 3D grid (3D variant). It then uses convolutions to capture the local geometry and predicts the occupancy of query points from interpolated grid features.
\item\textbf{SAP~\cite{Peng2021SAP}}
estimates the oriented normals and $k$ point offsets for each input point to densify the point cloud. The resulting point cloud of size $k\mid\cP\mid$ is then used by a differentiable Poisson solver \cite{screened_poisson} to predict a mesh. The entire pipeline is trained end-to-end.
\item\textbf{DGNN~\cite{dgnn}}
uses a graph neural network to predict the occupancy of Delaunay cells. 
\item\textbf{POCO~\cite{boulch2022poco}} extracts point features using point cloud convolution \cite{boulch2020convpoint}, then estimates the occupancy of query points with a learning-based interpolation method.
\end{itemize}}

\vspace{0.1cm}
\noindent\textbf{Neural optimization-based methods.} These methods are optimized for each input shape independently, and do not rely on any training set.
\begin{itemize}
    \item \textbf{IGR~\cite{Gropp2020}}  learns an implicit function for an input cloud based on its points' position and normals. We train the model for $100,000$ iterations for each scan.
    \item \textbf{LIG~\cite{lig}}  trains an autoencoder from ground truth \acl{sdf} and only retains its decoder for inference.  As the training code is unavailable, we use the available decoder pretrained on ShapeNet (without noise). The input point clouds are augmented with $10$ new points along each point's jet-estimated normals \cite{jet-normal}, whose occupancy is determined by the normals orientation. We use the proposed post-processing to remove falsely enclosed volumes
    \item\textbf{P2M~\cite{point2mesh}}
    iteratively moves the vertices of an initial mesh to fit a point cloud.
    \item{\textbf{\ac{sapopt}~\cite{Peng2021SAP}} is the variant of \ac{sap} that is purely optimization-based, rather than learning-based. It}
uses non-learned normals and densification. 
\end{itemize}
Note that while LIG employs a pretrained network, it also involves optimizing a unique function for each new input point cloud. Consequently, for the purposes of our analysis and classification, we categorize LIG as an optimization-based method, acknowledging its hybrid nature.

\vspace{0.1cm}
\noindent\textbf{Traditional optimization-based methods.} These test-of-time  methods do not rely on neural networks {or training, but nonetheless have a few parameters, which can be set according to default values or adjusted depending on given ground-truth data (akin to a train set)}.
\begin{itemize}
\item\textbf{SPSR~\cite{screened_poisson}}
is a classic traditional method which approximates the surface as a level-set of an implicit function estimated from point positions and normal information. We use the sensor position to orient jet-estimated normals \cite{jet-normal}.
In experiments E1-E4, we perform a grid-search on the train set with octree depths of $\{6,8,10,12\}$, and boundary conditions in $\{\text{dirichlet, neumann, free}\}$. 
In experiments E5-E7,
we use an octree of depth $10$ and Dirichlet boundary condition.
We also experimented with post-processing the reconstructed surface with the trimming tool provided by the authors but could not find parameters that consistently improve the evaluation metrics.
\item\textbf{\ac{resr}~\cite{Labatut2009a}} is a classical graph-cut-based method exploiting on visibility information. 
For experiments E1-E4, we perform a grid-search on the train set with point weights $\alpha = \{16, 32, 48\}$ and $\sigma = \{0.001,0 .01, 0.1, 1\}$, and regularization strength $\lambda = \{1.5, 2.5, 5, 10\}$.
In the optimization setting, we
use the parametrization suggested by the authors: point weights $\alpha_{vis} = 32$ and $\sigma = 0.01$ and regularization strength $\lambda = 5$.
\end{itemize}

\subsection{Evaluation metrics}
\label{ch2:sec:metrics}

We want the reconstructed surface $\cSr$ to be as close as possible to the  ground-truth surface $\cS$ in terms of geometry and topology. To measure this \emph{closeness} we use various metrics described below.

\vspace{0.1cm}
\noindent{\textbf{Geometric metrics.}
We evaluate reconstructions
with the volumetric \ac{iou}, symmetric \ac{cd}, and \ac{nc}:
\begin{itemize}
\item\textbf{Volumetric IoU:}  We denote by $\cVg$ the space  inside the ground-truth surface $\cSg$, and $\cVr$ the space inside the reconstructed surface $\cVr$.
The volumetric \ac{iou} estimates the intersection over union between $\cVg$ and $\cVr$. To approximate this value, we sample a set $\cP$ of $100,000$ points in the union of the bounding boxes of $\cVg$ and $\cVr$ and calculate the following:
\begin{align*}
\text{IoU}\left(\cSg,\cSr\right) =
    &\frac{\vert \{ p \in \cP  \cap \cVg \cap \cVr \}\vert}{\vert \{p \in \cP \cap (\cVr \cup  \cVg) \vert}~.
\end{align*}
\item\textbf{Chamfer distance:}
We sample a set of points $\cPg$ and $\cPr$ on the facets of the ground-truth mesh and the reconstructed mesh, respectively, with $\vert \cPg \vert = \vert \cPr \vert = 100,000$.
We approximate the symmetric Chamfer distance between $\cSg$ and $\cSr$ as follows:
\begin{align*}\nonumber
    \text{CD}(\cSg,\cSr) =
    & \frac{1}{2\vert \cPg \vert} \sum_{x \in \cPg} \min_{y \in \cPr} {\vert \vert x - y \vert \vert }_2 \\
    + & \frac{1}{2\vert \cPr \vert}\sum_{y \in \cPr} \min_{x \in \cPg} {\vert \vert y - x \vert \vert }_2~.
\end{align*}
\item\textbf{Normal consistency:}
Let $n(x)$ be the unit normal of a point $x$ on a mesh, and  $\langle\cdot{,}\cdot\rangle$ the Euclidean scalar product in $\bR^3$.
The normal consistency is estimated as:
\resizebox{\linewidth}{!}{
  \begin{minipage}{\linewidth}
\begin{align*}\nonumber
    \text{NC}(\cSg,\cSr) =
    & \frac{1}{2\vert \cPg \vert} \sum_{x \in \cPg} \left\langle n(x),n\left(\argmin_{y \in \cPr} {\vert \vert x - y \vert \vert }_2\right) \right\rangle  \\
    + & \frac{1}{2\vert \cPr \vert}\sum_{y \in \cPr} \left\langle n(y),n\left(\argmin_{x \in \cPg} {\vert \vert y - x \vert \vert }_2\right) \right \rangle~.
\end{align*}
\end{minipage}}
\end{itemize}}

\vspace{0.1cm}
\noindent{\textbf{Topologic metrics.}
All surfaces considered in our benchmark have a single component, and are closed and manifold. In consequence, we can measure the geometric quality of the reconstructed surfaces with the following indices:
\begin{itemize}
\item {\bf Number of components:} The reconstructed surfaces should also have one component. 
\item {\bf Number of boundary edges:}  All reconstructed meshes should have no boundary edges, \ie edges that belong to only one facet.
\item {\bf Number of non-manifold edges:} All the edges of the reconstructed meshes should be manifold.
\end{itemize}}
\section{Experiments}
\label{ch2:experiments}

\noindent{We {report here the results of the} 
series of experiments {presented in \secref{ch2:sec:setup} and} designed to assess the robustness and performance of various deep and non-deep surface reconstruction methods from point clouds (\cf\tabref{ch2:tab:benchmark}). We start in \secref{sec:e1E4}, {for methods with dataset-driven parameterization}, by measuring the impact of {various inconsistencies} 
between train and test sets. We then evaluate in \secref{sec:E5} the performance of 
optimization-based methods {without dataset-driven parameterization}. Finally, we compare in \secref{sec:E6E7} {all evaluated methods, both} learning- and optimization-based, 
on synthetic and real point acquisitions.}

\subsection{{Surface reconstruction with dataset-driven parameterization (E1 - E4)}}
\label{sec:e1E4}

\begin{table*}
\caption{\textbf{Comparison of methods with dataset-driven parameterization (\protect\hyperlink{e1}{E1} to \protect\hyperlink{e4}{E4}):}
We evaluate across 6 metrics various learning-based methods and traditional optimization-based methods with adjustable parameters.
 We observe that learning-based methods outperform traditional methods in the simple setting where the training and testing set have the same characteristics (\protect\hyperlink{e1}{E1}). However, non-learning methods perform better when the test set contains unseen point defects or more complex objects. $^\dagger$methods not based on neural networks.}
\centering
\resizebox{0.9\textwidth}{!}{
\begin{tabular}{lcccccc|ccccc}
\toprule
& & \multicolumn{5}{c|}{\textbf{Volumetric IoU} (\%)~~[$\uparrow$]}
& \multicolumn{5}{c}{\textbf{Normal consistency} (\%)~~[$\uparrow$]}
\\
\textbf{Method} & &  \multicolumn{1}{c}{E1} & \multicolumn{1}{c}{E2} &  \multicolumn{1}{c}{E3} &  \multicolumn{1}{c}{E4} &   Mean &  \multicolumn{1}{c}{E1} & \multicolumn{1}{c}{E2} & \multicolumn{1}{c}{E3} &  \multicolumn{1}{c}{E4} & Mean \\
\midrule
\textbf{ConvONet2D   } &          \cite{Peng2020} &            85.0 &          47.3 &          79.3 &        68.3 &            70.0 &          92.7 &          76.4 &            90.0 &        87.8 &            86.7 \\

\rowcolor{gray!10}\textbf{ConvONet3D   } &          \cite{Peng2020} &          84.8 &          15.1 &          83.6 &               51.0 &          58.6 &            93.0 &          71.8 &          93.1 &              82.5 &          85.1 \\
\textbf{SAP          } &       \cite{Peng2021SAP} &          88.7 &          59.8 &          89.2 &             54.9 &          73.2 &          93.5 & \textbf{86.7} &          94.1 &             87.1 & \textbf{90.4} \\
\rowcolor{gray!10}\textbf{DGNN} &              \cite{dgnn} &          84.5 &          38.1 &            87.0 &  \textbf{84.4} &          73.5 &          85.4 &          68.8 &          88.5 &          85.5 &          82.1 \\
\textbf{POCO         } &    \cite{boulch2022poco} & \textbf{89.5} &          8.74 & \textbf{90.6} &        40.9 &          57.4 & \textbf{93.6} &          75.6 & \textbf{94.2} &        82.9 &          86.6\\
\rowcolor{gray!10}$^\dagger$\textbf{SPSR         } &  \cite{screened_poisson} &          77.1 & \textbf{80.7} &          80.7 &           74.6 & \textbf{78.3} &          87.7 &          83.2 &          89.1 & \textbf{87.0} &          86.9 \\
$^\dagger$\textbf{RESR} &      \cite{Labatut2009a} &          80.3 &          60.4 &          83.9 &           80.3 &          76.2 &            81.0 &            73.0 &          84.6 &        81.0 &          79.9 \\
\midrule
& & \multicolumn{5}{c|}{\textbf{Chamfer distance} ($\times 10^3$) ~~[$\downarrow$]}
& \multicolumn{5}{c}{\textbf{Number of components} ~~[$\downarrow$]}
\\
\textbf{Method} &  &             \multicolumn{1}{c}{E1} &             \multicolumn{1}{c}{E2} &             \multicolumn{1}{c}{E3} &             \multicolumn{1}{c}{E4} &                 Mean &            \multicolumn{1}{c}{E1} &            \multicolumn{1}{c}{E2} &            \multicolumn{1}{c}{E3} &         \multicolumn{1}{c}{E4}  &         Mean \\
\midrule
\textbf{ConvONet2D   } &          \cite{Peng2020} &          0.55 &           7.51 &          1.00 &          0.98 &           2.51 &           1.6 &          34.8 &          2.6&           3.2 &         10.6 \\
\rowcolor{gray!10}\textbf{ConvONet3D   } &          \cite{Peng2020} &          0.55 &           10.90 &           0.76 &        2.44 &            3.66 &          1.4 &          13.6 &           1.6 &            1.5 &         4.5 \\
\textbf{SAP          } &       \cite{Peng2021SAP} &          0.44 &           2.09 &          0.55 &      0.92 &          1.00 &          2.7 &            86.0 &          3.5 &        10.5 &         25.7 \\
\rowcolor{gray!10}\textbf{DGNN         } &              \cite{dgnn} &          0.55 &           2.54 &          0.63 &    \textbf{0.55} &          1.07 &          1.3 &          16.1 &          1.1 &          1.3 &         5.0 \\
\textbf{POCO         } &    \cite{boulch2022poco} & \textbf{0.42} &           10.50 & \textbf{0.52} &          1.32 &           3.19 &          2.3 &           178 &          2.8 &            16.3 &         49.9 \\
\rowcolor{gray!10}$^\dagger$\textbf{SPSR         } &  \cite{screened_poisson} &          0.80 & \textbf{0.66} &          0.87 &        0.89 & \textbf{0.81} &          9.3 &           185 &          11.1 &              3.2 &         52.0 \\
$^\dagger$\textbf{RESR} &      \cite{Labatut2009a} &          0.67 &           6.97 &          0.75 &           0.67 &           2.27 & \textbf{1.2} & \textbf{9.0} & \textbf{1.0}& \textbf{1.2} & \textbf{3.1} \\
\midrule
& & \multicolumn{5}{c|}{\textbf{Number of boundary edges} ~~[$\downarrow$]}
& \multicolumn{5}{c}{\textbf{Number of non-manifold edges}~~[$\downarrow$]}
\\
\textbf{Method} &        &         \multicolumn{1}{c}{E1} &         \multicolumn{1}{c}{E2} &         \multicolumn{1}{c}{E3} &         \multicolumn{1}{c}{E4} &         Mean &         \multicolumn{1}{c}{E1} &         \multicolumn{1}{c}{E2} &         \multicolumn{1}{c}{E3} &         \multicolumn{1}{c}{E4} &        Mean \\
\midrule
\textbf{ConvONet2D   } &          \cite{Peng2020} & \textbf{0} & \textbf{0} & \textbf{0} & \textbf{0} & \textbf{0} & \textbf{0} & \textbf{0} & \textbf{0}  & \textbf{0} & \textbf{0}  \\
\rowcolor{gray!10}\textbf{ConvONet3D   } &          \cite{Peng2020} & \textbf{0} & \textbf{0} &  \textbf{0} & \textbf{0} & \textbf{0} & \textbf{0} & \textbf{0} & \textbf{0} & \textbf{0} & \textbf{0} \\
\textbf{SAP          } &       \cite{Peng2021SAP} & \textbf{0} &  \textbf{0} & \textbf{0} &       8.4 &       2.1 & \textbf{0} & \textbf{0} & \textbf{0} & \textbf{0} & \textbf{0} \\
\rowcolor{gray!10}\textbf{DGNN         } &              \cite{dgnn} & \textbf{0} & \textbf{0} &   \textbf{0} & \textbf{0} & \textbf{0} &       1.4 &       2.2 &      0.6  &       1.7 &       1.5 \\
\textbf{POCO         } &    \cite{boulch2022poco} & \textbf{0} &        121 & \textbf{0} &     41.7 &       40.7 & \textbf{0} &    \textbf{0} & \textbf{0} & \textbf{0} & \textbf{0}  \\
\rowcolor{gray!10}$^\dagger$\textbf{SPSR         } &  \cite{screened_poisson} & \textbf{0} & \textbf{0} & \textbf{0} & \textbf{0} & \textbf{0} & \textbf{0} & \textbf{0} & \textbf{0} & \textbf{0} & \textbf{0}  \\
$^\dagger$\textbf{RESR} &      \cite{Labatut2009a} & \textbf{0} & \textbf{0} & \textbf{0}  & \textbf{0} & \textbf{0} &       9.4 &       28.5 &       8.47 &       9.4 &       13.9 \\
\bottomrule
\if 1 0
\fi
\end{tabular}
}
\label{tab:all_learning_experiments}
\end{table*}

\newcommand{\fpath}[3]{%
\includegraphics[width=\mywidth,mytrim]{figures/learning_results/#1/#2/#3}}
\def \shapenet {02691156/d18592d9615b01bbbc0909d98a1ff2b4}
\def \modelnet {bed/0585}
\def \reconbench {daratech/daratech}
\begin{figure*}
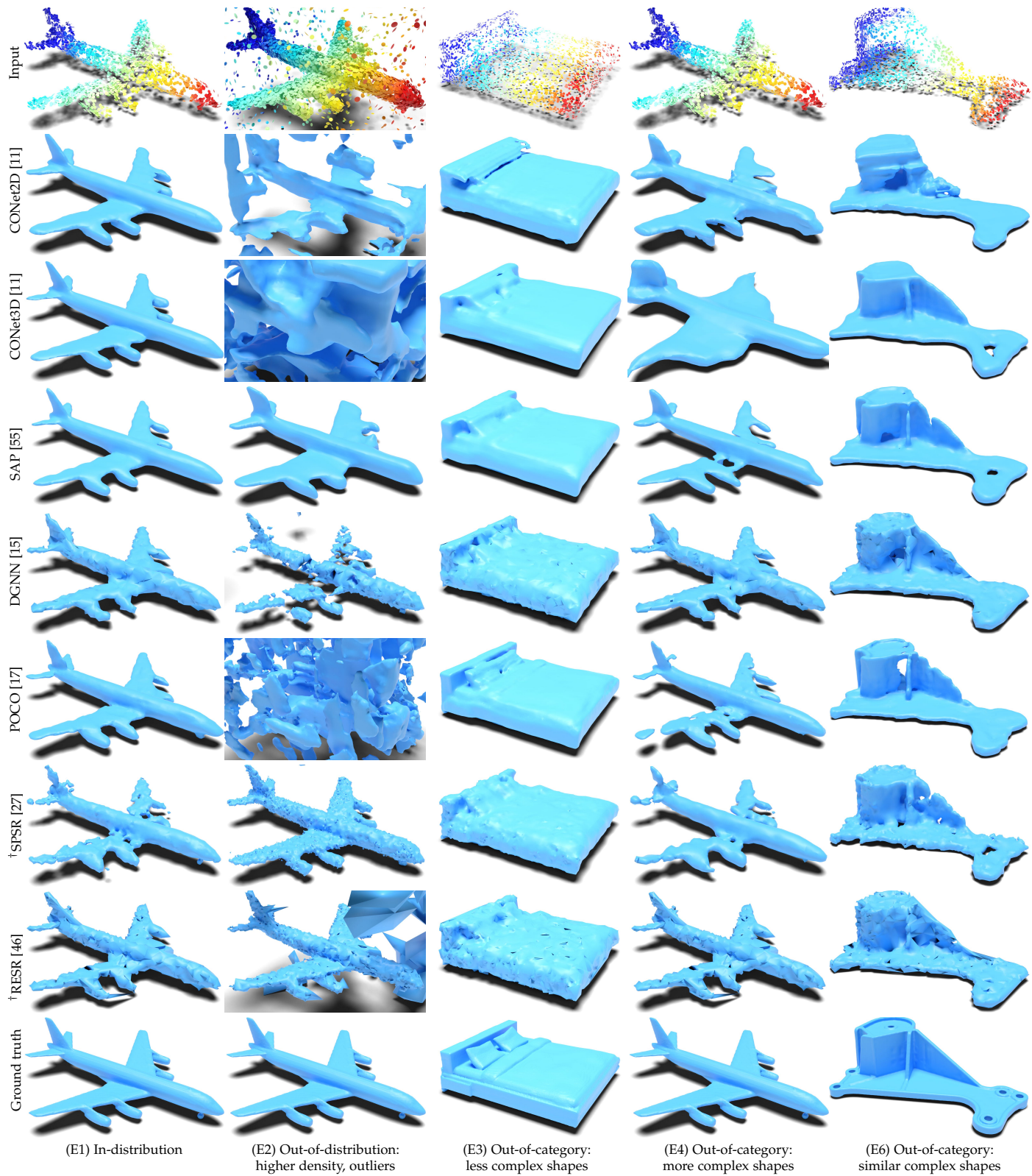

	\centering\scriptsize 
	\newcommand{\mywidth}{0.20\textwidth}
	\definetrim{mytrim}{70 10 70 30}
\begin{tabular}{@{}c@{}@{}c@{}c@{}c@{}c@{}c}

\rotatebox{90}{\hspace{9mm}Input}&
\fpath{shapenet3000}{\shapenet}{input.jpg}&
\fpath{shapenet10000}{\shapenet}{input.jpg}&
\fpath{modelnet}{\modelnet}{input.jpg}&
\fpath{shapenet}{\shapenet}{input.jpg}&
\fpath{reconbench}{\reconbench}{input.jpg}
\\

\rotatebox{90}{\hspace{4mm}CONet2D~\cite{Peng2020}}&
\fpath{shapenet3000}{\shapenet}{ConvONet2D.jpg}&
\fpath{shapenet10000}{\shapenet}{ConvONet2D.jpg}&
\fpath{modelnet}{\modelnet}{ConvONet2D.jpg}&
\fpath{shapenet}{\shapenet}{ConvONet2D.jpg}&
\fpath{reconbench}{\reconbench}{ConvONet2D.jpg}
\\

\rotatebox{90}{\hspace{4mm}CONet3D~\cite{Peng2020}}&
\fpath{shapenet3000}{\shapenet}{ConvONet3D.jpg}&
\fpath{shapenet10000}{\shapenet}{ConvONet3D.jpg}&
\fpath{modelnet}{\modelnet}{ConvONet3D.jpg}&
\fpath{shapenet}{\shapenet}{ConvONet3D.jpg}&
\fpath{reconbench}{\reconbench}{ConvONet3D.jpg}
\\

\rotatebox{90}{\hspace{5mm}SAP~\cite{Peng2021SAP}}&
\fpath{shapenet3000}{\shapenet}{SAP.jpg}&
\fpath{shapenet10000}{\shapenet}{SAP.jpg}&
\fpath{modelnet}{\modelnet}{SAP.jpg}&
\fpath{shapenet}{\shapenet}{SAP.jpg}&
\fpath{reconbench}{\reconbench}{SAP.jpg}
\\

\rotatebox{90}{\hspace{5mm}DGNN~\cite{dgnn}}&
\fpath{shapenet3000}{\shapenet}{DGNN.jpg}&
\fpath{shapenet10000}{\shapenet}{DGNN.jpg}&
\fpath{modelnet}{\modelnet}{DGNN.jpg}&
\fpath{shapenet}{\shapenet}{DGNN.jpg}&
\fpath{reconbench}{\reconbench}{DGNN.jpg}
\\

\rotatebox{90}{\hspace{5mm}POCO~\cite{boulch2022poco}}&
\fpath{shapenet3000}{\shapenet}{POCO.jpg}&
\fpath{shapenet10000}{\shapenet}{POCO.jpg}&
\fpath{modelnet}{\modelnet}{POCO.jpg}&
\fpath{shapenet}{\shapenet}{POCO.jpg}&
\fpath{reconbench}{\reconbench}{POCO.jpg}
\\

\rotatebox{90}{\hspace{5mm}$^\dagger$SPSR~\cite{screened_poisson}}&
\fpath{shapenet3000}{\shapenet}{SPSR.jpg}&
\fpath{shapenet10000}{\shapenet}{SPSR.jpg}&
\fpath{modelnet}{\modelnet}{SPSR.jpg}&
\fpath{shapenet}{\shapenet}{SPSR.jpg}&
\fpath{reconbench}{\reconbench}{SPSR.jpg}
\\

\rotatebox{90}{\hspace{2mm}$^\dagger$RESR~\cite{Labatut2009a}}&
\fpath{shapenet3000}{\shapenet}{Labatut_rt_2.jpg}&
\fpath{shapenet10000}{\shapenet}{Labatut_rt_2.jpg}&
\fpath{modelnet}{\modelnet}{Labatut_rt_2.jpg}&
\fpath{shapenet}{\shapenet}{Labatut_rt_2.jpg}&
\fpath{reconbench}{\reconbench}{Labatut_rt_2.jpg}
\\

\rotatebox{90}{\hspace{5mm}Ground truth}&
\fpath{shapenet3000}{\shapenet}{gt.jpg}&
\fpath{shapenet10000}{\shapenet}{gt.jpg}&
\fpath{modelnet}{\modelnet}{gt.jpg}&
\fpath{shapenet}{\shapenet}{gt.jpg}&
\fpath{reconbench}{\reconbench}{gt.jpg}
\\

&
(\hyperlink{e1}{E1}) In-distribution &
(\hyperlink{e2}{E2}) Out-of-distribution: &
(\hyperlink{e3}{E3}) Out-of-category: &
(\hyperlink{e4}{E4}) Out-of-category: &
(\hyperlink{e6}{E6}) Out-of-category:
\\ &
& higher density, outliers
& less complex shapes
& more complex shapes
& similar complex shapes
\end{tabular}
	\caption{\textbf{Learning-based and traditional methods (\protect\hyperlink{e1}{E1}-\protect\hyperlink{e4}{E4}, \protect\hyperlink{e6}{E6}):}  
    DGNN \cite{dgnn}, SAP \cite{Peng2021SAP} and \ac{spsr} \cite{screened_poisson} provide visually the best reconstructions, without prevalent defects. $^\dagger$non deep-learning-based methods.
	}
	\label{fig:learning_results}
\end{figure*}

We examine the precision and versatility of recent supervised learning-based methods, as well as two traditional optimization-based methods, whose parameters were trained or tuned on a training set.
We report qualitative results in \figref{fig:learning_results}, and quantitative results in \tabref{tab:all_learning_experiments}.

\vspace{0.1cm}\noindent{\textbf{\hyperlink{e1}{E1}: In distribution.}
In this simple setting, learning-based methods demonstrate superior performance compared to traditional approaches, achieving more than a 5\% improvement in volumetric IoU over \ac{spsr} and \ac{resr}. Implicit field-based methods like \ac{poco}, \ac{sap}, and \ac{conet} show particularly strong results. However, \Ac{dgnn} underperforms relative to other learning methods, likely due to the sparsity of the point clouds in this experiment.
}



\vspace{0.1cm}\noindent{\textbf{\hyperlink{e2}{E2}: Out-of-distribution - unseen point {characteristics}.} 
Learning-based methods exhibit a noticeable decline in performance when confronted with point characteristics (higher density, outliers) that were not seen during training. On the other hand, traditional methods, especially \Ac{spsr}, display a robust response to outliers and take advantage of a higher point density. While \ac{sap}'s reconstructions tend to be overly smooth, missing finer details, they are not as severely affected by defects as other learning-based approaches. The technique of \ac{resr} encounters issues due to its lower regularization weight, which was optimized for outlier-free point clouds. A higher regularization might be beneficial in minimizing erroneous components caused by outliers.


\vspace{0.1cm}\noindent{\textbf{\hyperlink{e3}{E3}: Out-of-distribution - unseen classes (simpler).} 
In the task of reconstructing simpler, out-of-category shapes from ModelNet (\hyperlink{e3}{E3}), neural implicit field methods stand out for their visually superior reconstructions. \ac{sap} and \ac{poco} lead in terms of quantitative performance. The interpolating approach \ac{dgnn} outperforms \ac{conet} in this experiment, demonstrating its generalization capabilities.
}



\vspace{0.1cm}\noindent{\textbf{\hyperlink{e4}{E4}: Out-of-distribution - unseen classes (more complex).} 
In this experiment, most methods overfit the  simpler training shapes from ModelNet shapes and subsequently struggle with reconstructing the more complex shapes from ShapeNet. The non-learning-based method \ac{spsr} is also affected as its parameters are tuned on ModelNet. Specifically, the optimalo octree depth for ModelNet reconstructions is $d=8$, as opposed to $d=10$ for ShapeNet, resulting in a decrease in volumetric IoU from $77.1$  in \hyperlink{e1}{E1} to $74.6$ in \hyperlink{e4}{E4}. In contrast, the parameter tuning for \ac{resr} remains consistent across both datasets.
\ac{dgnn} stands out as the only learning-based method that does not exhibit overfitting on the ModelNet dataset, leading to the most successful outcomes both quantitatively and qualitatively. In fact, its performance is on par with its results when directly trained on the ShapeNet dataset. We attribute this behavior to the local nature of \ac{dgnn}, which only consider small subgraphs during training and inference.
}

\vspace{0.1cm}\noindent{\textbf{Methods Performance.} We propose an analysis of the performance of each model:
\begin{itemize}
\item 
\ac{conet} perform best when both the training and test sets share identical point cloud characteristics and shape categories. On more complicated  settings, this method performs below traditional approaches. 
\item
\ac{sap} stands out for its superior reconstructions and robustness against outliers, partly due to its unique capability of explicitly predicting normals. This feature contributes to \ac{sap} achieving the highest mean normal consistency across all experiments.
\item
\ac{dgnn}, with its blend of local learning and global regularization, consistently delivers competitive results in nearly all scenarios, with the notable exception of the outlier-rich environment in \hyperlink{e2}{E2}. Overall, it achieves the highest volumetric IoU amongst learning-based methods when averaged across all experiments. 
\item
\ac{poco}'s local attention-based learning mechanism yields optimal outcomes when reconstruction does not require adapting to unseen domains. It particularly excels in experiments where the point cloud characteristics remain consistent between training and testing (\hyperlink{e1}{E1}, \hyperlink{e3}{E3}, \hyperlink{e4}{E4}). However, \ac{poco} is significantly impacted by outliers, as evidenced in \hyperlink{e2}{E2}, and shows a tendency to overfit on simpler training shapes, as observed in \hyperlink{e4}{E4}. Notably, both \ac{poco} and \ac{sap} produce watertight reconstructions, with boundary edges only appearing where the reconstruction intersects the bounding box.
\item
\Ac{spsr} demonstrates a high degree of robustness to various defects and shape complexities, yielding consistently good results across different metrics. However, it tends to produce less compact reconstructions with a higher number of components. 
\item
\ac{resr}, while slightly less robust to outliers {and now relatively old}, achieves a mean volumetric IoU higher than any of the learning-based methods, and presents the most compact reconstructions with an average of 2.7 components only. However, this  approach produces a significant number of non-manifold edges.
\end{itemize}
}

\subsection{{Surface reconstruction without dataset-driven parameterization (E5)}}

\label{sec:E5}


\begin{table*}
\caption{\textbf{Comparison of methods without dataset-driven parameterization (\protect\hyperlink{e5}{E5}):}
We report the performance of neural and traditional methods that do not train or tune their parameters on a training set. We evaluate on point clouds obtained with different scan settings: 
\textbf{LR:} low resolution, \textbf{HR:} high resolution, \textbf{HRN:} high resolution and noise, \textbf{HRO:} high resolution and outliers, and \textbf{HRNO:} high resolution  with noise and outliers. $^\dagger$methods not based on neural networks.}
\centering
\resizebox{0.9\textwidth}{!}{

\begin{tabular}{l@{~}ccccccc|cccccc}
\toprule
& & \multicolumn{6}{c|}{\textbf{Volumetric IoU} (\%)~~[$\uparrow$]}
& \multicolumn{6}{c}{\textbf{Normal consistency} (\%)~~[$\uparrow$]}
\\
\textbf{Method} &                                   &            LR &            HR &            HRN &            HRO &            HRNO &          Mean &            LR &            HR &            HRN &            HRO &            HRNO &          Mean \\
\midrule
\textbf{IGR          } &         \cite{Gropp2020} &          80.8 &          92.5 & \textbf{83.6} &          63.7 &          62.7 &          76.7 &            88.0 & \textbf{96.3} &        83.9 &          77.8 &          71.5 &          83.5 \\
\rowcolor{gray!10}\textbf{LIG          } &               \cite{lig} &          46.9 &          50.3 &          63.9 &            66 &          63.8 &          58.2 & \textbf{88.7} &          92.2 & \textbf{89.0} &            77 &          75.2 &          84.4 \\
\textbf{P2M          } &        \cite{point2mesh} &          75.2 &          83.3 &          75.5 &          71.3 &          67.8 &          74.6 &          86.3 &          92.2 &        88.1 &          84.5 &          82.1 &          86.6 \\
\rowcolor{gray!10}\textbf{\ac{sapopt}       } &       \cite{Peng2021SAP} &          75.6 &          89.1 &          72.4 &          55.3 &          34.9 &          65.4 &          83.4 &          94.8 &        61.6 &          74.5 &          55.3 &          73.9 \\
 $^\dagger$\textbf{SPSR         } &  \cite{screened_poisson} &          77.7 &          90.2 &          82.8 &          90.3 & \textbf{82.1} &          84.6 &          88.1 &            96 &        88.1 & \textbf{96.2} & \textbf{85.8} & \textbf{90.9} \\
\rowcolor{gray!10} $^\dagger$\textbf{RESR} &      \cite{Labatut2009a} & \textbf{81.3} & \textbf{93.4} &          80.1 & \textbf{93.4} &          79.1 & \textbf{85.5} &          87.6 &            96 &        66.3 &          94.9 &          66.5 &          82.3 \\
\midrule
& & \multicolumn{6}{c|}{\textbf{Chamfer distance} ($\times 10^3$) ~~[$\downarrow$]}
& \multicolumn{6}{c}{\textbf{Number of components} ~~[$\downarrow$]}
\\
\textbf{Method} &                               &             LR &            HR &            HRN &            HRO &            HRNO &           Mean &            LR &            HR &            HRN &            HRO &            HRNO &         Mean \\
\midrule
\textbf{IGR          } &         \cite{Gropp2020} &          0.67 &          0.32 & \textbf{0.55} &           7.96 &           7.72 &           3.45 &          6.8 & \textbf{1.2} &         35.2 &           44 &       97.4 &          36.9 \\
\rowcolor{gray!10}\textbf{LIG          } &               \cite{lig} &          0.75 &          0.58 &          0.78 &           7.89 &            7.80 &           3.56 &   \textbf{1} &   \textbf{1} &   \textbf{1} &          1.6 & \textbf{1} &          1.12 \\
\textbf{P2M          } &        \cite{point2mesh} &          0.82 &          0.47 &          0.73 &           1.53 &           2.13 &           1.13 & \textbf{1.2} &   \textbf{1} & \textbf{1.2} &          1.4 &        1.6 &          1.28 \\
\rowcolor{gray!10}\textbf{\ac{sapopt}} &       \cite{Peng2021SAP} &          0.85 &           0.32 &          0.70 &           3.99 &           3.93 &           1.96 &         73.2 &         85.6 &          937 &      1800 &   1.60 &           971 \\
 $^\dagger$\textbf{SPSR         } &  \cite{screened_poisson} &          0.79 &          0.37 &          0.57 &          0.36 & \textbf{0.61} &          0.54 & \textbf{1.2} &          1.6 &          3.6 &          3.8 &       20.2 &          6.08 \\
\rowcolor{gray!10}$^\dagger$\textbf{RESR} &       \cite{Labatut2009a} & \textbf{0.64} & \textbf{0.31} &          0.61 & \textbf{0.34} &          0.64 & \textbf{0.51} &   \textbf{1} &   \textbf{1} & \textbf{1.2} & \textbf{1.2} & \textbf{1} & \textbf{1.08} \\
\midrule
& & \multicolumn{6}{c|}{\textbf{Number of boundary edges} ~~[$\downarrow$]}
& \multicolumn{6}{c}{\textbf{Number of non-manifold edges}~~[$\downarrow$]}
\\
\textbf{Method} &        &         LR &            HR &            HRN &            HRO &            HRNO &       Mean &         LR &            HR &            HRN &            HRO &            HRNO &       Mean \\
\midrule
\textbf{IGR          } &         \cite{Gropp2020} & \textbf{0} & \textbf{0} & \textbf{0} & \textbf{0} & \textbf{0} & \textbf{0} & \textbf{0} &        0.8 &        0.8 &        5.2 &        4.2 &        2.2 \\
\rowcolor{gray!10}\textbf{LIG          } &               \cite{lig} &         69 &       42.8 &       17.2 & \textbf{0} & \textbf{0} &       25.8 & \textbf{0} & \textbf{0} & \textbf{0} & \textbf{0} & \textbf{0} & \textbf{0} \\
\textbf{P2M          } &        \cite{point2mesh} & \textbf{0} & \textbf{0} & \textbf{0} & \textbf{0} & \textbf{0} & \textbf{0} & \textbf{0} & \textbf{0} & \textbf{0} & \textbf{0} & \textbf{0} & \textbf{0} \\
\rowcolor{gray!10}\textbf{\ac{sapopt}          } &       \cite{Peng2021SAP} & \textbf{0} & \textbf{0} & \textbf{0} & \textbf{0} &        449 &       89.8 & \textbf{0} & \textbf{0} & \textbf{0} & \textbf{0} & \textbf{0} & \textbf{0} \\
$^\dagger$\textbf{SPSR         } &  \cite{screened_poisson} & \textbf{0} & \textbf{0} & \textbf{0} & \textbf{0} & \textbf{0} & \textbf{0} & \textbf{0} & \textbf{0} & \textbf{0} & \textbf{0} & \textbf{0} & \textbf{0} \\
\rowcolor{gray!10}$^\dagger$\textbf{RESR} &     \cite{Labatut2009a} & \textbf{0} & \textbf{0} & \textbf{0} & \textbf{0} & \textbf{0} & \textbf{0} &          1 &        5.8 &       24.4 &        3.8 &         22 &       11.4 \\
\bottomrule
\end{tabular}
}
\label{tab:all_optim_experiments}
\end{table*}

\vspace{0.1cm}{\textbf{Setting.} We now evaluate  methods that do not learn or tune their parameters on a training set. We consider neural-network-based optimization approaches that learn a new function to fit each input point cloud. We also assess the performance of non-learning-based methods, using their default parameters for a fair evaluation. These methods are tested on the benchmark shapes from Berger et al. \cite{Berger_benchmark}, {which uses synthetic range scans} 
with varying resolutions, noise levels, and outlier proportions.}

\vspace{0.1cm}\noindent{\textbf{Analysis.}
We report the results of this experience in \tabref{tab:all_optim_experiments}.
Interestingly, traditional methods such as \ac{spsr} and \ac{resr} consistently surpass network-based approaches in most metrics and scenarios. Their superior performance is particularly evident when dealing with outliers, where they exhibit over a 10-point advantage in volumetric IoU and normal consistency. \Ac{spsr} provides accurate and robust normal estimation under various conditions. \Ac{resr} achieves the highest mean \ac{iou} and mean Chamfer distance, and yields the most compact reconstructions. However, it also results in a significant number of non-manifold edges.}

{
\Ac{igr} stands out among learning-based methods, particularly when processing outlier-free point clouds. It even surpasses traditional methods in terms of volumetric IoU and Chamfer distance in high-resolution and noisy scenarios (HRN). On the other hand, \Ac{lig} demonstrates generally weaker performance, likely due to its dependency on a model pretrained on defect-free  and high-density point clouds. Its post-processing approach also leads to non-watertight reconstructions. \Ac{p2m} delivers reconstructions that are geometrically sound and topologically superior, characterized by fewer components and consistently producing watertight and manifold surfaces. \Ac{sapopt}, although providing satisfactory reconstructions in the absence of outliers, struggles in more challenging scenarios.
}

\subsection{Learning- and optimization-based surface reconstruction from synthetic MVS point clouds (E6)}
\label{sec:E6E7}

\begin{table*}[t]
\caption{\textbf{Comparison of all methods - synthetic data with ground truth (\protect\hyperlink{e6}{E6}):}
We report the performance of learning-based models trained on ShapeNet, optimization methods without training, and traditional optimization approaches, on shapes of Berger \etal \cite{Berger_benchmark} scanned with our MVS procedure. $^\dagger$methods not based on neural networks.}
\centering
\resizebox{0.9\textwidth}{!}{
\begin{tabular}{ll@{}ccccccc}
& &   
& \multicolumn{1}{c}{  \textbf{Vol. IoU}}
& \multicolumn{1}{c}{ \textbf{Normal consist.}}
& \multicolumn{1}{c}{ \textbf{Chamfer dist.}}
&\multicolumn{1}{c}{ \textbf{Components}} 
&\multicolumn{1}{c}{ \textbf{BE} }
&\multicolumn{1}{c}{ \textbf{NME}}\\
\multicolumn{2}{l}{\textbf{Method}} &   
& \multicolumn{1}{c}{ (\%) ~~[$\uparrow$]}
& \multicolumn{1}{c}{ (\%) ~~[$\uparrow$]}
& \multicolumn{1}{c}{  ($\times 10^3$) ~~[$\downarrow$]}
& \multicolumn{1}{c}{ [$\downarrow$]}
& \multicolumn{1}{c}{ [$\downarrow$]}
& \multicolumn{1}{c}{[$\downarrow$]}\\
\midrule
\multirow{5}{*}{
\rotatebox{90}{\textbf{Learning}}}&
\textbf{ConvONet2D} &          \cite{Peng2020} &          65.1 &            78.0 &           1.43 &        3.6 &\textbf{0} &\textbf{0} \\ 
&\cellcolor{gray!10}\textbf{ConvONet3D} ~~ &           \cellcolor{gray!10}\cite{Peng2020} &   \cellcolor{gray!10}76.4 &\cellcolor{gray!10}87.2 &\cellcolor{gray!10}0.89 &\cellcolor{gray!10}2.6 &\cellcolor{gray!10}\textbf{0} &\cellcolor{gray!10}\textbf{0} \\
&\textbf{SAP          } &       \cite{Peng2021SAP} &          78.3 &            89.0 &          0.73 &        5.6 & \textbf{0} & \textbf{0} \\
&\cellcolor{gray!10}\textbf{DGNN}& \cellcolor{gray!10}\cite{dgnn} &\cellcolor{gray!10}82.9 &\cellcolor{gray!10}85.2 &\cellcolor{gray!10}0.59 &\cellcolor{gray!10}\textbf{1} &\cellcolor{gray!10}\textbf{0} &\cellcolor{gray!10}0.4 \\
&\textbf{POCO         } &    \cite{boulch2022poco} & \textbf{83.9} & \textbf{89.5} & \textbf{0.58} &          2 & \textbf{0} & \textbf{0} \\
\midrule
\multirow{6}{*}{
\rotatebox{90}{\textbf{Optimization}}}&
\textbf{IGR          } &         \cite{Gropp2020} &          78.3 &          83.8 &          0.78 &       15.4 & \textbf{0} &        0.4 \\
&\cellcolor{gray!10}\textbf{LIG          } & \cellcolor{gray!10}\cite{lig} &\cellcolor{gray!10}45.7 &\cellcolor{gray!10}\underline{86.6} &\cellcolor{gray!10}0.83 &\cellcolor{gray!10}\textbf{1} &\cellcolor{gray!10}65.6 &\cellcolor{gray!10}\textbf{0} \\
&\textbf{P2M          } &        \cite{point2mesh} &          74.5 &            85.0 &          0.77 &          2 &\textbf{0} &\textbf{0} \\
&\cellcolor{gray!10}\textbf{\ac{sapopt}} &\cellcolor{gray!10}\cite{Peng2021SAP} &\cellcolor{gray!10}71.9 &\cellcolor{gray!10}77.0 &\cellcolor{gray!10}0.81 &\cellcolor{gray!10}133 &\cellcolor{gray!10}\textbf{0} &\cellcolor{gray!10}\textbf{0} \\
&$^\dagger$\textbf{SPSR} &  \cite{screened_poisson} &          77.6 &          86.4 &          0.79 &          8 &\textbf{0} &\textbf{0} \\
&\cellcolor{gray!10}$^\dagger$\textbf{RESR} &\cellcolor{gray!10}\cite{Labatut2009a} &\cellcolor{gray!10}\underline{79.4} &\cellcolor{gray!10}80.8 &\cellcolor{gray!10}\underline{0.67} &\cellcolor{gray!10}\textbf{1} &\cellcolor{gray!10}\textbf{0} &\cellcolor{gray!10}9.6 \\
\bottomrule
\end{tabular}
}
\label{tab:learning_optim_experiments}
\end{table*}

{
In this other experiment, we compare learning- and optimization-based reconstruction  methods on the same dataset. We scan the shapes of Berger \etal \cite{Berger_benchmark} with the same synthetic scanning method as in \hyperlink{e5}{E5}.
We consider learning-based models trained on shapes from ShapeNet scanned with the same procedure,  network-based optimization models without training, and traditional optimization approaches.  
We report the results in \tabref{tab:learning_optim_experiments} and provide visualizations in the supplementary material. 
 Among the learning-based methods, \ac{dgnn} and \ac{poco}  achieve the most accurate surface reconstructions.
Consistent with findings from \hyperlink{e5}{E5}, \ac{resr} emerges as the most effective among the optimization-based techniques. However, learning-based models achieve higher precision by leveraging priors from ShapeNet thanks to the similar scanning protocol.
}



\subsection{Learning- and optimization-based surface reconstruction from real point clouds (E7)}
\begin{figure*}
\captionsetup[sub]{labelfont=scriptsize,textfont=scriptsize,justification=centering}
    \centering
    \newcommand{\mywidth}{0.16\textwidth}
	\definetrim{mytrim}{50 50 50 50}
    \begin{tabular}{@{}c@{}c@{}c@{}c@{}c@{}c@{}c@{}}

 \rotatebox{90}{\hspace{3mm}Learning}
        & \subfloat[Input]{ \includegraphics[width=\mywidth,mytrim]{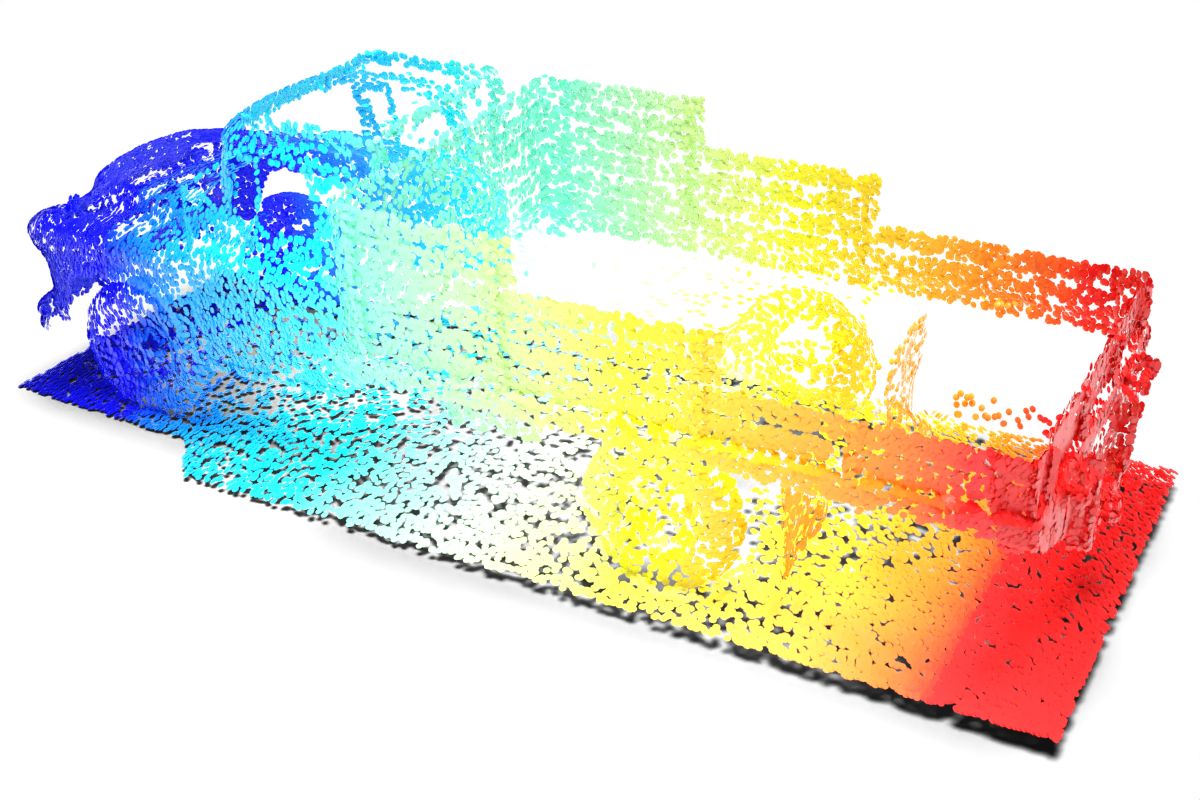}\label{real:truck:input}}
         &  \subfloat[CONet2D]{\includegraphics[width=\mywidth,mytrim]{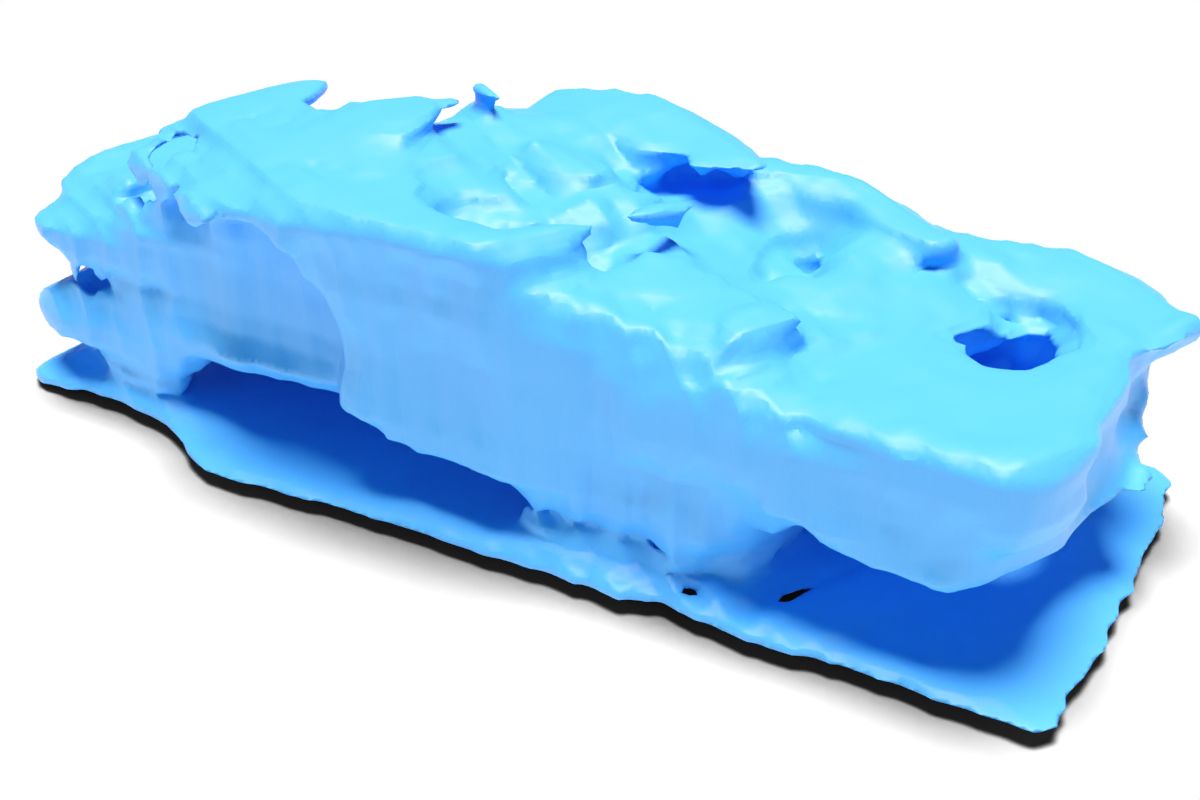}\label{real:truck:start}}
         &  \subfloat[CONet3D]{\includegraphics[width=\mywidth,mytrim]{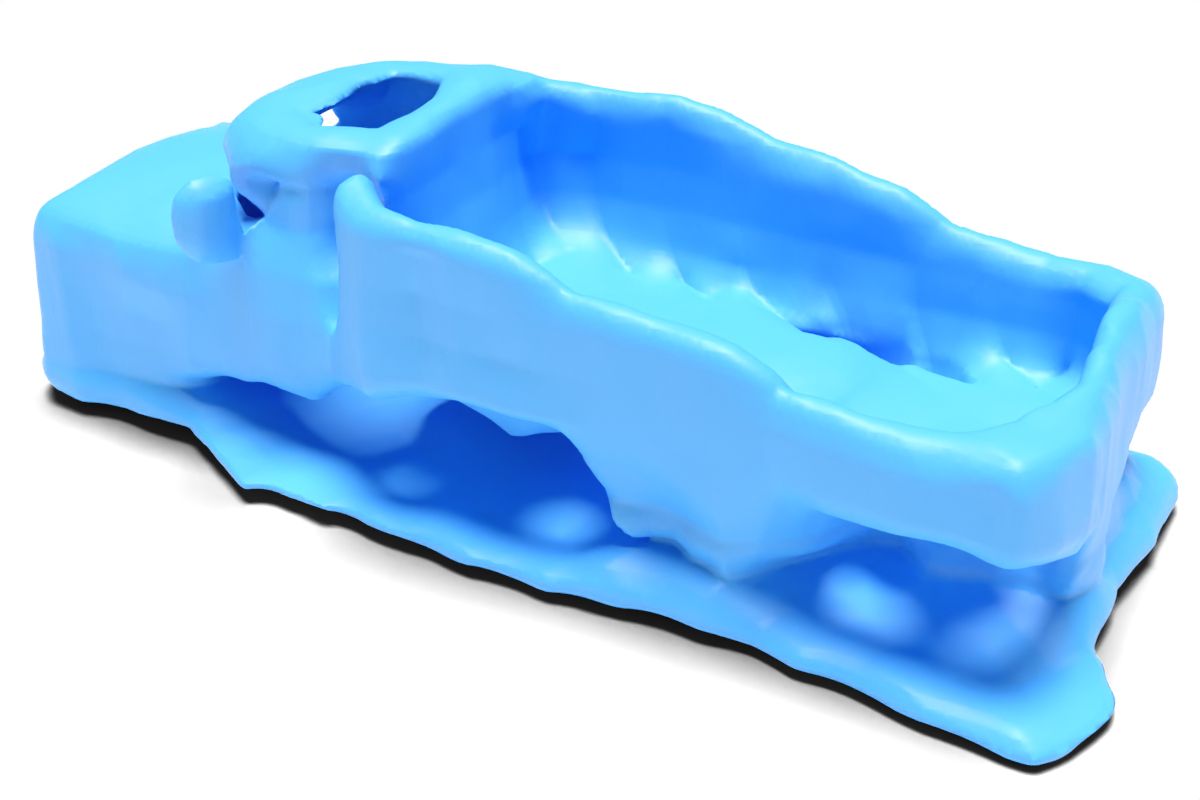}}
         &  \subfloat[SAP]{\includegraphics[width=\mywidth,mytrim]{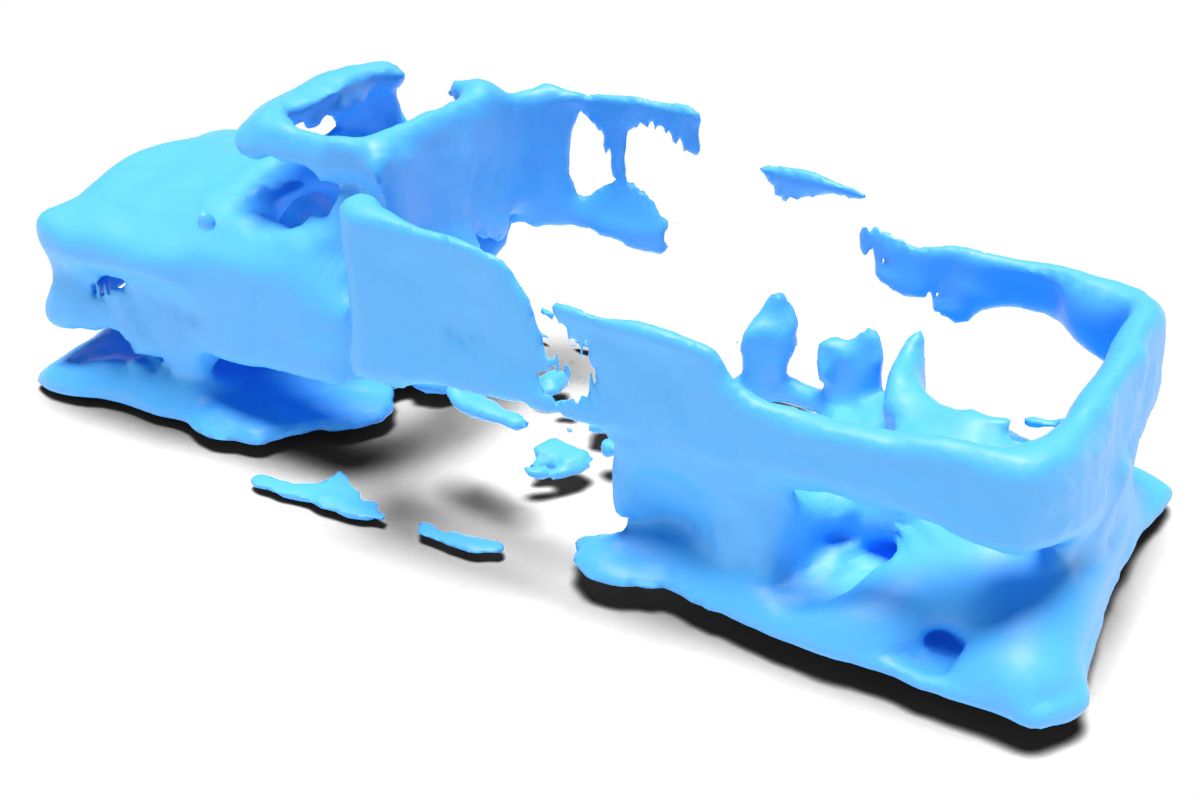}}
         &  \subfloat[POCO]{\includegraphics[width=\mywidth,mytrim]{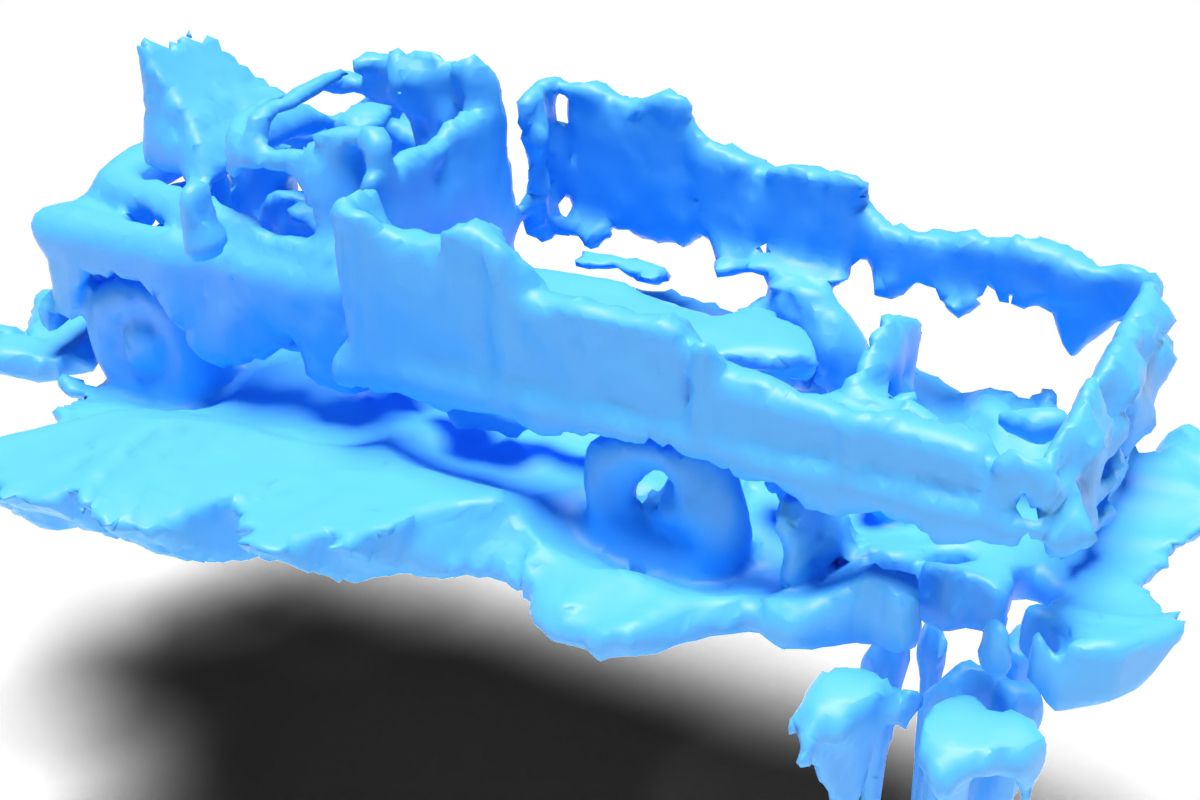}}
         &  \subfloat[DGNN]{\includegraphics[width=\mywidth,mytrim]{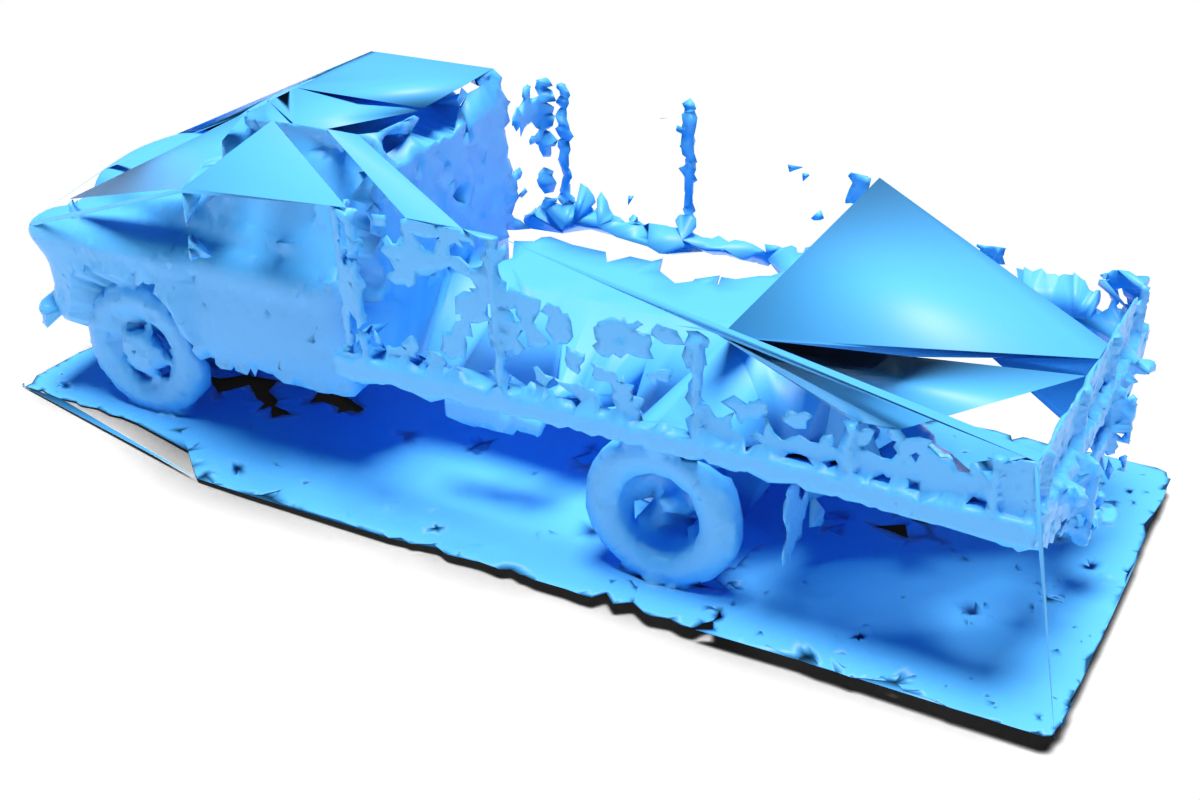}}   
         \\[6mm]
         \rotatebox{90}{\hspace{-1mm}Optimization}
         & \subfloat[IGR]{\includegraphics[width=\mywidth,mytrim]{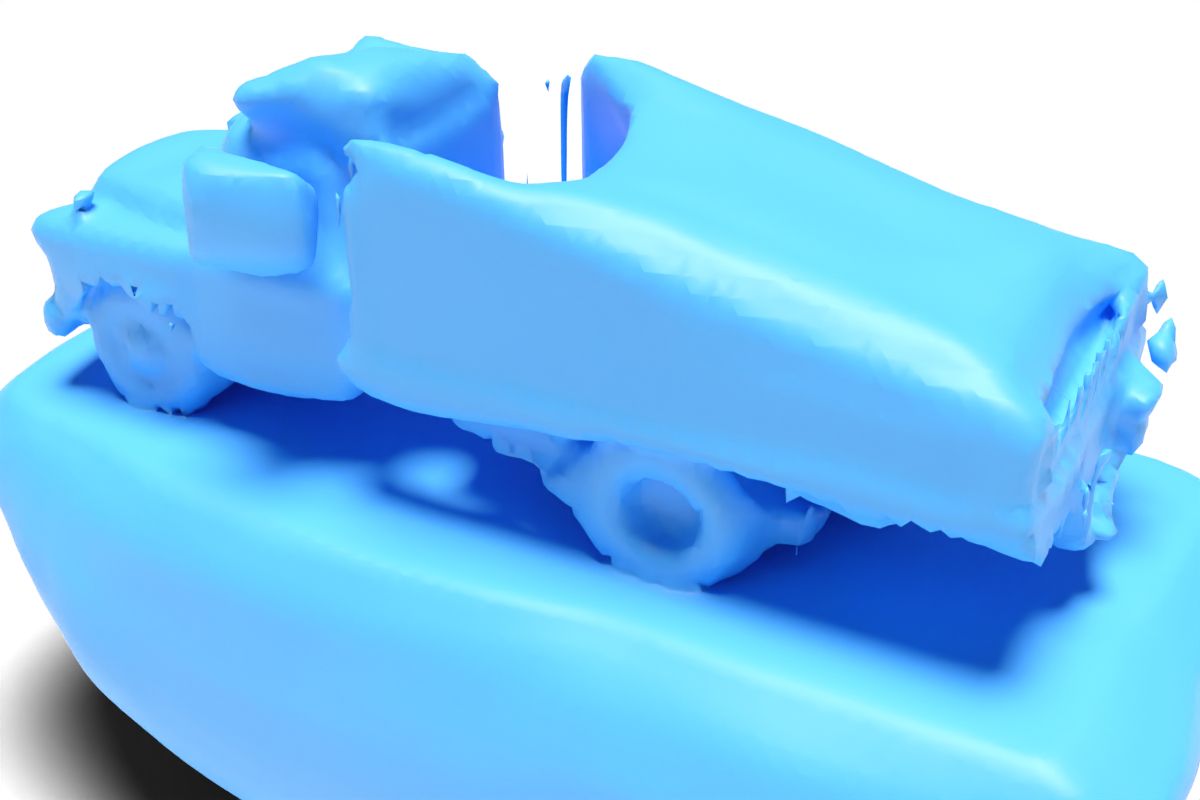}}
         &  \subfloat[LIG]{\includegraphics[width=\mywidth,mytrim]{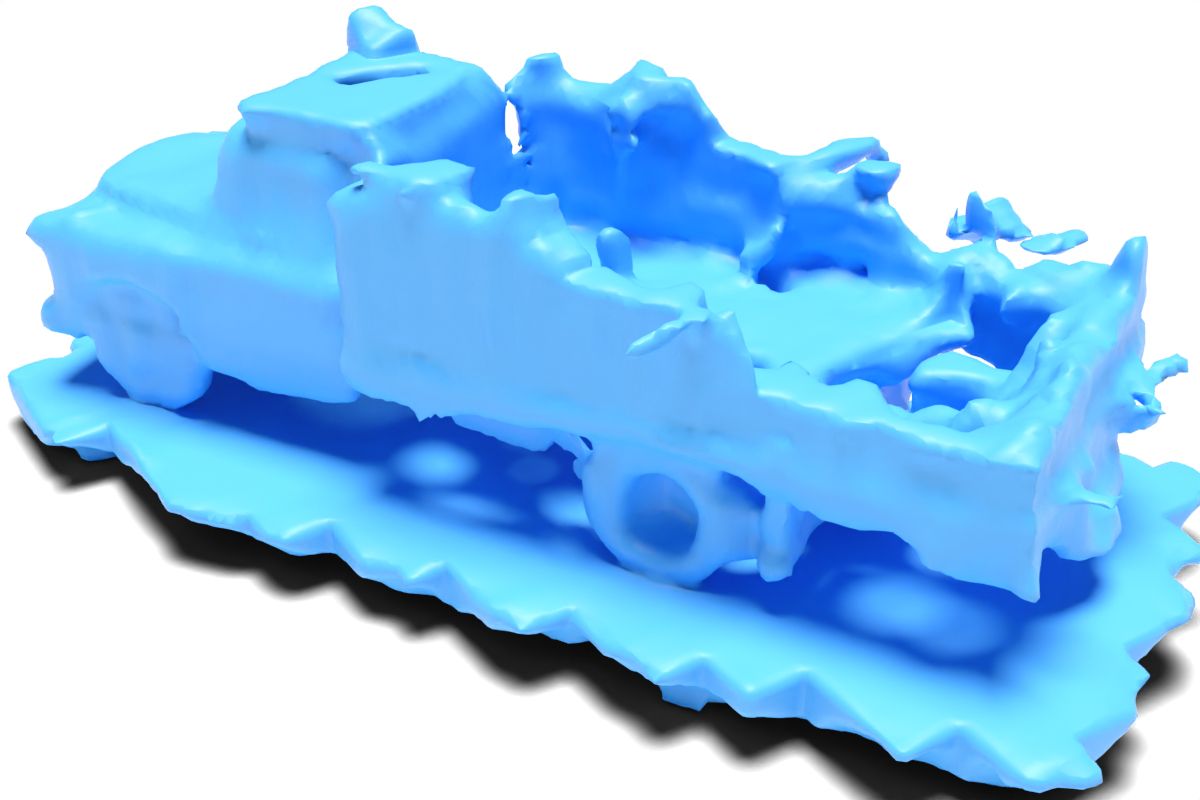}}
         &  \subfloat[P2M]{\includegraphics[width=\mywidth,mytrim]{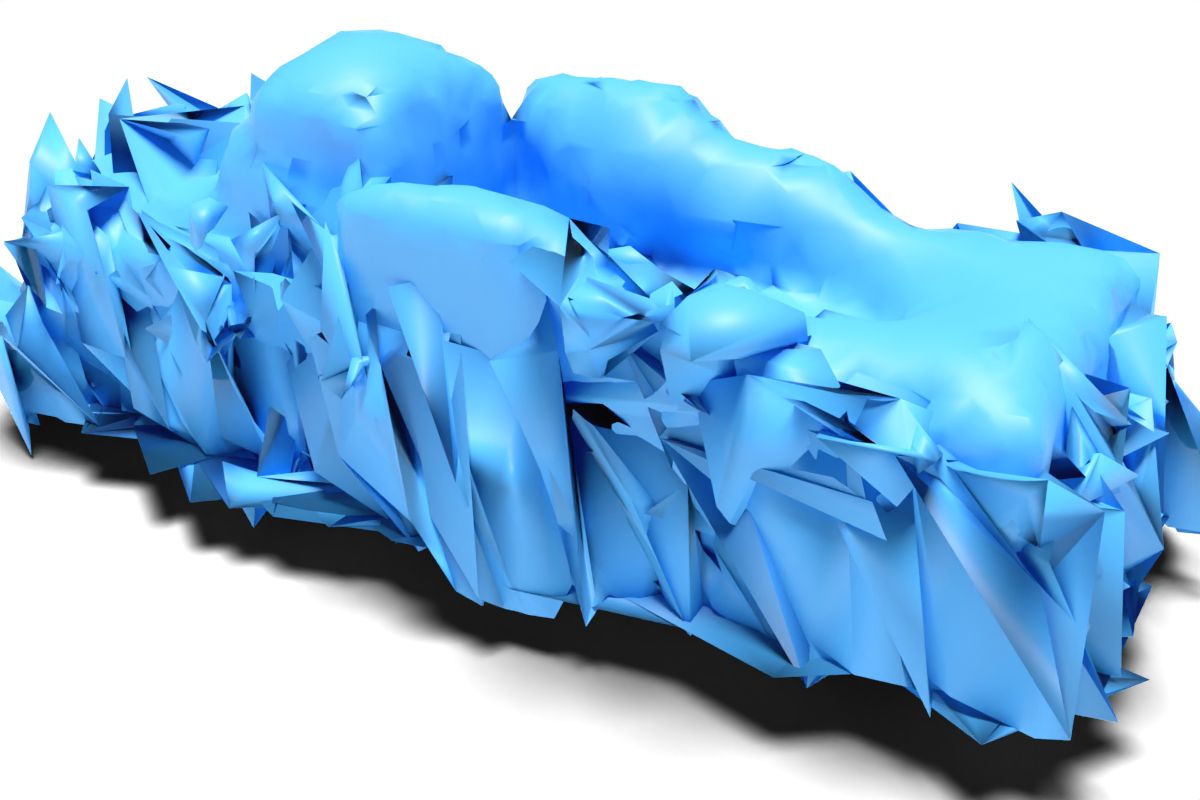}}
         &  \subfloat[SAP\optim]{\includegraphics[width=\mywidth,mytrim]{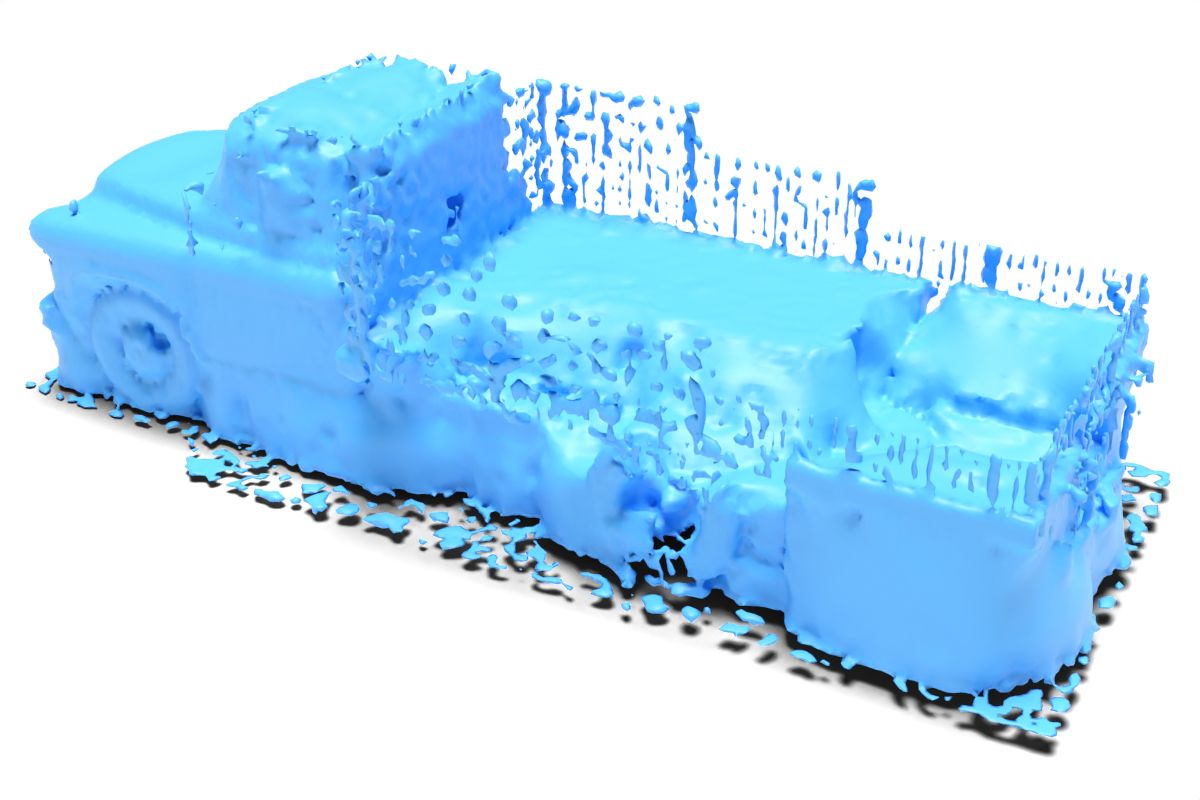}}
         &  \subfloat[SPSR]{\includegraphics[width=\mywidth,mytrim]{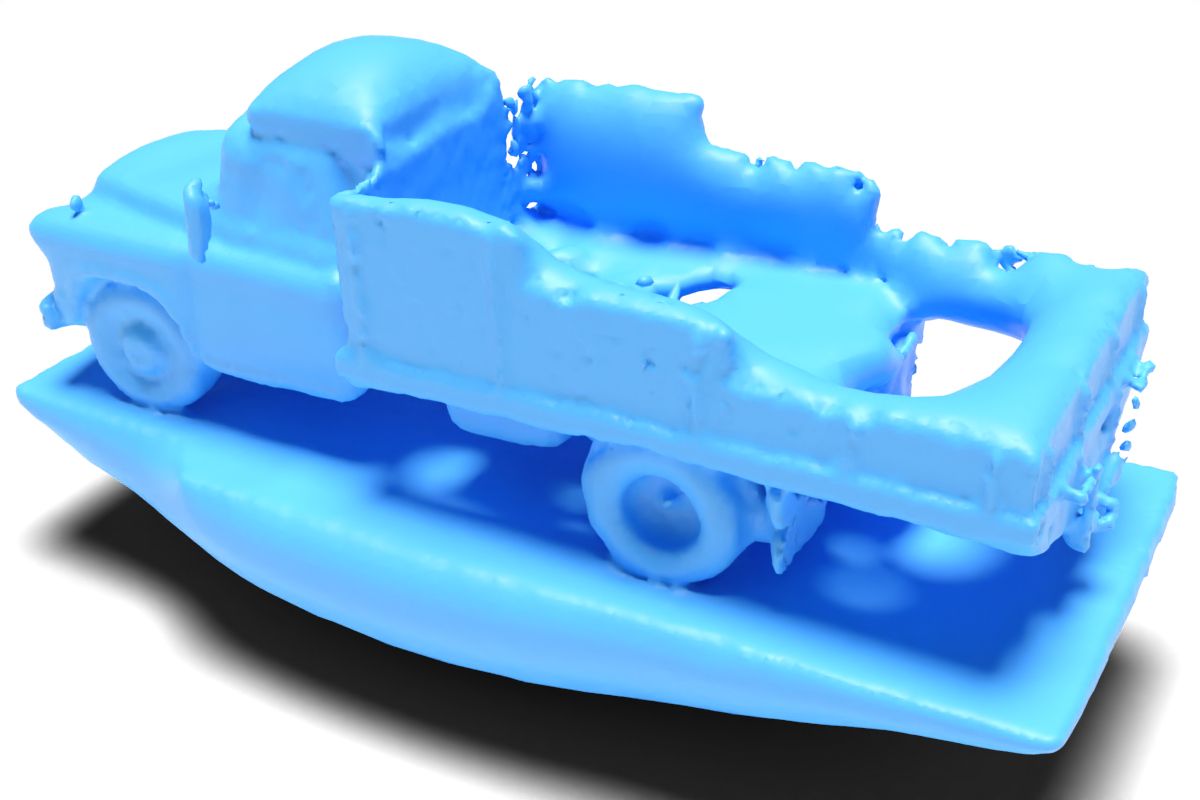}}         
         &  \subfloat[RESR]{\includegraphics[width=\mywidth,mytrim]{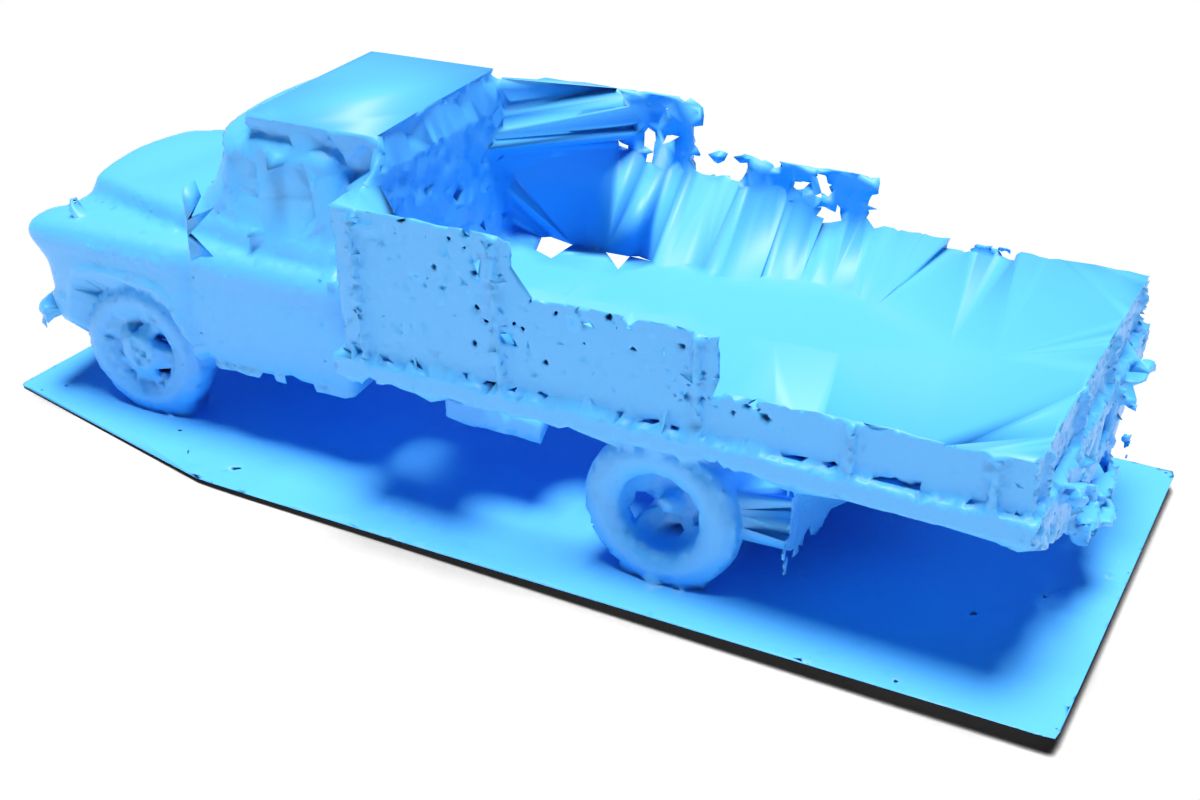}\label{real:truck:end}}   
\\[5mm] \midrule \\[-3mm]
        \rotatebox{90}{\hspace{13mm}Learning}
        & \subfloat[Input]{ \includegraphics[width=\mywidth,mytrim]{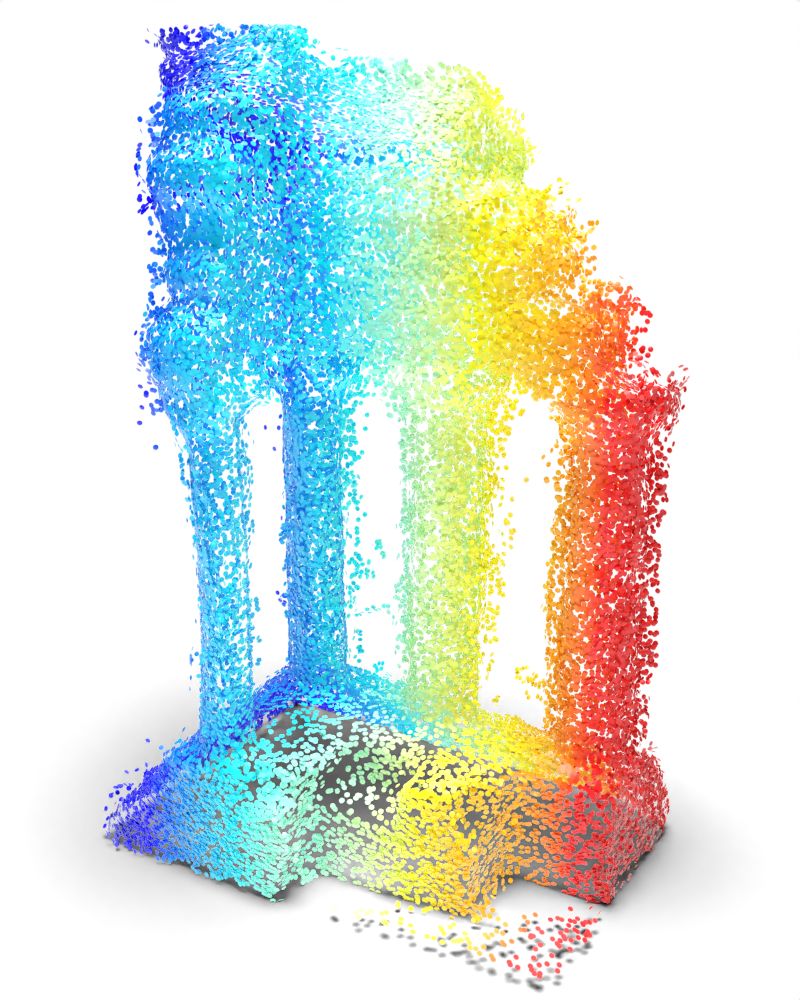}\label{real:temple:input}}
         &  \subfloat[CONet2D]{\includegraphics[width=\mywidth,mytrim]{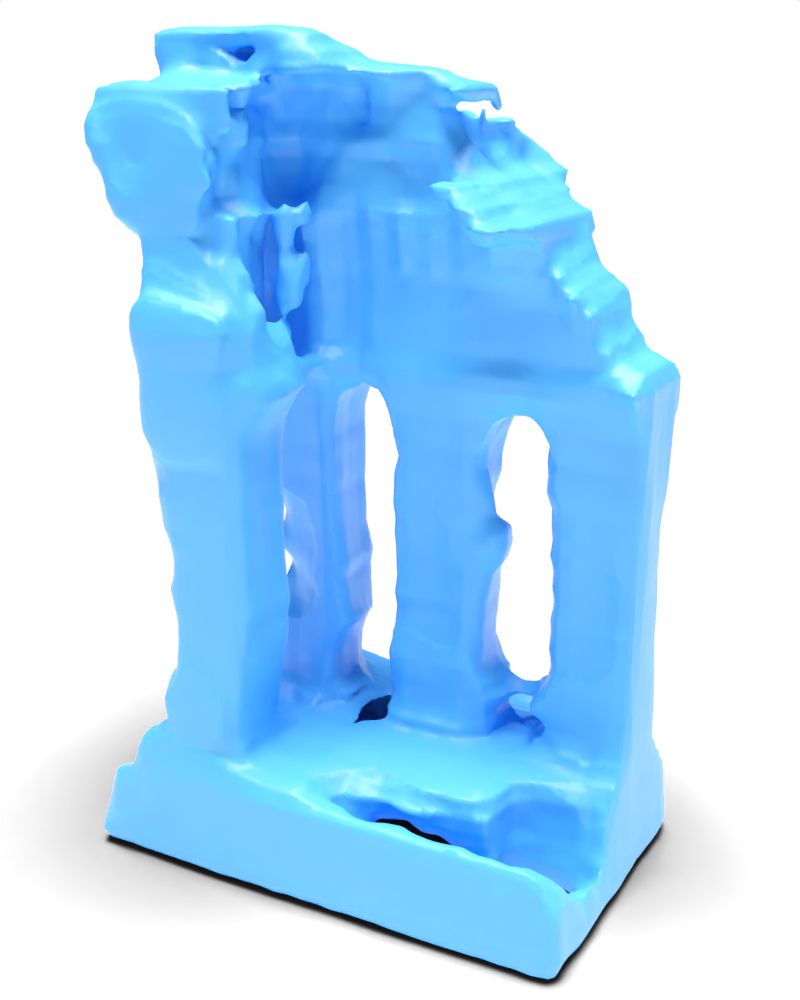}\label{real:temple:start}}
         &  \subfloat[CONet3D]{\includegraphics[width=\mywidth,mytrim]{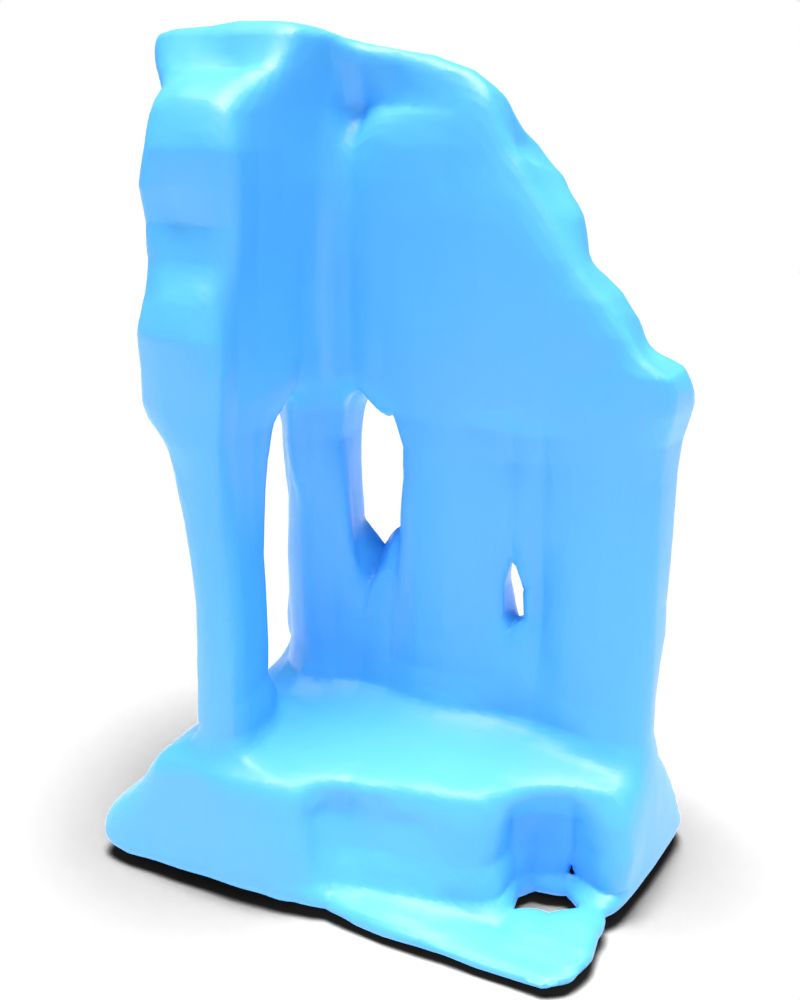}}
         &  \subfloat[SAP]{\includegraphics[width=\mywidth,mytrim]{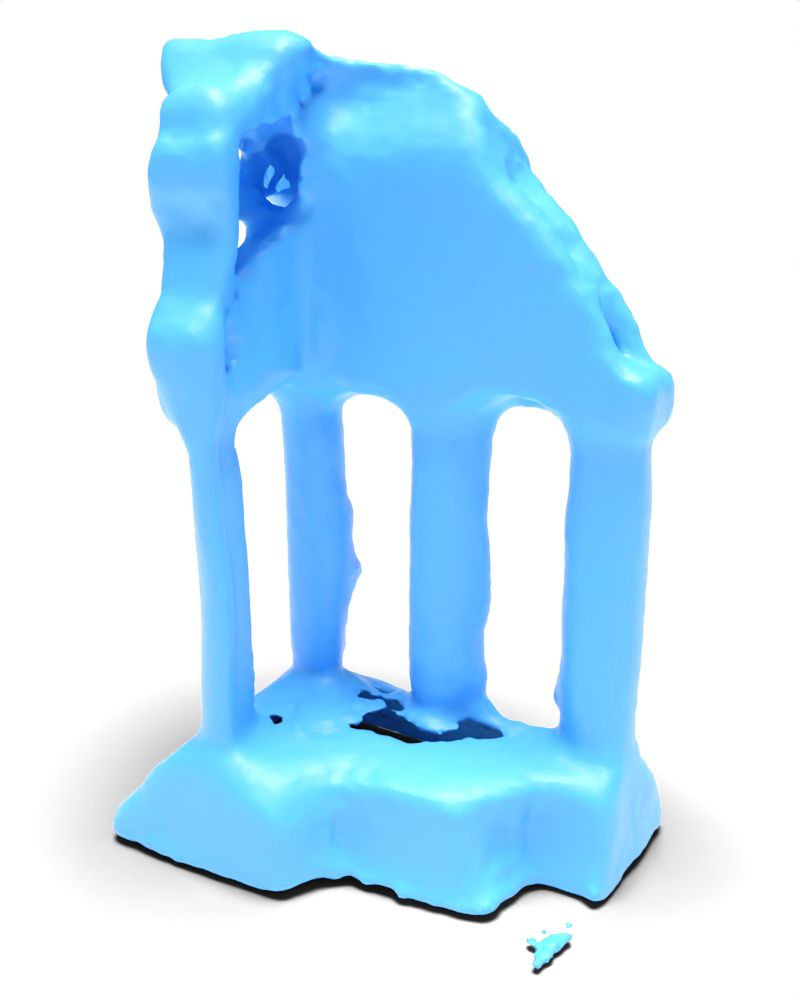}\label{real:temple:sap}}
         &  \subfloat[POCO]{\includegraphics[width=\mywidth,mytrim]{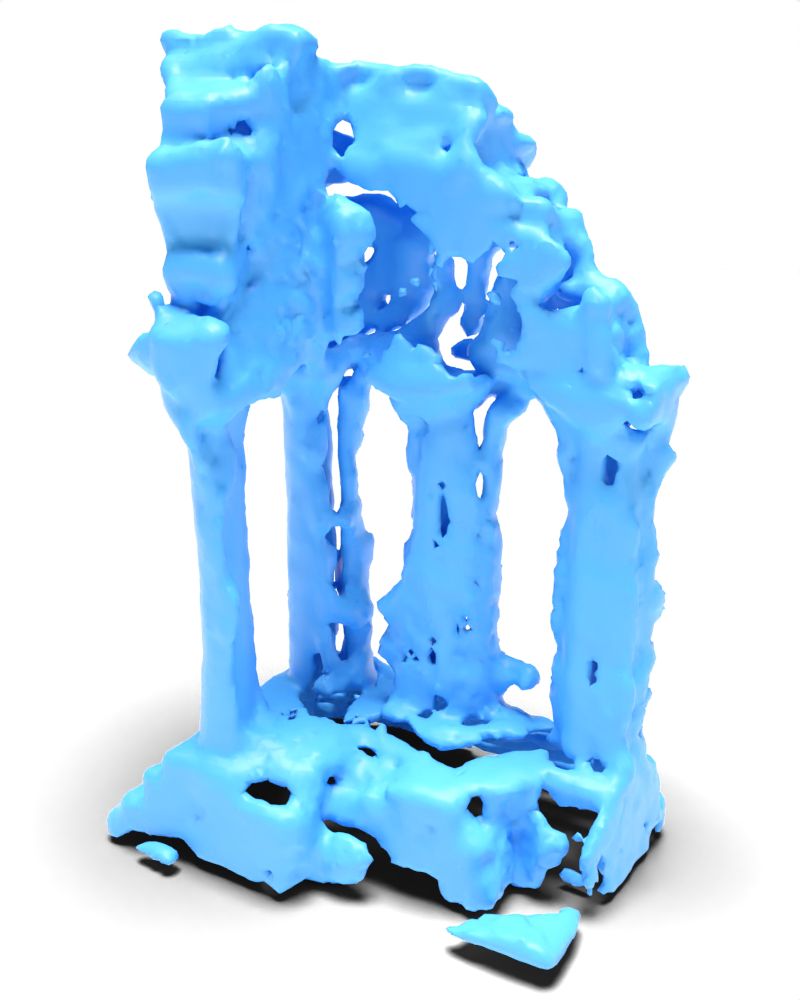}}
         &  \subfloat[DGNN]{\includegraphics[width=\mywidth,mytrim]{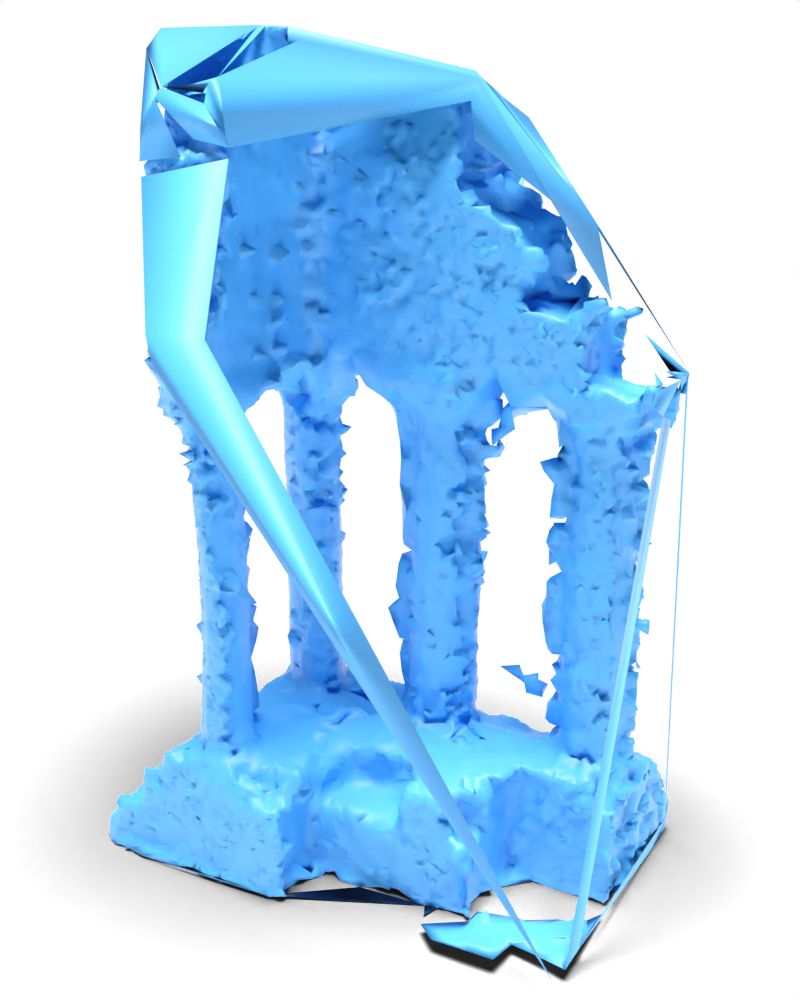}}   
         \\[5mm]
         \rotatebox{90}{\hspace{9mm}Optimization}
         & \subfloat[IGR]{\includegraphics[width=\mywidth,mytrim]{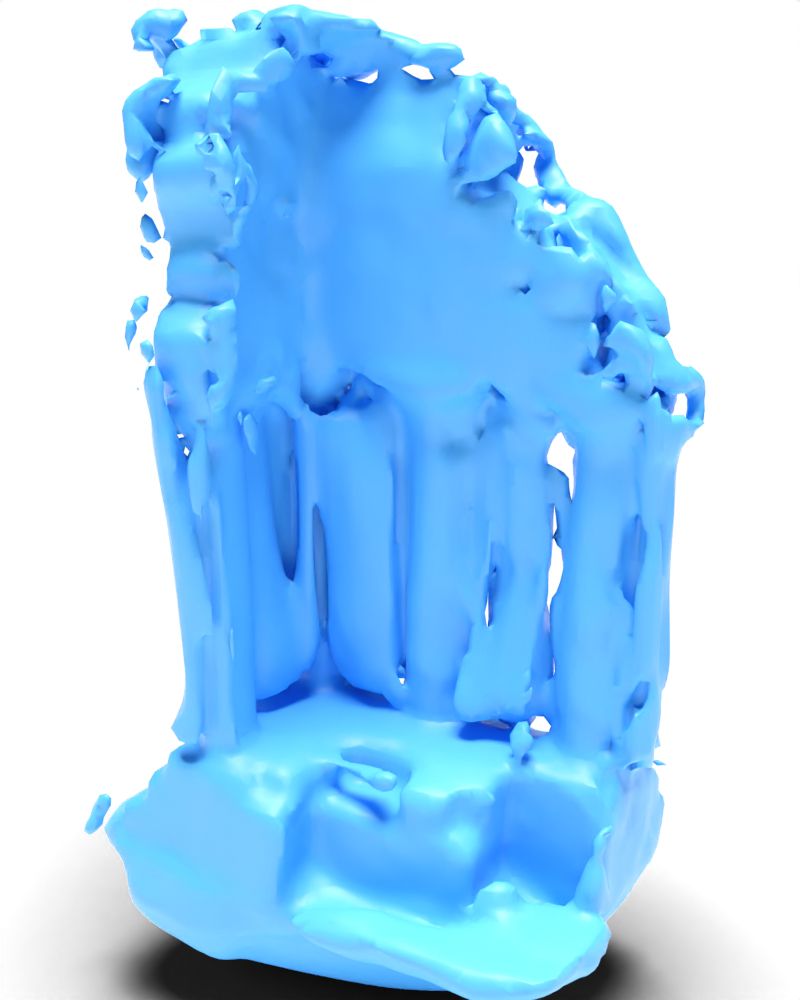}}
         &  \subfloat[LIG]{\includegraphics[width=\mywidth,mytrim]{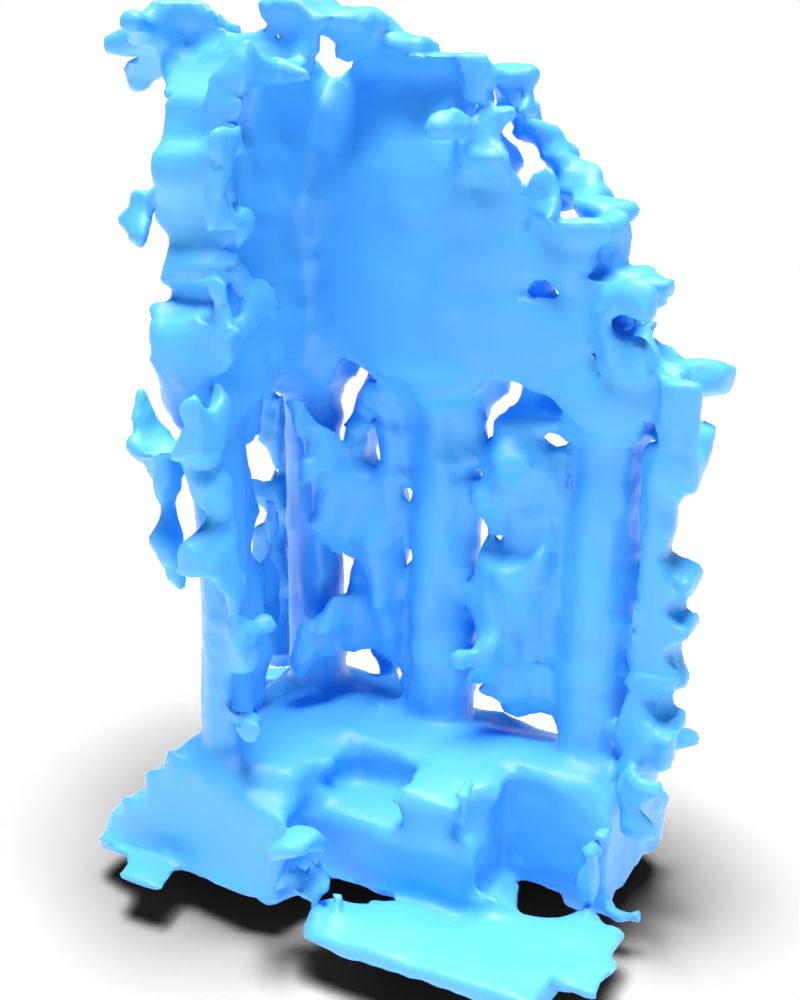}}
         &  \subfloat[P2M]{\includegraphics[width=\mywidth,mytrim]{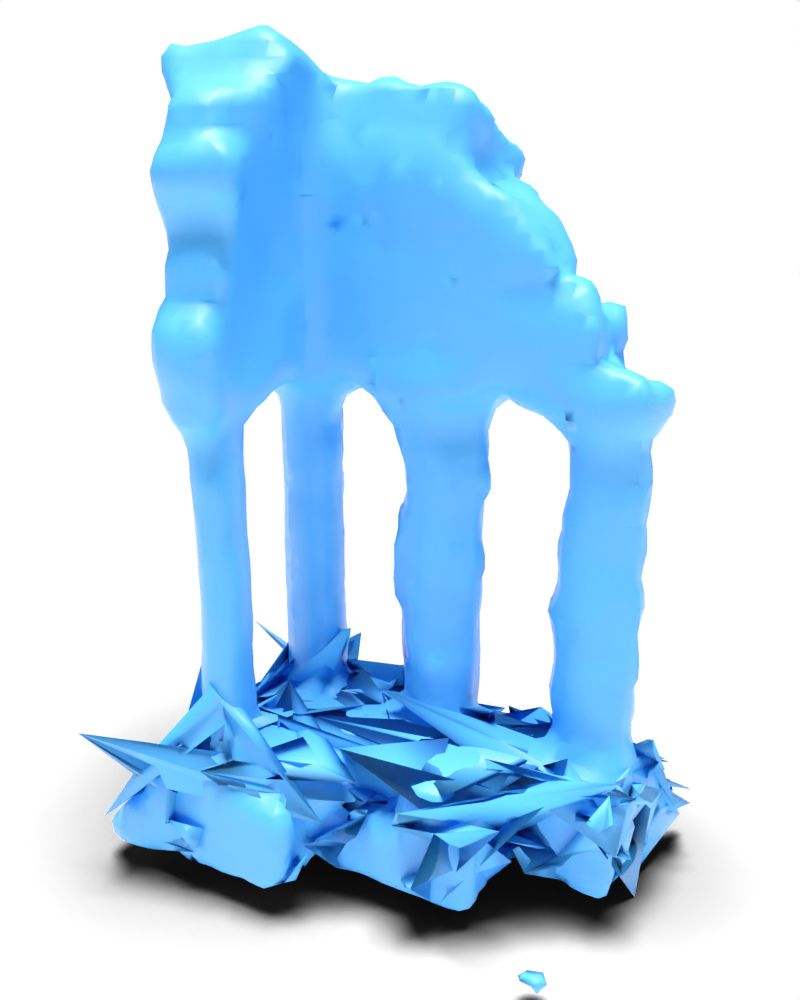}\label{real:temple:p2m}}
         &  \subfloat[SAP\optim]{\includegraphics[width=\mywidth,mytrim]{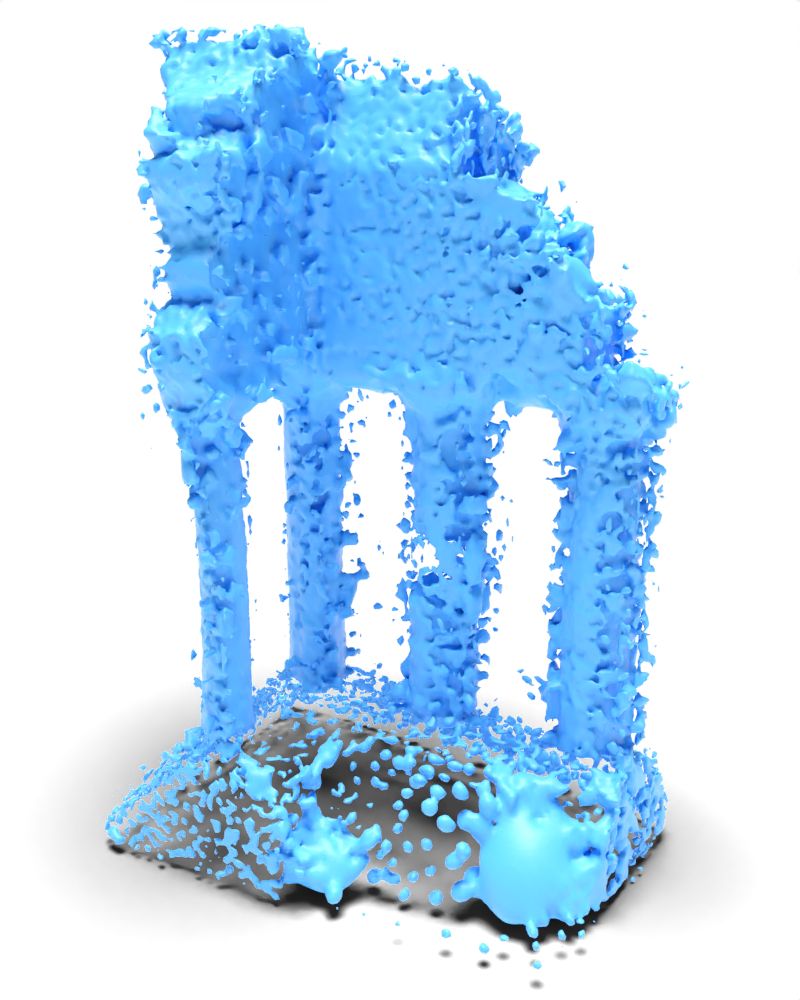}}
         &  \subfloat[SPSR]{\includegraphics[width=\mywidth,mytrim]{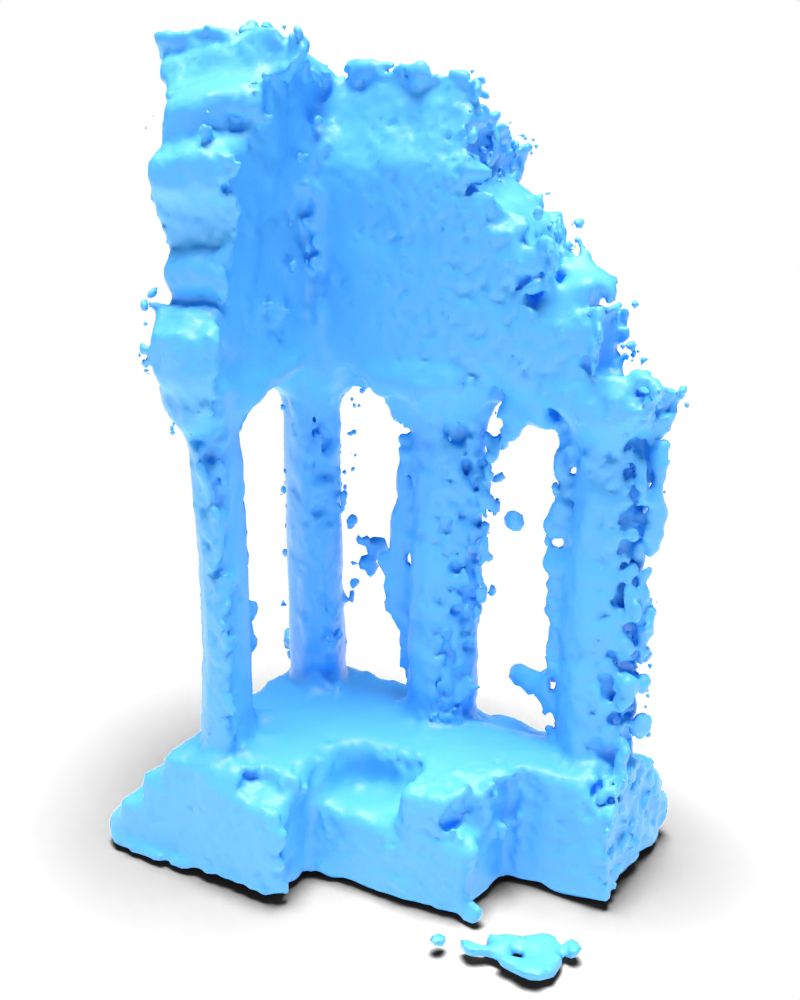}\label{real:temple:spsr}}         
         &  \subfloat[RESR]{\includegraphics[width=\mywidth,mytrim]{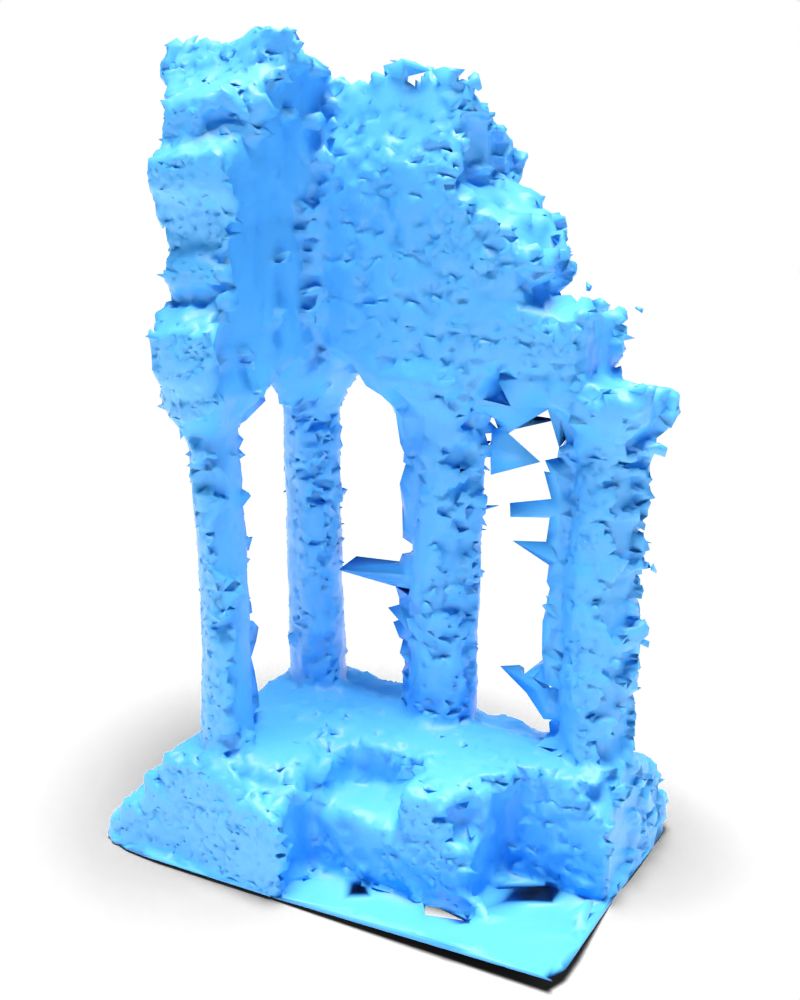}\label{real:temple:end}\label{real:temple:labatut}}   
\\[5mm] \midrule \\[-3mm]
        \rotatebox{90}{\hspace{3mm}Learning}
        & \subfloat[Input]{ \includegraphics[width=\mywidth,mytrim]{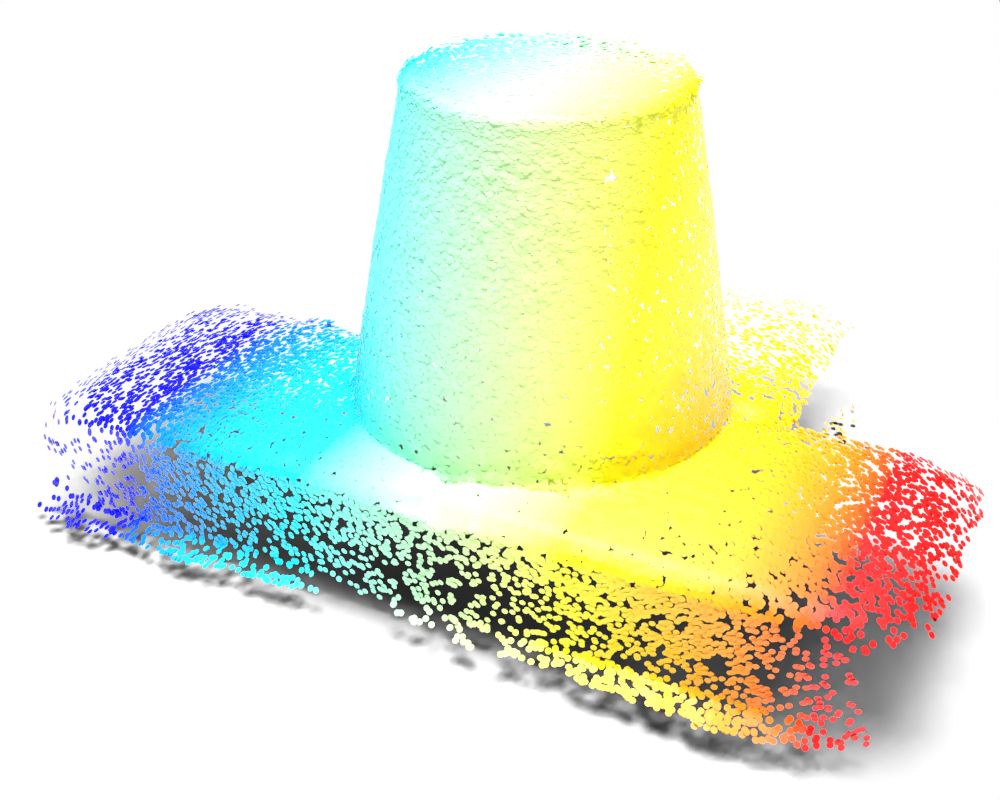}\label{real:scan1:input}}
         &  \subfloat[CONet2D]{\includegraphics[width=\mywidth,mytrim]{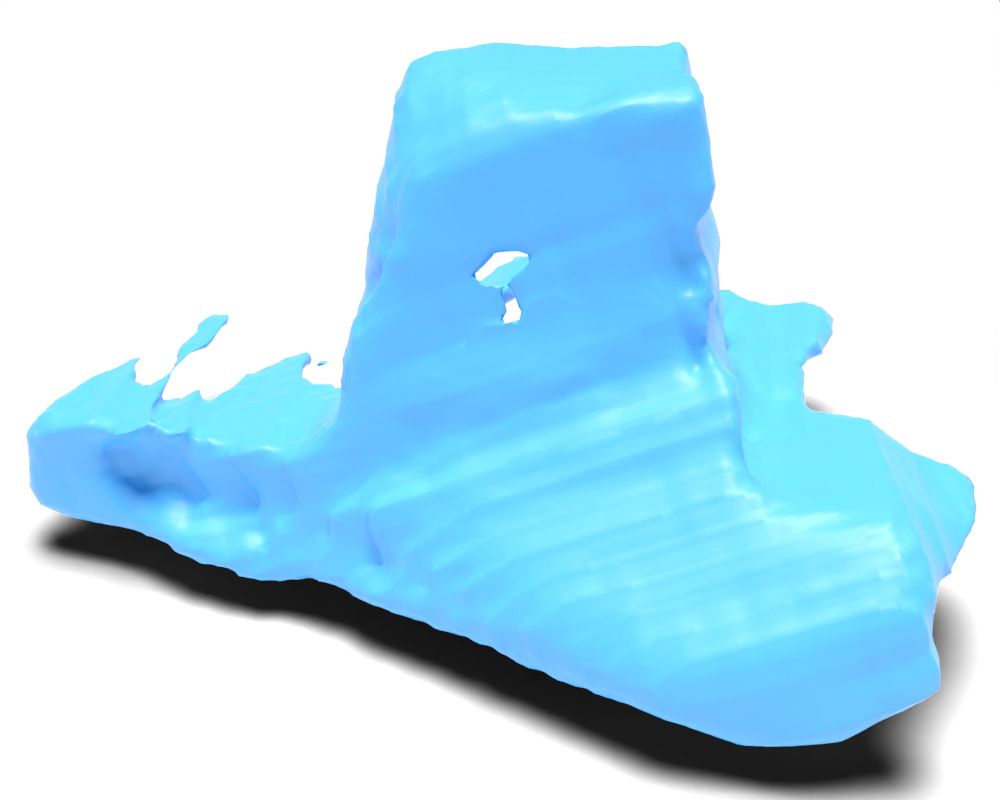}\label{real:scan1:start}}
         &  \subfloat[CONet3D]{\includegraphics[width=\mywidth,mytrim]{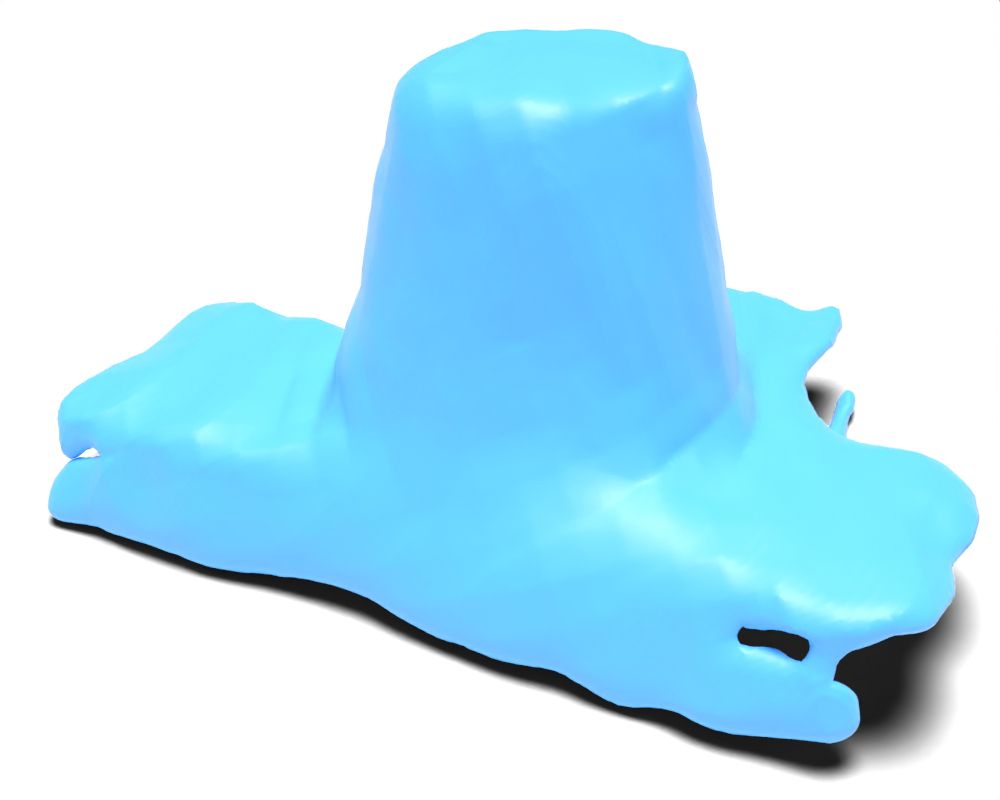}}
         &  \subfloat[SAP]{\includegraphics[width=\mywidth,mytrim]{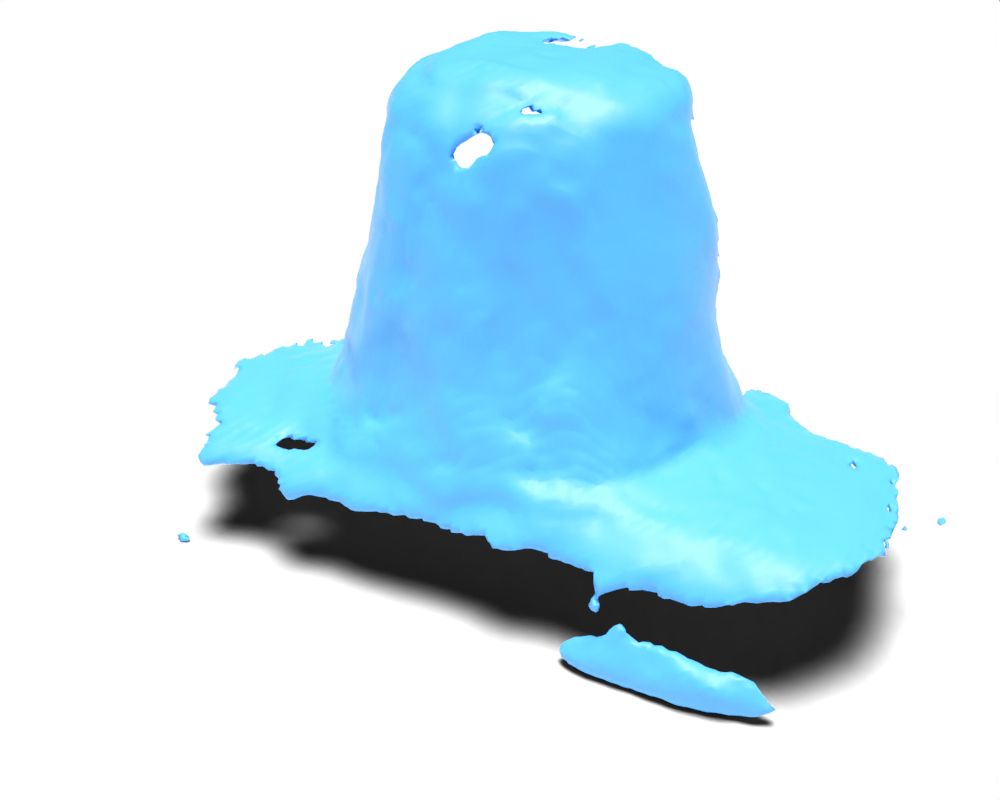}}
         &  \subfloat[POCO]{\includegraphics[width=\mywidth,mytrim]{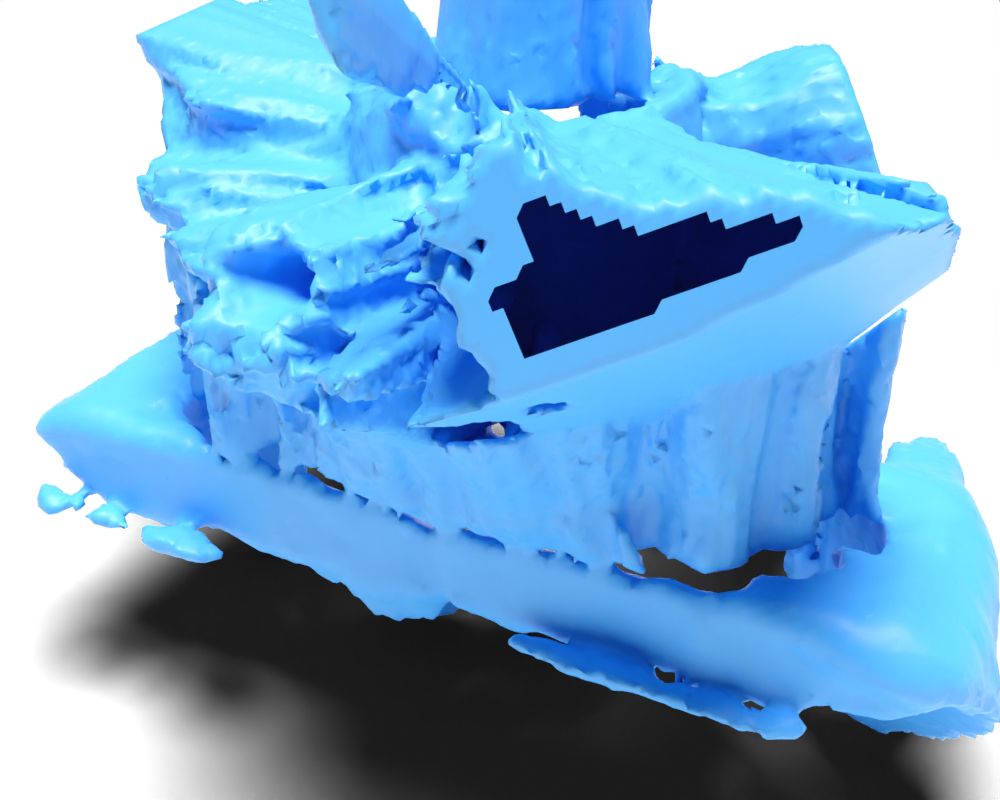}}
         &  \subfloat[DGNN]{\includegraphics[width=\mywidth,mytrim]{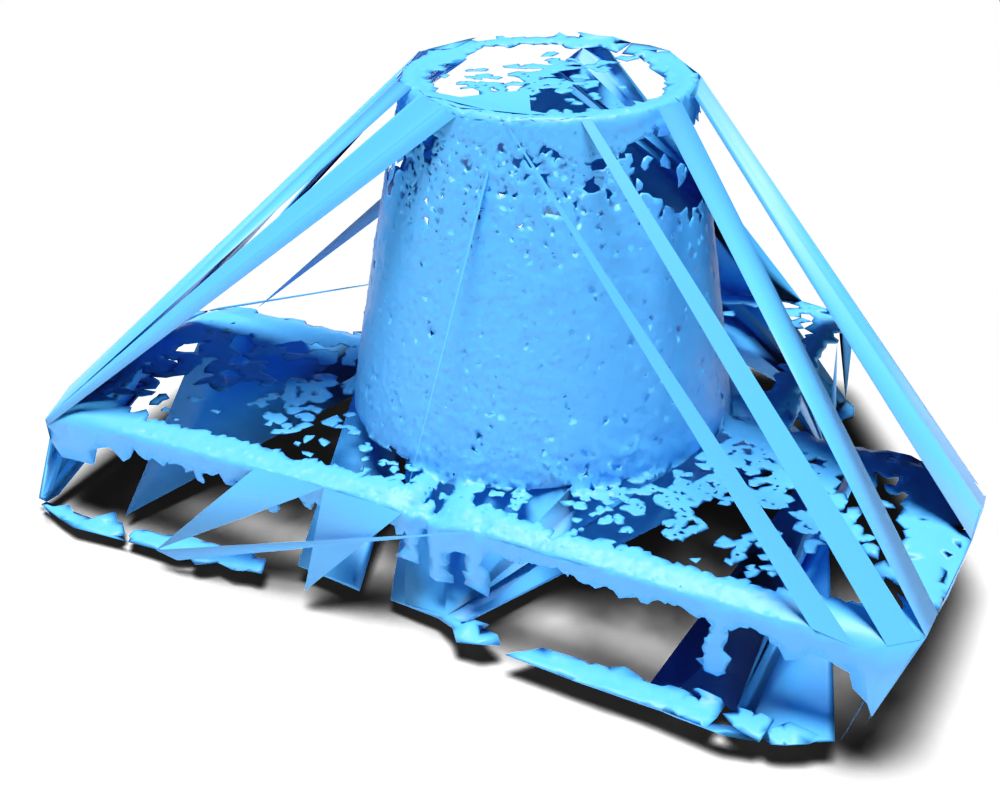}}   
         \\[5mm]
         \rotatebox{90}{\hspace{-1mm}Optimization}
         & \subfloat[IGR]{\includegraphics[width=\mywidth,mytrim]{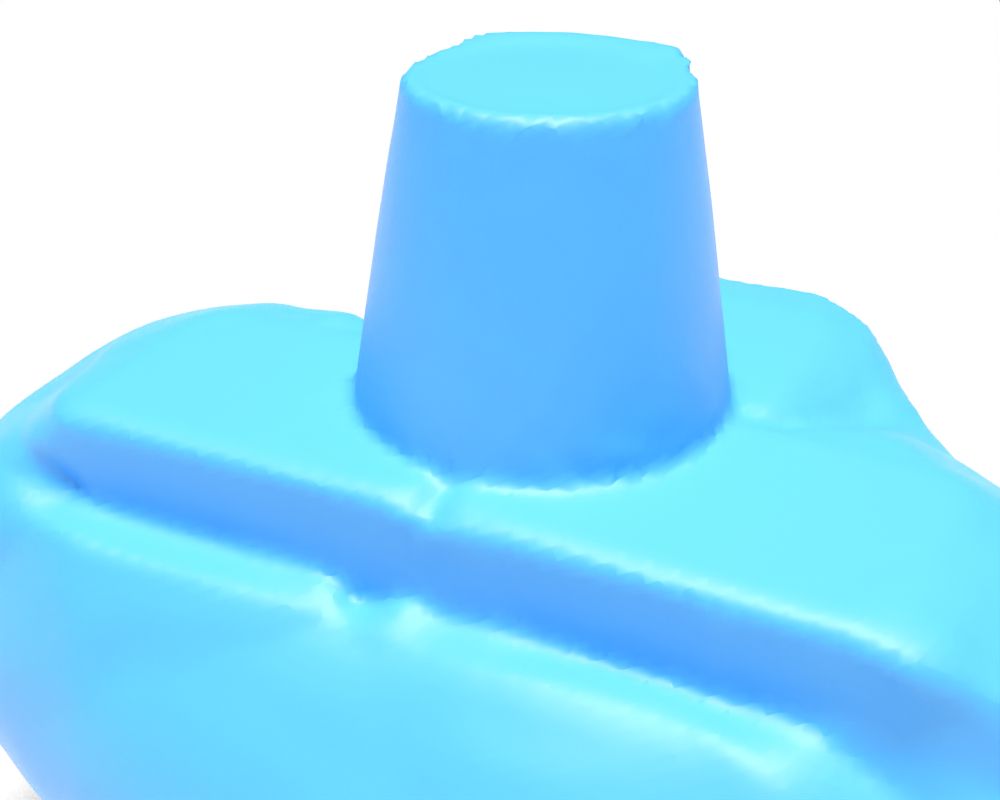}}
         &  \subfloat[LIG]{\includegraphics[width=\mywidth,mytrim]{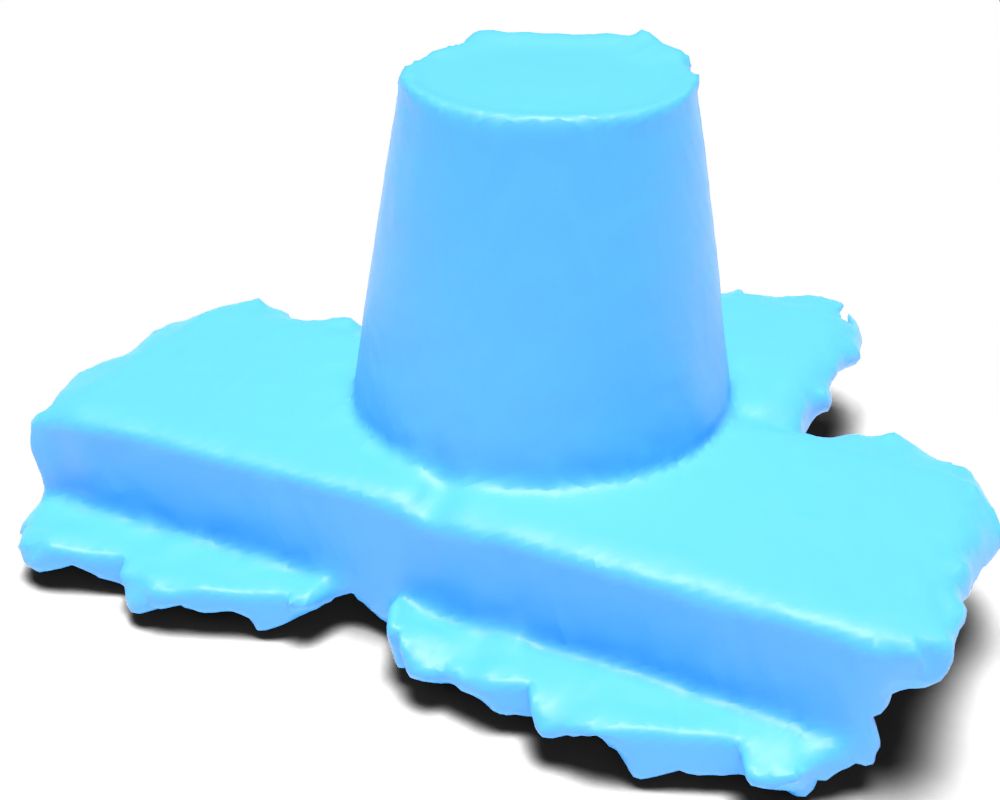}}
         &  \subfloat[P2M]{\includegraphics[width=\mywidth,mytrim]{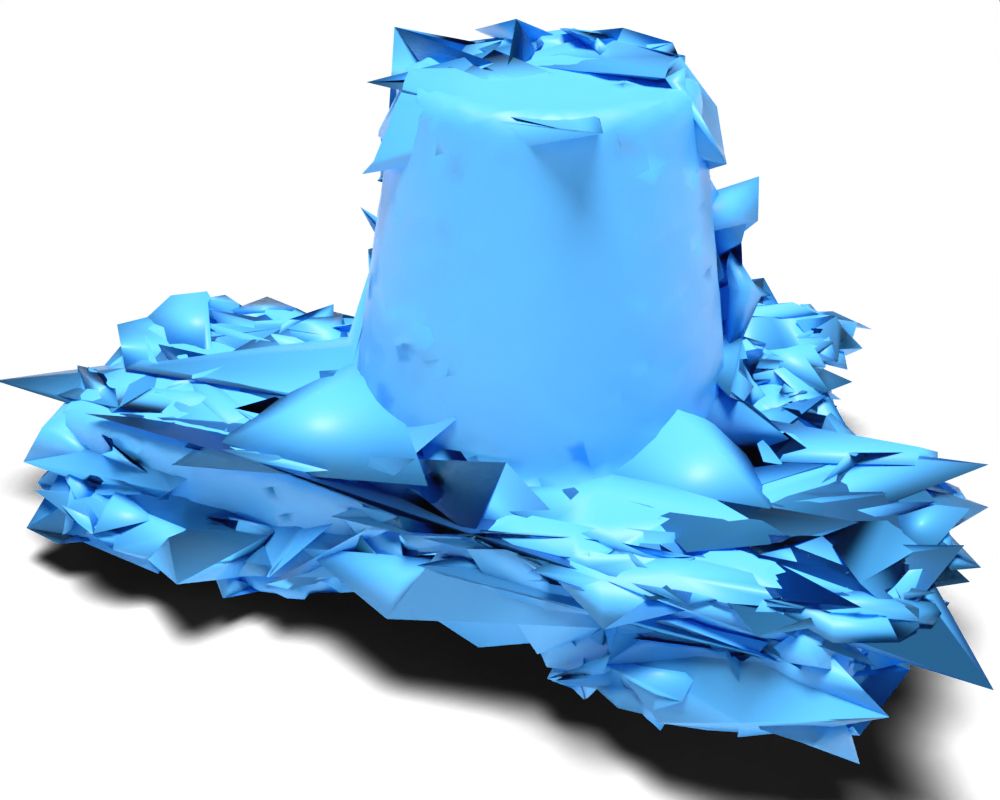}}
         &  \subfloat[SAP\optim]{\includegraphics[width=\mywidth,mytrim]{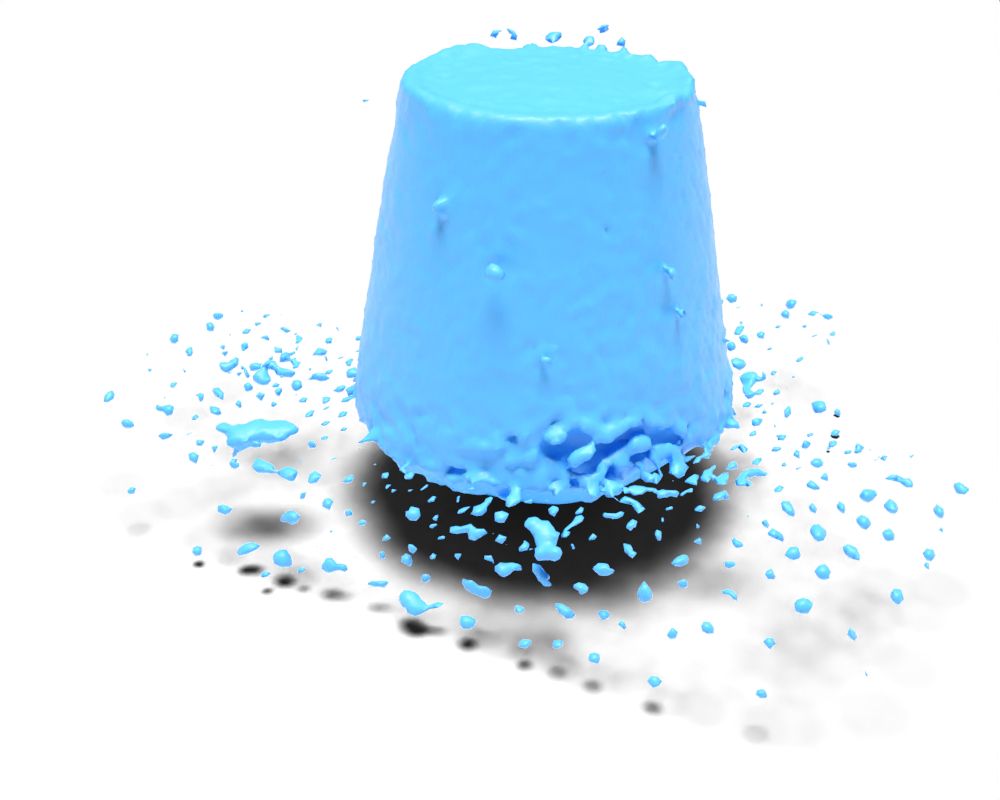}}
         &  \subfloat[SPSR]{\includegraphics[width=\mywidth,mytrim]{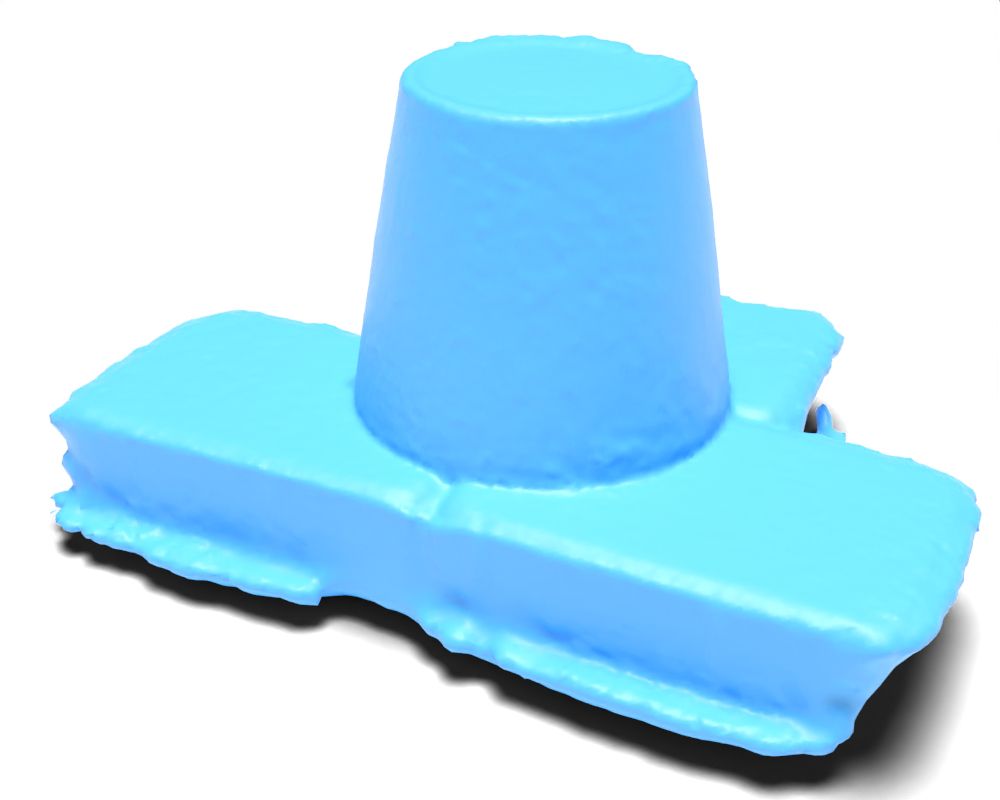}}         
         &  \subfloat[RESR]{\includegraphics[width=\mywidth,mytrim]{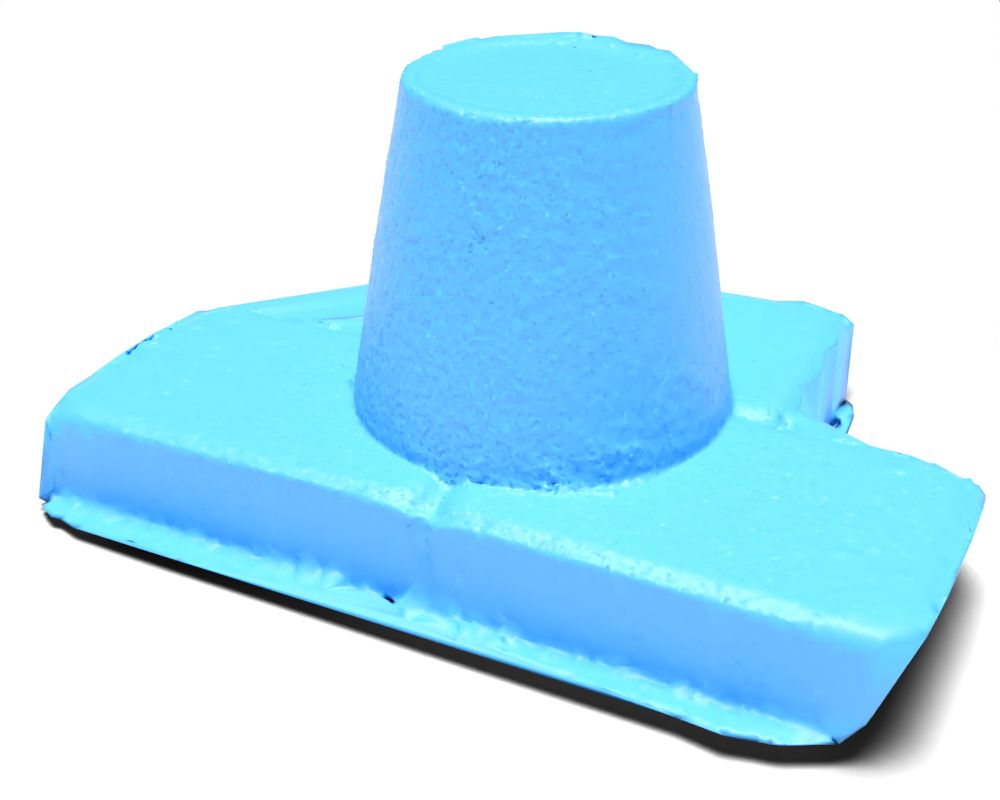}\label{real:scan1:end}}   
    \end{tabular}
\caption{\textbf{Comparison of all methods - real data without ground truth (\protect\hyperlink{e7}{E7}):} We show reconstructions of two real‐world MVS and one range-scanned point clouds.
The learning methods were trained on synthetic MVS scans from ShapeNet (\cf \protect\hyperlink{e1}{E1}). Optimization-based methods are optimized per scan using standard settings. The two traditional methods \ac{spsr} \cite{screened_poisson} and \ac{resr} \cite{Labatut2009a} reconstruct the visually best surfaces. Their reconstructions only show noticeable defects from the heavy noise of the \emph{Temple Ring} MVS point cloud (Figs.~(\protect\subref{real:temple:spsr}) and (\protect\subref{real:temple:end})).
}
    \label{fig:ch2:real}
\end{figure*}

In this final experiment, we qualitatively assess the performance of learning- and optimization-based methods on real \ac{mvs} and range-scanned point clouds (without ground truth). For the learning-based approaches, we use models trained on synthetic \ac{mvs} scans from ShapeNet (\hyperlink{e4}{E4}), while optimization-based methods are applied independently to each point cloud. Our findings are presented in \figref{fig:ch2:real}.

\vspace{0.1cm}\noindent\textbf{Tanks and Temples range scanning, \figsref{real:truck:input}-\ref{real:truck:end}:} Except for DGNN, the learning-based methods struggle to adapt to real range scanning acquisitions. IGR manages to create a reasonable mesh, but traditional methods like \ac{spsr} and \ac{resr} yield the most complete and accurate surfaces.
     
\vspace{0.1cm}\noindent\textbf{Middlebury \ac{mvs}, \figsref{real:temple:input}-\ref{real:temple:end}:} \Ac{sap} stands out as the only learning-based method capable of reconstructing this challenging \ac{mvs} scan, resulting in a smooth and complete surface, albeit with minor topological noise, such as holes. The optimization-based method \ac{p2m} delivers a visually appealing reconstruction with minimal defects (\figref{real:temple:p2m}).
     
\vspace{0.1cm}\noindent\textbf{DTU \ac{mvs}, \figsref{real:scan1:input}-\ref{real:scan1:end}:} Both learning- and optimization-based methods find this scene challenging, while traditional methods succeed in producing convincing surfaces.

This experiment reinforces our findings from synthetic point clouds: learning-based methods are vulnerable to unfamiliar defects, whereas traditional methods offer a robust and reliable baseline.

\subsection{Runtimes}

\begin{table}
\caption[Runtimes for learning-based reconstruction]{
	\textbf{Runtimes of surface reconstruction methods:} Times (in seconds) for reconstructing one object from a point cloud of 3,000 points averaged over the ShapeNet test set. GC stand for Graph-cut; SE stands for surface extraction, such as marching cubes or triangle-from-tetrahedron. Note that different variants and implementations of marching cubes are used by different methods, which also influences the runtimes. \Ac{sap}~\cite{Peng2021SAP} has the fastest total runtime. $^\dagger$methods not based on neural networks.
}
\centering
\resizebox{\columnwidth}{!}{%
\begin{tabular}{l@{~}cccccc}
\toprule
\textbf{Model}        &                     & {Feature extraction}        & {Decoding/GC}       & {SE}        & {Total}         \\ \midrule
\textbf{ConvONet2D}&\cite{Peng2020}         & 0.016                     & 0.32              & 0.17      & 0.51          \\
\rowcolor{gray!10}\textbf{ConvONet3D} ~~ &\cite{Peng2020}         & 0.008                     & 0.21              & 0.17      & 0.40          \\
\textbf{SAP}&\cite{Peng2021SAP}             & 0.022                     & 0.02             & 0.05     & \textbf{0.09}\\
\rowcolor{gray!10}\textbf{DGNN}&  \cite{dgnn}                 & 0.110                      & 0.28              & 0.01      & 0.39          \\ 
\textbf{POCO}&\cite{boulch2022poco}         & 0.088                     & 13.72             & 0.33      & 15.74         \\
\rowcolor{gray!10}\textbf{P2S}&\cite{points2surf}             &\multicolumn{2}{c}{\qquad69.06}                      & 11.51     & 80.57         \\
$^\dagger$\textbf{SPSR}&  \cite{screened_poisson}     &-&-&-& 1.25\\ 
\rowcolor{gray!10}$^\dagger$\textbf{RESR}&\cite{Labatut2009a}  & 0.1                       & 0.07              & 0.01      & 0.18          \\ 
\bottomrule
\end{tabular}}
\label{tab:ch2:timing}
\end{table}




{The detailed runtimes for the methods evaluated in the learning-based experiments are presented in \tabref{tab:ch2:timing}. \ac{sap} stands out as the fastest reconstruction method. \ac{dgnn} also exhibits efficient runtimes, whereas \ac{poco} tends to be slower, attributable primarily to its intensive {and costly need to search for query neighbors.} 
Additionally, we include runtime comparisons for \ac{p2s}. Due to its extensive computational demands during both training and inference, \ac{p2s} was excluded from experiments \hyperlink{e1}{E1} through \hyperlink{e4}{E4}.}

\subsection{Summary and analysis}

When the point clouds in the training and test sets have similar characteristics, learning-based methods produce high-quality reconstructions and even show robustness to noise and missing data. They are able to generalize to unseen shape categories, provided that the training set is sufficiently large (30k shapes in our experiments) and contains shapes of sufficient complexity.
 This ability implies that the methods primarily learn priors associated with point cloud attributes rather than the shapes themselves. However, learning-based methods do not produce satisfying results when the training shapes are too simple, or when the point clouds include unknown defects, such as outliers.
 Hybrid approaches such as  \ac{sap} or \ac{dgnn} achieve higher robustness to domain shifts and shorter reconstruction times. 
With the exception of \ac{igr}, the newer optimization-based methods we tested do not exhibit strong resilience to acquisition defects and generally do not outperform the traditional methods \ac{spsr} and \ac{resr}.

\section{Conclusion}

{Surface reconstruction from point clouds is a central subject in digital geometry processing, continually evolving with great strides in acquisition technologies and innovative approaches to surface reconstruction and analysis.
In this paper, we have conducted an extensive survey of the field, benchmarking across various datasets a range of recent learning-based and optimization-based methods, alongside established traditional techniques.
Our findings reveal that learning-based methods demonstrate compelling performance when tested on point clouds bearing resemblance to their training sets. However, their effectiveness is contingent upon training with shapes of comparable complexity to the test set and they exhibit limited robustness to out-of-distribution acquisition defects. While they may lag behind in in-distribution scenarios, traditional methods consistently offer robust and reliable performance across a broader range of conditions.
We also observe that recent learning-free optimization-based methods are outperformed by these traditional methods in almost all settings.
}

{
Real-world scenes often include a variety of complex acquisition defects.
Consequently, learning-based methods face challenges in reconstructing such complex point clouds accurately. Thus, the creation or compilation of training data that encompasses a wide array of complex shapes, scanned with realistic defects, becomes crucial yet poses significant challenges. Future directions in learning-based surface reconstruction should prioritize training on point clouds with realistic acquisition defects, such as those common in typical sensor and acquisition setups, or focus on enhancing the resilience of these methods to novel and unseen defects.
}

\ifCLASSOPTIONcaptionsoff
  \newpage
\fi



%
\FloatBarrier
\bibliographystyle{ieee/IEEEtran} 
\bibliography{ieee/IEEEabrv,references}
\balance




\end{document}